\documentclass{article}

\usepackage{graphicx}
\usepackage[bookmarks=false]{hyperref}
\usepackage{cite}
\usepackage[T1]{fontenc}
\usepackage{lmodern}
\usepackage{textcomp}
\usepackage{amsmath, latexsym, amssymb, bm, ascmac, here, mathtools, comment,amsthm}
\usepackage{authblk}

\usepackage[margin=1in]{geometry}
 
\usepackage{xcolor,subfigure}

\newcommand{\RR}{\mathbb{R}}

\newcommand{\cL}{\mathcal{L}}
\newcommand{\cR}{\mathcal{R}}
\newcommand{\cS}{\mathcal{S}}
\newcommand{\cT}{\mathcal{T}}
\newcommand{\cY}{\mathcal{Y}}
\newcommand{\argmin}{\mathop{\text{argmin}}}

\newtheorem{theorem}{Theorem}[section]
\newtheorem{definition}{Definition}[section]

\begin{document}
\title{Safe Triplet Screening for Distance Metric Learning}
% \title{Safe Triplet Screening for Large Margin Metric Learning}
% \titlenote{Produces the permission block, and copyright information}
% \subtitle{Extended Abstract}
% \subtitlenote{The full version of the author's guide is available as \texttt{acmart.pdf} document}

\author[1]{Tomoki Yoshida}
\affil[1,2,3]{Nagoya Institute of Technology}
% \email{yoshida.t.mllab.nit@gmail.com}

\author[2]{Ichiro Takeuchi}
\affil[2,3]{National Institute for Material Science}
\affil[2]{RIKEN Center for Advanced Intelligence Project}
% \email{takeuchi.ichiro@nitech.ac.jp}

\author[3]{Masayuki Karasuyama}
\affil[3]{Japan Science and Technology Agency}
% \email{karasuyama@nitech.ac.jp}

\maketitle

\begin{abstract}
We study \emph{safe screening} for metric learning.
Distance metric learning can optimize a metric over a set of triplets, each one of which is defined by a pair of same class instances and an instance in a different class.
However, the number of possible triplets is quite huge even for a small dataset.
Our safe triplet screening identifies triplets which can be \emph{safely} removed from the optimization problem without losing the optimality.
Compared with existing safe screening studies, triplet screening is particularly significant because of (1) the huge number of possible triplets, and (2) the semi-definite constraint in the optimization.
We derive several variants of screening rules, and analyze their relationships.
Numerical experiments on benchmark datasets demonstrate the effectiveness of safe triplet screening.  
\end{abstract}

\begin{center}
\end{center}

\section{Introduction}

\begin{figure}[tb]
 \centering
 \includegraphics[width=\linewidth]{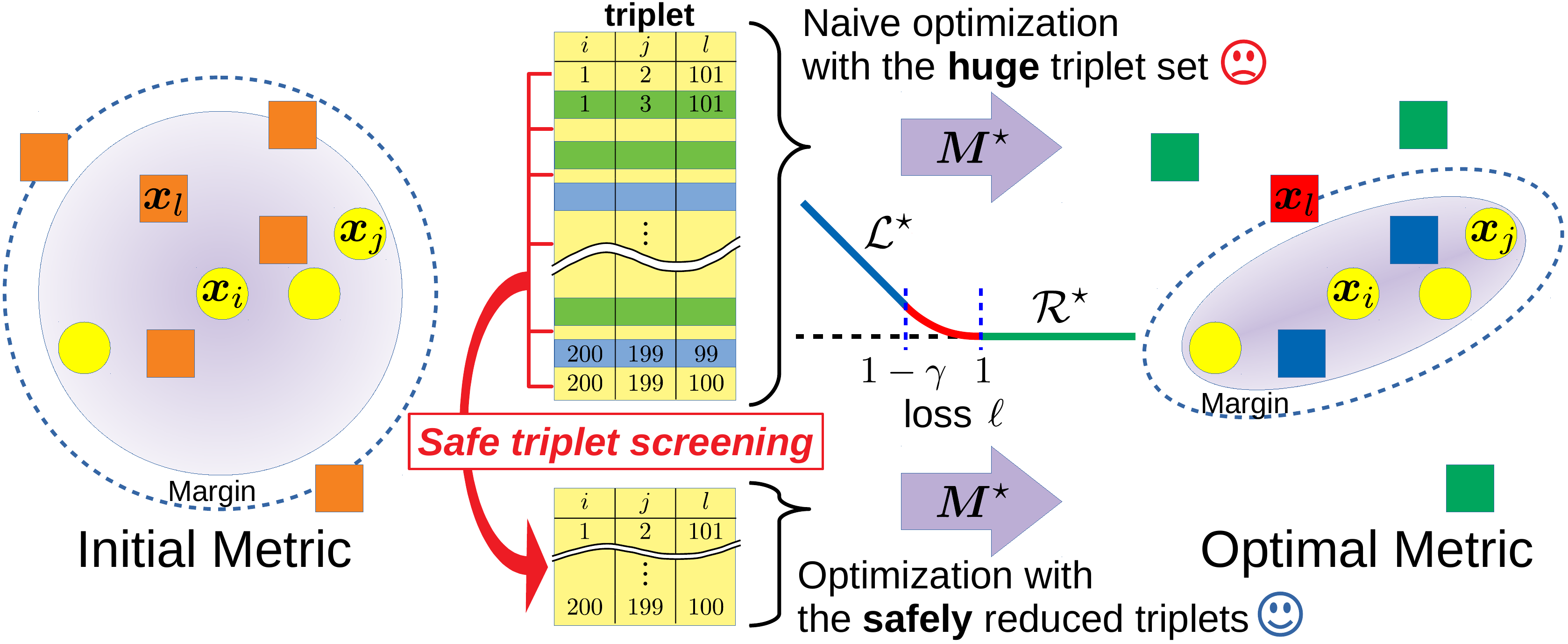}
 \caption{
 Metric learning with safe triplet screening.
 The naive optimization needs to minimize the sum of loss function values for a huge number of triplets $(i, j, l)$. 
 Safe triplet screening identifies a subset of $\cL^{\star}$ (blue points in the right drawing) and $\cR^{\star}$ (green points in the right drawing), corresponding to the location of the loss function on which each triplet lies by using the optimal $\bm{M}^\star$. 
 This enables reducing the number of triplets in the optimization problem.
 }
 \label{fig:metric_triplet}
\end{figure}

\emph{Distance metric learning} (e.g., \cite{weinberger2009distance,schultz2004learning,davis2007information,kulis2013metric}) is a widely accepted technique to acquire the optimal metric from observed data. 
The most standard problem setting is to learn the following parameterized Mahalanobis distance:
\[
 d_{\bm{M}}(\bm{x}_i,\bm{x}_j) \coloneqq \sqrt{(\bm{x}_i-\bm{x}_j)^\top \bm{M} (\bm{x}_i-\bm{x}_j)}, 
\]
where $\bm{x}_i$ and $\bm{x}_j$ are $d$-dimensional feature vectors, and $\bm{M} \in \RR^{d \times d}$ is a positive semi-definite matrix.
Using a better distance metric can provide better prediction performance for a variety of machine learning tasks including classification \cite{weinberger2009distance}, clustering \cite{xing2002distance} and ranking \cite{mcfee2002metric}.
Further, the metric optimization has also attracted wide interest even from recent deep network studies \cite{schroff2015facenet,hoffer2015deep}.

The seminal work of distance metric learning \cite{weinberger2009distance} shows a \emph{triplet} based formulation. 
A triplet $(i,j,l)$ is defined by a pair of $\bm{x}_i$ and $\bm{x}_j$ which have a same label (same class), and $\bm{x}_l$ which has a different label (different class). %  from $i$ and $j$
For a triplet $(i,j,l)$, a desirable metric would satisfy $d_{\bm{M}}(\bm{x}_i,\bm{x}_j) < d_{\bm{M}}(\bm{x}_i,\bm{x}_l)$, meaning that the same class pair is closer than the pair in different classes. 
For each one of triplets, \cite{weinberger2009distance} defines a loss function penalizing violations of this constraint, which has been widely used as a standard approach to metric learning.
Although pairwise approaches have also been considered (e.g., \cite{davis2007information}), the triplet based loss is the current standard since the relative evaluation $d_{\bm{M}}(\bm{x}_i,\bm{x}_j) < d_{\bm{M}}(\bm{x}_i,\bm{x}_l)$ is more appropriate for most metric learning application tasks such as nearest neighbor classification \cite{weinberger2009distance}, and similarity search \cite{jain2009online}.

However, a set of triplets is quite huge even for a small dataset.
For example, considering a two class problem having $100$ instances in each class, the number of possible triplets is $1,980,000$.
Since dealing with a huge number of triplets causes prohibitive computations, a small subset of triplets are sometimes used in practice (e.g., \cite{capitanine2016constraint}) though the optimality of such a sub-sampling strategy is not clearly understood.
Our \emph{safe triplet screening} enables the identification of triplets which can be \emph{safely} removed from the optimization problem without losing the optimality of the resulting metric.
This means that our approach can accelerate the time-consuming metric learning optimization with the optimality guarantee.
\figurename~\ref{fig:metric_triplet} shows a schematic illustration of safe triplet screening.

Safe screening is originally proposed for the feature selection by LASSO \cite{ghaoui2010safe}, in which unnecessary features are identified by the following procedure: (\textbf{Step 1}) Identifying a bounded region in which the optimal dual solution is guaranteed to exist, and (\textbf{Step 2}) For each one of features, verifying possibility to be selected under the condition created by \textbf{Step 1}.
%
% \begin{description}
%  \setlength{\leftskip}{1em}
%  \item[\textbf{Step 1}] 
% 	    Identifying a bounded region in which the optimal dual solution is guaranteed to exist
% 	    % must lie, based on a current feasible solution which we call \emph{reference solution}
%  \item[\textbf{Step 2}] 
% 	    For each one of features, verifying possibility to be selected under the condition created by \textbf{Step 1}
% \end{description}
%
This procedure is useful to mitigate the optimization difficulty of LASSO for high dimensional problems, and so many papers further propose a variety of approaches to creating bounded regions for obtaining a tighter bound which results in higher screening performance \cite{wang2013lasso,liu2014safe,fercoq2015mind,xiang2017screening}.
As another direction of the research, the screening idea has been applied to other learning methods including SVM non-support vector screening \cite{ogawa2013safe}, nuclear norm regularization subspace screening \cite{zhou2015safe}, and group LASSO group screening \cite{ndiaye2016safe}.
To the best of our knowledge, however, no studies have considered screening for metric learning, and our safe triplet screening is particularly significant compared with those exiting studies due to (1) the huge number of possible triplets, and (2) the \emph{semi-definite constraint}.
Our technical contributions are summarized into the following:
\begin{itemize}
	\item Deriving six sphere regions in which the optimal $\bm{M}^\star$ must lie based on three different approaches, and analyzing their relationships
	\item Deriving three types of screening rules, each one of which employs a different approach for the semi-definite constraint
	\item Building an extension for the \emph{regularization path} calculation 
\end{itemize}
We further demonstrate the effectiveness of our approach based on several benchmark datasets having a huge number of triplets.

% This paper is organized as follows. 
% %
% In section~\ref{sec:LMML}, we define the optimization problem of large margin metric learning.
% %
% Section~\ref{sec:STS} derives rules and bounds for our safe triplet screening, and section~\ref{sec:screeningRange} shows an extension for the regularization path. 
% %
% In section~\ref{sec:experiment}, we evaluate our approach through numerical experiments. 
% %
% Section~\ref{sec:summary} concludes the paper.

\subsection*{Notation}
%The notation used in this paper is described. 
We denote by $[n]$ the set $\{1,2,\ldots, n\}$ for any integer $n\in\mathbb{N}$. 
The inner product of the matrices is denoted by $\langle \bm{A}, \bm{B}\rangle\coloneqq\sum_{ij}A_{ij}B_{ij}=\mathrm{tr}(\bm{A}^\top\bm{B})$. 
The squared Frobenius norm is represented by $\left\|\bm{A}\right\|_F^2\coloneqq\langle \bm{A},\bm{A}\rangle$.
The positive semi-definite matrix $\bm{M}\in\RR^{d\times d}$ is denoted by $\bm{M}\succeq\bm{O}$ or $\bm{M}\in\RR_+^{d\times d}$.
% When eigenvalue decomposition of matrix $\bm{M}$ is $\bm{M}=\bm{V}\bm{\Lambda}\bm{V}^\top$, 
Through eigenvalue decomposition of matrix $\bm{M}=\bm{V}\bm{\Lambda}\bm{V}^\top$, matrices
$\bm{M}_+$ and $\bm{M}_-$ are defined as follows:
\[
\bm{M}=\bm{V}\underbrace{(\bm{\Lambda}_++\bm{\Lambda}_-)}_{\bm{\Lambda}}\bm{V}^\top=\underbrace{\bm{V}\bm{\Lambda}_+\bm{V}^\top}_{\eqqcolon\bm{M}_+}+\underbrace{\bm{V}\bm{\Lambda}_-\bm{V}^\top}_{\eqqcolon\bm{M}_-},
\vspace{-.4em}
\]
where $\bm{\Lambda}_+$ and $\bm{\Lambda}_-$ are constructed only by the positive and negative components of the diagonal matrix $\bm{\Lambda}$.
% Here, $\bm{\Lambda}_+, \bm{\Lambda}_-$ are respectively matrices leaving only the positive and negative components of $\bm{\Lambda}$.
Note that 
% $\langle \bm{M}_+, \bm{M}_-\rangle=0$ because
$\langle \bm{M}_+,\bm{M}_-\rangle = 
\mathrm{tr}(\bm{V}\bm{\Lambda}_+\bm{V}^\top\bm{V}\bm{\Lambda}_-\bm{V}^\top) 
% =\mathrm{tr}(\bm{V}\bm{\Lambda}_+\bm{\Lambda}_-\bm{V}^\top) 
=\mathrm{tr}(\bm{V}\bm{O}\bm{V}^\top)=0$, and
$\bm{M}_+$ is \emph{projection} of $\bm{M}$ onto the semi-definite cone, i.e., 
$\bm{M}_+ = \argmin_{\bm{A} \succeq \bm{O}} \| \bm{A} - \bm{M} \|_F^2$.
% We call $\bm{M}_+$ as \emph{projection} of $\bm{M}$ onto a semi-definite cone. 

% \section{Large Margin Metric Learning}
\section{Distance Metric Learning}
\label{sec:LMML}

We first formulate a general form of metric learning problem as a \emph{regularized triplet loss minimization} (RTLM) problem.
% We first formulate the metric learning problem based on \emph{large margin metric learning} (LMML) originally proposed by \cite{weinberger2009distance}, which is reduced to a convex optimization problem with the semi-definite constraint.
%
For later analysis, we derive primal and dual formulations, and to discuss the optimality of the learned metric, we focus on the convex formulation of RTLM in this paper.

\subsection{Primal Problem}\label{sec:primal}

Let $\{(\bm{x}_i, y_i) \mid i \in [n]\}$ be $n$ pairs of a $d$ dimensional feature vector $\bm{x}_i \in \RR^d$ and a label $y_i \in \cY$, where $\cY$ is a discrete label space.
%
% We consider a metric learning problem based on the following Mahalanobis distance:
We consider learning the following Mahalanobis distance:
\[
d_{\bm{M}}(\bm{x}_i,\bm{x}_j)\coloneqq\sqrt{(\bm{x}_i-\bm{x}_j)^\top \bm{M} (\bm{x}_i-\bm{x}_j)}, 
\]
where $\bm{M} \in \RR_+^{d \times d}$ is a positive semi-definite matrix which parameterizes distance.
%
% LMML defines ``margin'' for  a \emph{triplet} of instances defined as 
We define a \emph{triplet} of instances as follows:
\[
\mathcal{T}=\left\{(i,j,l) \mid (i,j)\in\mathcal{S},~y_i\ne y_l,~l\in[n]\right\},
\]
where $\mathcal{S}=\left\{(i,j) \mid y_i = y_j,~i\ne j,~(i,j)\in[n]\times[n]\right\}$.
The set $\mathcal{S}$ contains index pairs from the same class, and $\mathcal{T}$ represents a triplet of indices consisting of $(i,j) \in \cS$, and $l$ which is in a different class from $i$ and $j$. 
%
% The large margin loss for a triplet is defined as
We refer to the following loss as \emph{triplet loss}:
\[
\ell\left(d_{\bm{M}}^2(\bm{x}_i,\bm{x}_l)-d_{\bm{M}}^2(\bm{x}_i,\bm{x}_j)\right), 
\text{ for }(i,j,l)\in\cT,
\]
where $\ell : \RR\to\RR$ is some loss function.
For the triplet loss, we consider the hinge function
% \[
$\ell(x)=\max\{0, 1-x\}$,
% \]
or the smoothed hinge function
\[
\ell(x)=\begin{cases}
	0,								&x>1, \\
	\frac{1}{2\gamma}(1-x)^2,		&1-\gamma\le x\le 1, \\
	1-x-\frac{\gamma}{2},			&x<1-\gamma, 
\end{cases}
\]
where $\gamma > 0$ is a parameter. 
Note that the smoothed hinge includes the hinge function as a special case ($\gamma \to 0$).
The triplet loss produces a penalty if a pair $(i,j) \in \cS$ is more distant than the threshold compared with a pair $i$ and $l$ which are in difference classes.
The both of two loss functions contain the ``zero part'', in which no penalty is imposed, and the ``linear part'', in which penalty is given linearly.
% , from $1$ or $1-\gamma$, respectively.
%
Using the standard squared regularization, we consider the following RTLM as a general form of metric learning:
% the optimization problem of metric learning is written as
\begin{align} 
\tag{Primal}\label{eq:Primal}
% \min_{\bm{M}\succeq\bm{O}}~ \sum_{(i,j,l)\in \mathcal{T}}\ell\left(d_{\bm{M}}^2(\bm{x}_i,\bm{x}_l)-d_{\bm{M}}^2(\bm{x}_i,\bm{x}_j)\right)+\lambda R(\bm{M}),\\
\min_{\bm{M}\succeq\bm{O}}~
P_{\lambda}(\bm{M})\coloneqq \sum_{ijl}\ell\left(\langle \bm{M},\bm{H}_{ijl}\rangle\right)+
\frac{\lambda}{2}\left\|\bm{M}\right\|_F^2,
% \lambda  R(\bm{M}),
\end{align}
where 
$\sum_{ijl}$ denotes $\sum_{(i,j,l) \in \cT}$, 
$\bm{H}_{ijl}\coloneqq (\bm{x}_i-\bm{x}_l)(\bm{x}_i-\bm{x}_l)^\top-(\bm{x}_i-\bm{x}_j)(\bm{x}_i-\bm{x}_j)^\top$, 
% $R : \RR^{d\times d}\to \RR$ is a regularization term, 
and $\lambda>0$ is a regularization parameter. 
%the objective function can be written as
%\[
%P_{\lambda}(\bm{M})\coloneqq \sum_{ijl}\ell\left(\langle \bm{M},\bm{H}_{ijl}\rangle\right)+\lambda R(\bm{M}),
%\]
% where $\sum_{ijl}$ denotes $\sum_{(i,j,l) \in \cT}$.
% In this paper, we employ the following standard squared regularization term: 
%\[
%R(\bm{M})=\frac{1}{2}\left\|\bm{M}\right\|_F^2.
%\]

\subsection{Dual Problem}\label{sec:dual}

The dual problem is written as
\[
 \tag{Dual1}\label{eq:Dual1}
\max_{\bm{0}\le\bm{\alpha}\le\bm{1},\,\bm{\Gamma}\succeq\bm{O}} D_{\lambda}(\bm{\alpha},\bm{\Gamma})
\coloneqq
% :=
% \sum_{ijl}\ell^*(-\alpha_{ijl})
-\frac{\gamma}{2}\|\bm{\alpha}\|_2^2+\bm{\alpha}^\top\bm{1}
-\frac{\lambda}{2}\|\bm{M}_{\lambda}(\bm{\alpha},\bm{\Gamma})\|_F^2,
\]
where $\bm{\alpha} \in \RR^{|\cT|}$, which contains $\alpha_{ijl}$ for $(i,j,l) \in \cT$, and $\bm{\Gamma} \in \RR^{d \times d}$ are dual variables, and
\begin{equation}\label{eq:Mlambda}
\bm{M}_{\lambda}(\bm{\alpha}, \bm{\Gamma})\coloneqq\frac{1}{\lambda}\Bigl[\sum_{ijl}\alpha_{ijl}\bm{H}_{ijl}+\bm{\Gamma}\Bigr].
\end{equation}
We omit the derivation due to the space limitation (see Appendix~\ref{app:dual}).
Since the last term 
$\displaystyle\max_{\bm{\Gamma}\succeq\bm{O}}-\frac{1}{2}\left\|\bm{M}_{\lambda}(\bm{\alpha},\bm{\Gamma})\right\|_F^2$
is equivalent to the projection onto a semi-definite cone \cite{boyd2005least,malick2004dual}, 
the above problem \eqref{eq:Dual1} can be simplified as 
\[
	\tag{Dual2}\label{eq:Dual2}
	\max_{\bm{0}\le\bm{\alpha}\le\bm{1}}~
	D_{\lambda}(\bm{\alpha})\coloneqq
	%-\sum_{ijl}\ell^*(-\alpha_{ijl})
	-\frac{\gamma}{2}\|\bm{\alpha}\|_2^2+\bm{\alpha}^\top\bm{1}
	-\frac{\lambda}{2}\bigl\|\bm{M}_{\lambda}(\bm{\alpha})\bigr\|_F^2,
\]
where
\[
\bm{M}_{\lambda}(\bm{\alpha})\coloneqq\frac{1}{\lambda}\Bigl[\sum_{ijl}\alpha_{ijl}\bm{H}_{ijl}\Bigr]_+.
\]
For the optimal $\bm{M}^\star$, each one of triplets in $\cT$ can be categorized into the following three groups:
\begin{equation}
	\begin{split}
		\mathcal{L}^\star&\coloneqq\{(i,j,l)\in\mathcal{T} \mid \langle \bm{H}_{ijl},\bm{M}^\star \rangle < 1-\gamma \}, \\
		\mathcal{C}^\star&\coloneqq\{(i,j,l)\in\mathcal{T} \mid 1-\gamma\le\langle \bm{H}_{ijl},\bm{M}^\star \rangle \le 1 \}, \\
		\mathcal{R}^\star&\coloneqq\{(i,j,l)\in\mathcal{T} \mid \langle \bm{H}_{ijl},\bm{M}^\star \rangle > 1 \}.
	\end{split}
	\label{eq:LCR}
\end{equation}
This indicates that triplets in $\cR^\star$ is in ``zero part'', and triplets in $\cL^\star$ is in ``linear part'' of the loss function.
The well-known KKT condition provides the following relation between the optimal dual variable and the derivative of the loss function (see Appendix~\ref{app:dual} for detail):
\begin{align}
 \alpha_{ijl}^\star=-\nabla\ell(\langle \bm{M}^\star,\bm{H}_{ijl}\rangle).
 \label{eq:alpha-star}
\end{align}
Considering this equation, \eqref{eq:LCR}, and the definition of the loss function, we obtain the following rules:
% by which the following rules are obtained.
\begin{equation}
	\begin{split}
	(i,j,l)\in\mathcal{L}^\star&\Rightarrow \alpha_{ijl}^\star=1, \\
	(i,j,l)\in\mathcal{C}^\star&\Rightarrow \alpha_{ijl}^\star\in[0,1], \\
	(i,j,l)\in\mathcal{R}^\star&\Rightarrow \alpha_{ijl}^\star=0.
	\end{split}
	\label{eq:LCRdual}
\end{equation}

\section{Safe Triplet Screening}\label{sec:STS}

The nonlinear semi-definite programming problem of RTLM can be solved by the gradient methods including the primal-based \cite{weinberger2009distance}, or the dual-based approach \cite{shen2014efficient}.
However, prohibitive amount of computations can be necessary because of the huge number of triplets.
Naive calculation of the objective function requires $O(d^2 |\cT|)$ computations for both of the primal and the dual cases. 
Our \emph{safe triplet screening} can reduce the number of triplets 
by identifying a part of $\cL^\star$ and $\cR^\star$ before solving the optimization problem.

Let $\hat{\cL} \subseteq \cL^\star$ and $\hat{\cR} \subseteq \cR^\star$ be subsets of $\cL^\star$ and $\cR^\star$, respectively.
When we have $\hat{\cL}$ and $\hat{\cR}$, the optimization problem \eqref{eq:Primal} can be transformed into
% the optimization problem can be reformulated as
\[
 \tilde{P}_{\lambda}(\bm{M})
 =
 \sum_{(i,j,l)\in\mathcal{T}-\hat{\cL}-\hat{\cR}}\ell(\langle\bm{M},\bm{H}_{ijl}\rangle)
 +\frac{\lambda}{2}\|\bm{M}\|_F^2
 +\left(1-\frac{\gamma}{2}\right)|\hat{\cL}|-\langle \bm{M}, \sum_{(i,j,l)\in \hat{\cL}}\bm{H}_{ijl} \rangle.
\]
This problem differs from the original \eqref{eq:Primal} as follows
\begin{itemize}
 \item In the first term, we remove $\hat{\cR}$ which does not produce any penalty at the optimal solution
 \item The loss function for $\hat{\cL}$ is fixed at ``linear part'' of the loss function by which the sum over triplets can be calculated beforehand (the last two terms)
% replaced with a simple linear term (the last two terms) for which the sum over triplets can be calculated beforehand
\end{itemize}
Note that this problem has the same optimal $\bm{M}^\star$ as the original $P_{\lambda}(\bm{M})$.
% , while $\tilde{P}_{\lambda}(\bm{M})$ and $P_{\lambda}(\bm{M})$ are different functions.
%
% Remarkable advantages of this problem is as follows:
% \begin{itemize}
%  \item The sum of the loss term is over $\cT - \hat{\cL} - \hat{\cR}$ instead of $\cT$.
%  \item The set of triplets $(i,j,l) \in \hat{\cR}$ is not necessary.
%  \item Instead of the all $\bm{H}_{ijl}$ for $(i,j,l) \in \hat{\cL}$, only the sum $\sum_{(i,j,l) \in \hat{\cL}} \bm{H}_{ijl}$ is required.
% \end{itemize}
Therefore, if the large number of $\hat{\cL}$ and $\hat{\cR}$ can be detected beforehand, the metric learning optimization can be accelerated dramatically. 
In the case of the dual problem, the dual variables $\alpha_{ijl}$ for $(i,j,l) \in \hat{\cL}$ and $(i,j,l) \in \hat{\cR}$ can be fixed by the rule \eqref{eq:LCRdual}, and the number of variables to be optimized is reduced.

Our safe triplet screening identifies $\hat{\cL}$ and $\hat{\cR}$ by the following procedure:
\begin{description}
	\setlength{\leftskip}{1em}
	\item[\textbf{Step 1}] 
		Identifying a sphere region, in which the optimal solution $\bm{M}^\star$ must lie, based on a current feasible solution which we call \emph{reference solution}
		% Find a small area containing the optimal solution (bound) from the current solution
	\item[\textbf{Step 2}] 
		For each one of triplets $(i,j,l) \in \cT$, verifying possibility of $(i,j,l) \in \cL^\star$ or $(i,j,l) \in \cR^\star$ under the condition that $\bm{M}^\star$ is in the region
		% As long as the solution is in the bound, find constraints that do not hold and delete it from the optimization problem
\end{description}
In Section~\ref{sec:screening}, we first describe \textbf{Step 2} of this procedure, and subsequently, we derive sphere shaped regions which must contain $\bm{M}^\star$, required for \textbf{Step 1}, in Section~\ref{sec:bound}.
% In this section, we describe the procedure of \textbf{Step 2} given the sphere region. 

\subsection{Screening Rule}
\label{sec:screening}

Letting $\mathcal{B}$ be a region which contains $\bm{M}^\star$, 
the following screening rule can be derived from \eqref{eq:LCR}: 
\begin{gather*}
	\tag{R1}\label{eq:R1}
	\max_{\bm{X}\in\mathcal{B}}~\langle \bm{X}, \bm{H}_{ijl}\rangle
	< 1-\gamma 
	%\Rightarrow \langle \bm{M}^\star, \bm{H}_{ijl}\rangle < 1-\gamma
	\Rightarrow (i,j,l)\in\mathcal{L}^\star
	\\
	\tag{R2}\label{eq:R2}
	\min_{\bm{X}\in\mathcal{B}}~\langle \bm{X}, \bm{H}_{ijl}\rangle
	> 1 
	%\Rightarrow \langle \bm{M}^\star, \bm{H}_{ijl}\rangle > 1
	\Rightarrow (i,j,l)\in\mathcal{R}^\star.
\end{gather*}
We will show how to evaluate these rules efficiently. % , and further show that the semi-definite constraint can also be incorporated into the conditions for obtaining tighter bounds.
Since \eqref{eq:R1} can be evaluated in the same way as \eqref{eq:R2}, we only deal with \eqref{eq:R2} hereafter.

\subsubsection{Sphere Rule}
\label{sec:noSD}
Suppose that the optimal $\bm{M}^\star$ lies in a hypersphere defined by a center $\bm{Q} \in \RR^{d \times d}$ and a radius $r \in \RR_+$.
% the center of the hypersphere $\mathcal{B}$ is $\bm{Q} \in \RR^{d \times d}$ and the radius is $r \in \RR_+$, 
To evaluate the condition of \eqref{eq:R2}, we consider the following minimization problem \eqref{eq:P1}: 
\[
\tag{P1}\label{eq:P1}
\min_{\bm{X}}~\langle \bm{X}, \bm{H}_{ijl}\rangle~~\mathrm{s.t.}~\left\|\bm{X}-\bm{Q}\right\|_F^2\le r^2.
\]
Letting $\bm{Y}\coloneqq\bm{X}-\bm{Q}$, this problem is transformed into
\[
\min_{\bm{Y}}~\langle \bm{Y}, \bm{H}_{ijl}\rangle+\langle \bm{Q}, \bm{H}_{ijl}\rangle~~\mathrm{s.t.}~\left\|\bm{Y}\right\|_F^2\le r^2.
\]
Since $\langle \bm{Q}, \bm{H}_{ijl}\rangle$ is a constant, 
this optimization problem is to minimize the inner product $\langle \bm{Y}, \bm{H}_{ijl}\rangle$ under the norm constraint.
The optimal $\bm{Y}^\star$ of this optimization problem is easily derived as
\[
\bm{Y}^\star=-r {\bm{H}_{ijl}} / {\left\|\bm{H}_{ijl}\right\|_F},
\]
and then the minimum value of \eqref{eq:P1} is
%\[
$\langle \bm{H}_{ijl}, \bm{Q}\rangle-r\left\|\bm{H}_{ijl}\right\|_F$.
%\]
% \figurename~\ref{fig:noSD} shows a schematic illustration.
This derives the following \emph{sphere rule}:
\begin{equation}
	\label{eq:sphere-ruleR2}
	\langle \bm{H}_{ijl}, \bm{Q}\rangle-r\left\|\bm{H}_{ijl}\right\|_F > 1
	\Rightarrow 
	(i,j,l) \in \cR^\star.
\end{equation}
Obviously, this condition can be calculated immediately for given $\bm{Q}$ and $r$ without any iterative procedure. 

% \begin{figure}[tb]
% 	\centering
% 	\includegraphics[width=0.5\linewidth]{figure/noSD.pdf}
% 	\caption{
% 	Sphere rule defined by \eqref{eq:P1}.
% 	The yellow sphere indicates the region in which the optimal $\bm{M}^\star$ must exist.
% 	The ``$\max$'' and ``$\min$'' respectively indicate the points that the maximum and the minimum value of 
% 	the inner product $\langle \bm{X}, \bm{H}_{ijl}\rangle$ are attained. 
% 	If $\langle \bm{X}^\star, \bm{H}_{ijl}\rangle > 1$ holds, the condition \eqref{eq:R2} is guaranteed to be satisfied.
% 	}
% 	\label{fig:noSD}
% \end{figure}

\subsubsection{Sphere Rule with Semi-definite Constraint}
\label{sec:sphere+SD}

Since sphere rule does not utilize the positive semi-definiteness of $\bm{M}^\star$, 
a stronger rule can be constructed by incorporating semi-definite constraint into \eqref{eq:P1}:
\[
\tag{P2}\label{eq:P2}
%\begin{array}{cc}\displaystyle
	\min_{\bm{X}}~\langle \bm{X}, \bm{H}_{ijl}\rangle~~
	\mathrm{s.t.}~\left\|\bm{X}-\bm{Q}\right\|_F^2\le r^2,~\bm{X}\succeq\bm{O}.
%\end{array}
\]
Although the analytical solution is not available, 
\eqref{eq:P2} can be solved efficiently by being transformed into
the \emph{Semi-Definite Least Squares} (SDLS) problem \cite{malick2004dual}.

Suppose that a feasible solution $\bm{X}_0$ of \eqref{eq:P2} satisfies $\langle \bm{X}_0,\bm{H}_{ijl}\rangle>1$,
because if $\langle \bm{X}_0,\bm{H}_{ijl}\rangle\le 1$, we immediately see that this triplet does not satisfy the condition of \eqref{eq:R2}.
Under this condition, we consider the following problem instead of \eqref{eq:P2}:
\[
\tag{SDLS}\label{eq:SDLS}
%\begin{array}{cc}\displaystyle
\min_{\bm{X}\in\mathbb{R}^{d\times d}}~\left\|\bm{X}-\bm{Q}\right\|_F^2~~
\mathrm{s.t.}~\langle \bm{X},\bm{H}_{ijl}\rangle=1,~\bm{X}\succeq\bm{O}.
%\end{array}
\]
If the optimal value of this problem is greater than $r^2$, i.e., $\| \bm{X}^\star - \bm{Q} \|_F^2 > r^2$, we can deduce
\[
\left\{\bm{X}~\middle|~
\langle \bm{X},\bm{H}_{ijl}\rangle \leq 1,~
\left\|\bm{X}-\bm{Q}\right\|_F^2\le r^2,~
\bm{X}\succeq\bm{O}
\right\} =
\emptyset,
\]
which indicates that the condition of \eqref{eq:R2} is satisfied. 

% The problem \eqref{eq:P2} can be transformed into 

We derive the following dual problem of \eqref{eq:SDLS} based on \cite{malick2004dual}:
\[
\max_{y}~
D_{\mathrm{SDLS}}(y) \coloneqq
-\left\|[\bm{Q}+y\bm{H}_{ijl}]_+\right\|_F^2+2Cy+\left\|\bm{Q}\right\|_F^2,
% \underbrace{-\left\|[\bm{Q}+y\bm{H}_{ijl}]_+\right\|_F^2+2Cy+\left\|\bm{Q}\right\|_F^2}_{\eqqcolon D_{\mathrm{SDLS}}(y)},
\]
where $y \in \RR$ is a dual variable, and $C = 1$ for \eqref{eq:R2} and $C = 1 - \gamma$ for \eqref{eq:R1}.
Unlike the primal problem, the dual is an unconstrained problem which only has one variable $y$, 
and thus standard gradient-based algorithms rapidly converge.
We call the quasi-Newton optimization for this problem \emph{SDLS dual ascent method}. 
During the dual ascent, we can stop the iteration before convergence 
if $D_{\mathrm{SDLS}}(y)$ becomes larger than $r^2$, 
since the value of the dual problem does not exceed the value of the primal problem (weak duality).

Although the computation of $[\bm{Q}+y\bm{H}_{ijl}]_+$ requires eigenvalue decomposition, 
this computational requirement can be alleviated when the center $\bm{Q}$ of the hypersphere is positive semi-definite.
From the definition, $\bm{H}_{ijl}$ has at most one negative eigenvalue, 
and then $\bm{Q}+y\bm{H}_{ijl}$ also has at most one negative eigenvalue.
Let $\lambda_{\min}$ be the negative (minimum) eigenvalue of $\bm{Q}+y\bm{H}_{ijl}$, 
and $\bm{q}_{\min}$ be the corresponding eigenvector.
The projection $[\bm{Q}+y\bm{H}_{ijl}]_+$ can be expressed as %follows 
%\[
$[\bm{Q}+y\bm{H}_{ijl}]_+=(\bm{Q}+y\bm{H}_{ijl})-\lambda_{\min}\bm{q}_{\min}\bm{q}_{\min}^\top$.
%\]
Computation of the minimum eigenvalue and eigenvector is much easier than the full eigenvalue decomposition \cite{lehoucq1996deflation}. 

As a special case, when $\bm{M}$ is a diagonal matrix, the semi-definite constraint is reduced to the non-negative constraint, and analytical calculation of the rule \eqref{eq:P2} is possible (see Appendix~\ref{app:P3}). 

% We further consider a special case that $\bm{M}$ is a diagonal matrix because we can evaluate screening rules with the semi-definite constraint much easier than the general case. 
% When $\bm{M}$ is a diagonal matrix, the semi-definite constraint is reduced to the nonnegativity constraint. 
% Then, the minimization problem \eqref{eq:P2} is simplified as
% \[
% \tag{P3}\label{eq:P3}
% %	\begin{array}{cc}\displaystyle
% 		\min_{\bm{x}\in\mathbb{R}^d}~ \bm{x}^\top\bm{h}_{ijl}~~
% 		\mathrm{s.t.}~\left\|\bm{x}-\bm{q}\right\|_2^2\le r^2,~\bm{x}\ge\bm{0}, 
% %	\end{array}
% \]
% where $\bm{h}_{ijl}\coloneqq\mathrm{diag}(\bm{H}_{ijl})$. 
% Considering the KKT condition of \eqref{eq:P3}, this optimization problem can be solved analytically with $O(d^2)$ 
% (see Appendix~\ref{app:P3}). 
% SDLS dual ascent is also applicable, 
% which can be faster than the analytical calculation for high dimensional case because the computations required for one iteration is $O(d)$. 

\subsubsection{Sphere Rule with Linear Constraint}
\label{sec:sphere+Linear}
% To deal with semi-definite constraint for $\bm{M}$ in the screening rule, the eigenvalue decomposition is required as we mentioned in section~\ref{sec:sphere+SD}.
To reduce computational complexity, we here consider relaxing the semi-definite constraint into a linear constraint.
Suppose that a region defined by a linear inequality 
$\{ \bm{X} \in \RR^{d \times d} \mid \langle \bm{P}, \bm{X}\rangle \ge 0 \}$ contains the semi-definite cone, 
i.e., $\RR_+^{d \times d} \subseteq \{ \bm{X} \in \RR^{d \times d} \mid \langle \bm{P}, \bm{X}\rangle \ge 0 \}$, 
for which we will describe how to obtain $\bm{P} \in \RR^{d \times d}$ later.
Using this relaxed constraint, the condition \eqref{eq:R2} is 
\[
	\tag{P4}\label{eq:P4}
%	\begin{array}{cc}\displaystyle
	\min_{X}~ \langle \bm{X}, \bm{H}_{ijl}\rangle~~
	\mathrm{s.t.}~\left\|\bm{X}-\bm{Q}\right\|_F^2\le r^2,~\langle \bm{P}, \bm{X}\rangle \ge 0.
%	\end{array}
\]
This problem can be solved analytically by considering the KKT condition 
as follows (Appendix~\ref{app:solveP4}). 

\begin{theorem}[Analytical solution of \eqref{eq:P4}] \label{thm:P4}
 The optimal solution of \eqref{eq:P4} is as follows:
	\[
	\langle\bm{H}_{ijl}, \bm{X}^\star \rangle=
	\begin{cases}
		0,																	& \text{ if } \bm{H}_{ijl} = a \bm{P},\\
		\langle\bm{H}_{ijl},\bm{Q}\rangle-r\|\bm{H}_{ijl}\|_F,			& \text{ if } \langle \bm{P},\bm{Q}-r\frac{\bm{H}_{ijl}}{\|\bm{H}_{ijl}\|_F} \rangle\ge0,\\
		\langle\bm{H}_{ijl}, \frac{\beta\bm{P}-\bm{H}_{ijl}}{\alpha}+\bm{Q}\rangle,&\mathrm{otherwise},
	\end{cases}
	\]
	where $a$ is a constant, and 
	\[
	\alpha=\sqrt{\frac{\|\bm{P}\|_F^2\|\bm{H}_{ijl}\|_F^2-\langle \bm{P},\bm{H}_{ijl}\rangle^2}{r^2\|\bm{P}\|_F^2-\langle\bm{P},\bm{Q}\rangle^2}},
	\beta=\frac{\langle \bm{P},\bm{H}_{ijl}\rangle-\alpha\langle \bm{P},\bm{Q}\rangle}{\|\bm{P}\|_F^2}.
	\]
\end{theorem}\noindent
% See Appendix~\ref{app:solveP4} for the proof.

\begin{figure}[tb]
	\centering
	\includegraphics[width=0.4\linewidth]{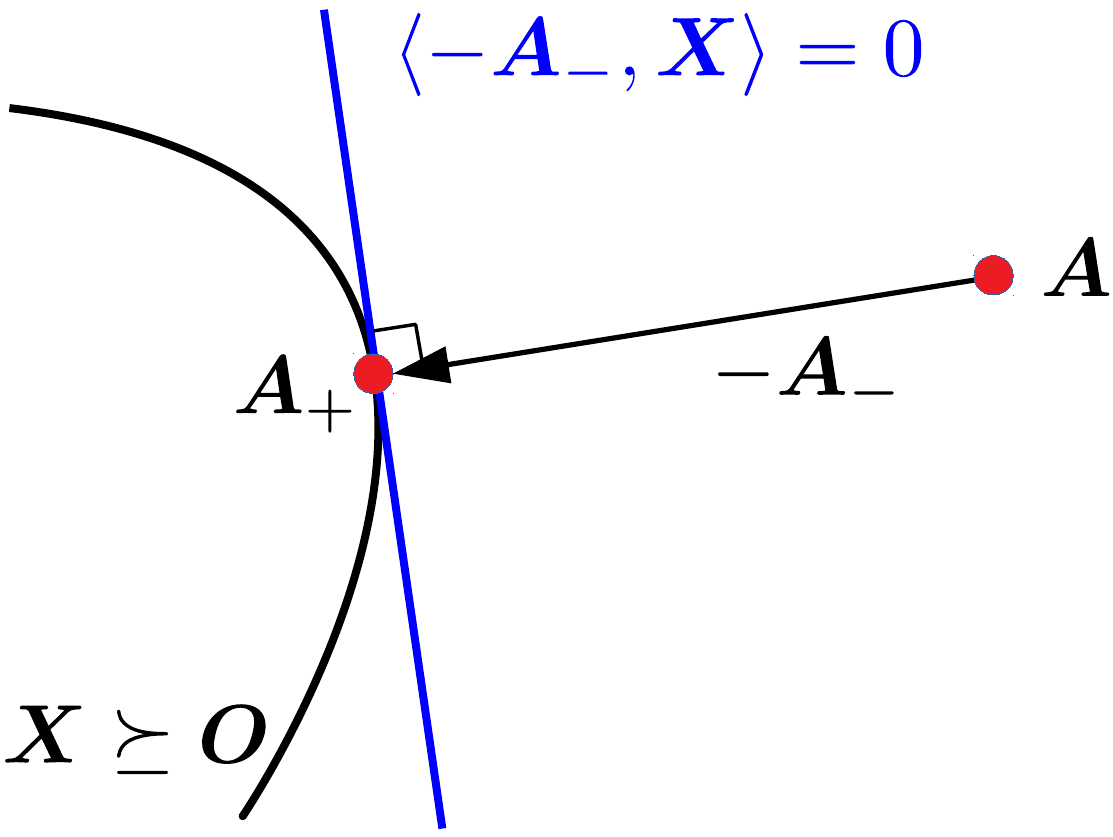}
	\caption{
	Linear relaxation of semi-definite constraint. 
	From the projection of $\bm{A}$ to $\bm{A}_+$, the supporting hyperplane $\langle - \bm{A}_-, \bm{X} \rangle = 0$ is constructed, 
	and the halfspace $\{ \bm{X} \mid \langle - \bm{A}_-, \bm{X} \rangle \geq 0 \}$ contains the semi-definite cone $\bm{X} \succeq \bm{O}$. 
	}
	\label{fig:linear}
\end{figure}

A simple way to obtain $\bm{P}$ is to utilize the projection onto the semi-definite cone.
Let $\bm{A} \in \RR^{d \times d}$ be a matrix which is in outside of the semi-definite cone as illustrated in \figurename~\ref{fig:linear}.
In the figure, $\bm{A}_+$ is the projection of $\bm{A}$ onto the semi-definite cone.
For example, when the projected gradient for the primal problem \cite{weinberger2009distance} is used as an optimizer, $\bm{A}$ can be an update of gradient descent $\bm{A} = \bm{M} - \eta \nabla P_\lambda(\bm{M})$ with some step size $\eta > 0$. 
% In practice, when the optimizer is the projected gradient in primal \cite{weinberger2009distance}.$\bm{A}$ can be $\bm{A} = \bm{M} - \eta \nabla P_\lambda(\bm{M})$ with some step size $\eta > 0$ 
Since $\bm{M} - \eta \nabla P_{\lambda}(\bm{M})$ is projected onto the semi-definite cone at every iteration of the optimization, 
no additional calculation is required to obtain $\bm{A}$ and $\bm{A}_+$.
Defining $\bm{A}_- \coloneqq \bm{A} - \bm{A}_+$, for any $\bm{X} \succeq \bm{O}$, we obtain
\[
\langle \bm{A}_+-\bm{A}, \bm{X}-\bm{A}_+\rangle \ge 0
~\Leftrightarrow~\langle -\bm{A}_-, \bm{X}\rangle \ge 0.
\]
The left inequality is from a property of a supporting hyperplane \cite{boyd2004convex}, and for the right inequality, we use $\langle \bm{A}_+, \bm{A}_- \rangle = 0$.
By setting $\bm{P} = - \bm{A}_-$, we obtain a linear approximation of the semi-definite constraint 
which is a superset of the original semi-definite cone.

\subsection{Sphere Bound}\label{sec:bound}
In previous section, we assume that the sphere region
% $\cB$ 
which contains the optimal $\bm{M}^\star$ is available.
% In previous section, we assume that we can obtain the sphere region $\cB$ which contains the optimal $\bm{M}^\star$.
% The sphere $\cB$ can be derived by various approaches.
In this section, we show that six variants of the regions created by three-types of different approaches.
We here omit detailed derivation, and see Appendix for the proofs.

\subsubsection{Gradient Bound (GB)}

We first introduce a hypersphere which we call \emph{Gradient Bound} (GB) because the center and radius of the hypersphere are represented by the subgradient of the objective function:
\begin{theorem}[GB]
	\label{thm:GB}
	Given any feasible solution $\bm{M} \succeq \bm{O}$, the optimal solution $\bm{M}^\star$ for $\lambda$ exists in the following hypersphere:
 \[
 \left\|\bm{M}^\star-
 \bm{Q}^{\mathrm{GB}}(\bm{M})
%\left(\bm{M}-\frac{1}{2\lambda}\nabla P_{\lambda}(\bm{M})\right)
 \right\|_F^2
 \le
 \left(\frac{1}{2\lambda}\left\|\nabla P_{\lambda}(\bm{M})\right\|_F\right)^2,
 \]
 where $\bm{Q}^{\mathrm{GB}}(\bm{M}) \coloneqq \bm{M}-\frac{1}{2\lambda}\nabla P_{\lambda}(\bm{M})$.
\end{theorem}\noindent
\begin{proof}%[proof]
	% In convex optimization, the following theorem holds generally:
	% \begin{theorem}
	% 	\textbf{(Optimality condition of convex optimization, \cite{bertsekas1999nonlinear}).}
	% 	\label{thm:cvxopt}
	% 	In the minimization problem 
	% 	$\min_{\bm{x}\in\mathcal{F}}f(\bm{x})$
	% 	where the feasible region $\mathcal{F}$ and the function $f(\bm{x})$ are convex, 
	% 	the necessary and sufficient condition that $\bm{x}^\star$ is the optimal solution is 
	% 	\[
	% 	\exists \nabla f(\bm{x}^\star)\in\partial f(\bm{x}^\star)\left[\nabla f(\bm{x}^\star)^\top (\bm{x}^\star-\bm{x})\le0,~\forall \bm{x}\in\mathcal{F}\right], 
	% 	\]
	% 	where $\partial f(\bm{x}^\star)$ represents the set of subgradient in $\bm{x}^\star$.
	% \end{theorem}\noindent
 From the standard optimality condition of the convex optimization problem \cite{bertsekas1999nonlinear} (shown as \textbf{Theorem}~\ref{thm:cvxopt} in our Appendix~\ref{app:GB}), we obtain
% 	In metric learning, the feasible region $\cF$ is a set of positive semi-definite matrices, and the condition is
	\begin{equation}
		\label{eq:primalopt}
		\langle \nabla P_{\lambda}(\bm{M}^\star), \bm{M}^\star-\bm{M}\rangle \le 0,~ \forall \bm{M} \succeq \bm{O}.
	\end{equation}
	In addition to this condition, we use the following two inequalities derived from the convexity of $\ell$:
	\begin{gather*}
	\ell(\langle\bm{M}^\star,\bm{H}_{ijl}\rangle)\ge\ell(\langle\bm{M},\bm{H}_{ijl}\rangle)+\langle \bm{\Xi}_{ijl}(\bm{M}),\bm{M}^\star-\bm{M}\rangle, \\
	\ell(\langle\bm{M},\bm{H}_{ijl}\rangle)\ge\ell(\langle\bm{M}^\star,\bm{H}_{ijl}\rangle)+\langle \bm{\Xi}_{ijl}(\bm{M}^\star),\bm{M}-\bm{M}^\star\rangle,
	\end{gather*}
	where $\bm{\Xi}_{ijl}(\bm{M})$ is an arbitrary subgradient at $\bm{M}$ of the loss function $\ell(\langle \bm{M},\bm{H}_{ijl}\rangle)$.
	\textbf{Theorem} \ref{thm:GB} is derived by combining three inequality shown above.
 See Appendix~\ref{app:GB} for the proof.
\end{proof}\noindent
This theorem is an extension of the sphere for SVM \cite{shibagaki2015regularization}, which can be treated as a simple unconstrained problem.
% We call this hypersphere \emph{Gradient Bound} (GB) because the right-hand-side contains the norm of the subgradient as a radius. 

\begin{figure*}[t]
 % \centering
 \subfigure[
 Projected Gradient Bound (PGB).
 The center of GB $\bm{Q}^{\rm GB}$ is projected onto the semi-definite cone.
 The radius is calculated as
 $r_{\rm PGB}^2=r_{\rm GB}^2-\|\bm{Q}^{\rm GB}-\bm{Q}_+^{\rm GB}\|_F^2$.
 ]{\includegraphics[width=0.28\linewidth]{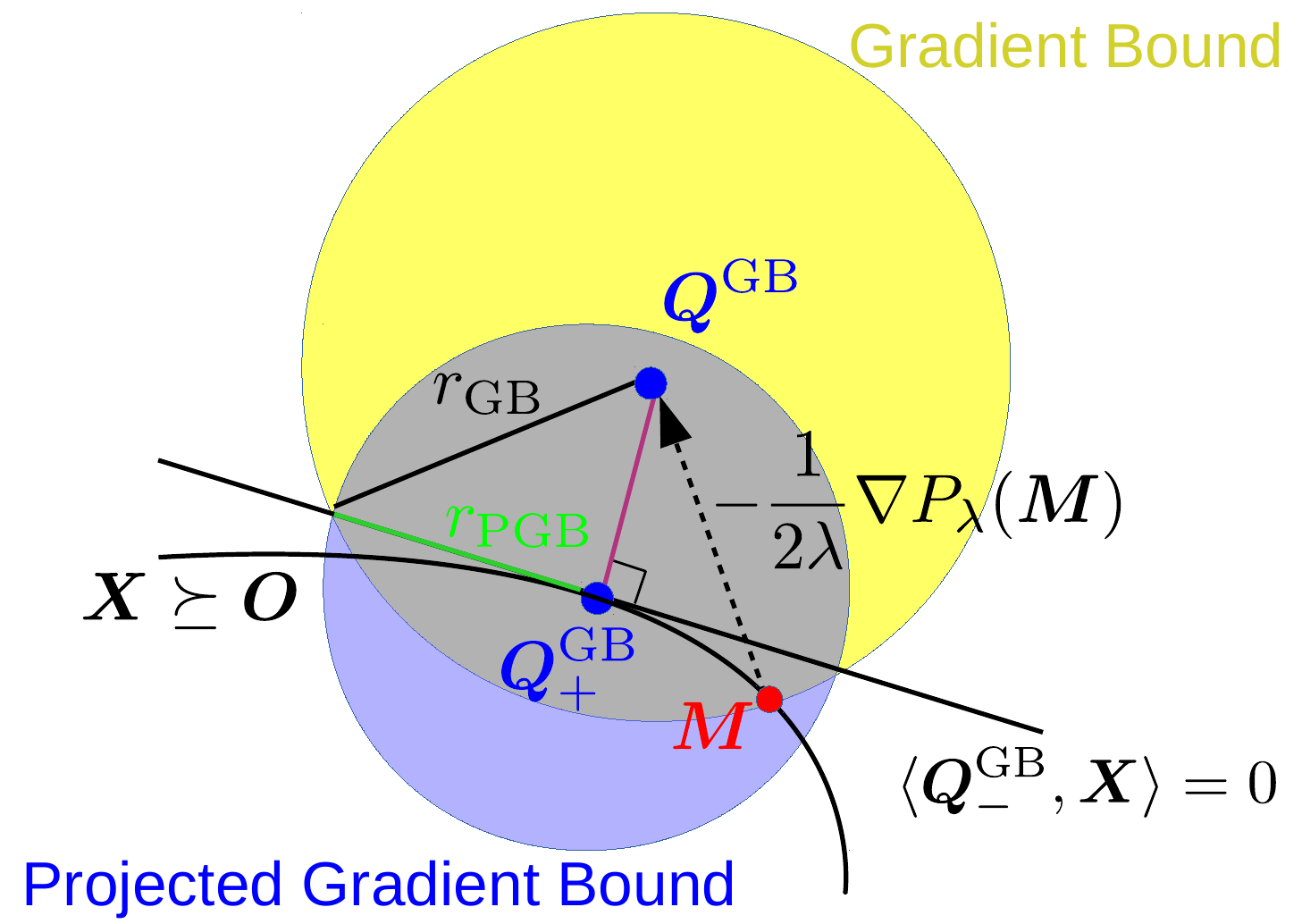}
 \label{fig:project}
 } 
 \hfill
 \subfigure[
 Duality Gap Bound (DGB). 
 The center is the reference solution, and the radius is determined from the duality gap.
 ]{\includegraphics[width=0.21\linewidth]{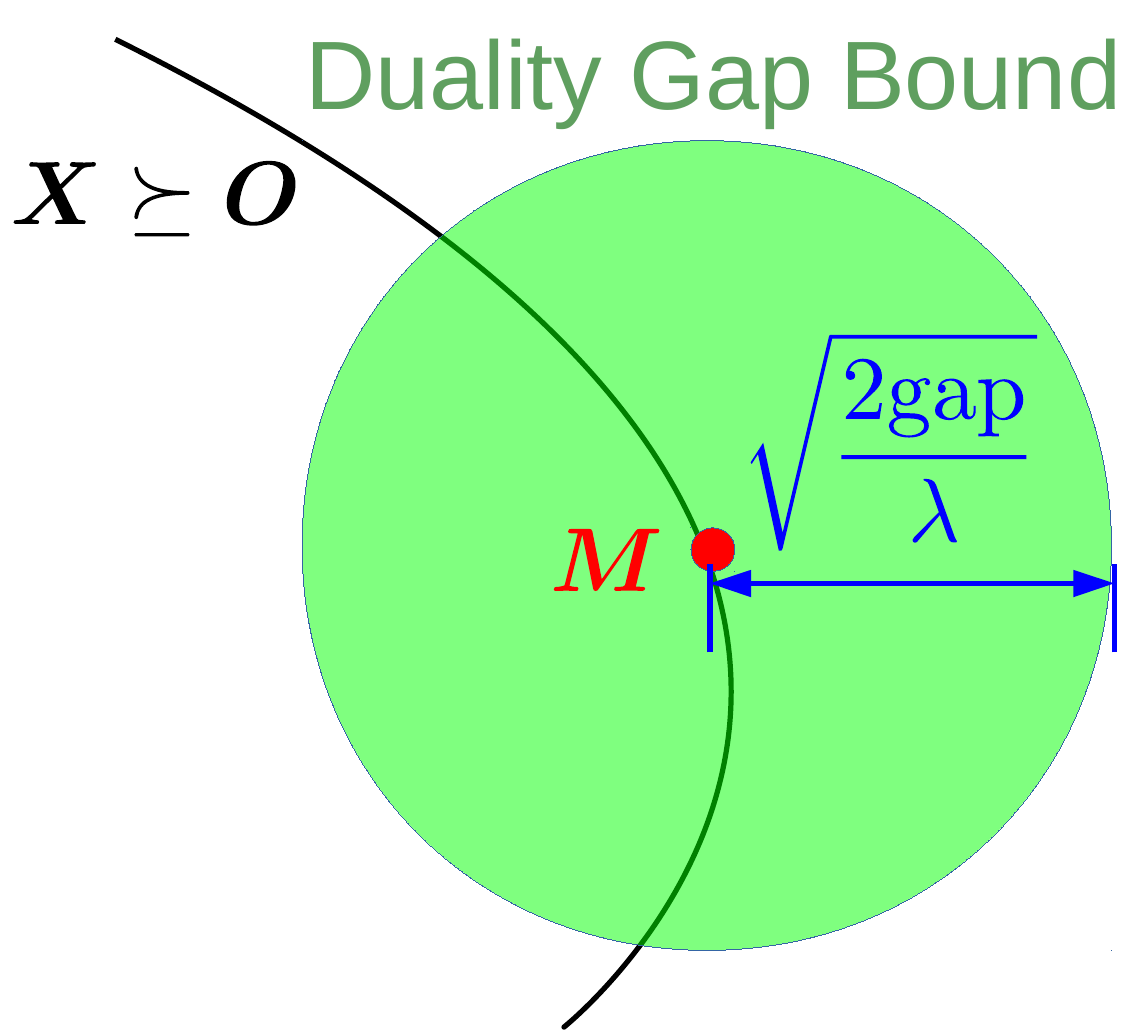}
 \label{fig:gap}
 }
 \hfill
 \subfigure[
 Relation between DGB and RPB (\textbf{Theorem} \ref{thm:DGBvsRPB}).
 ]{\includegraphics[width=0.23\linewidth]{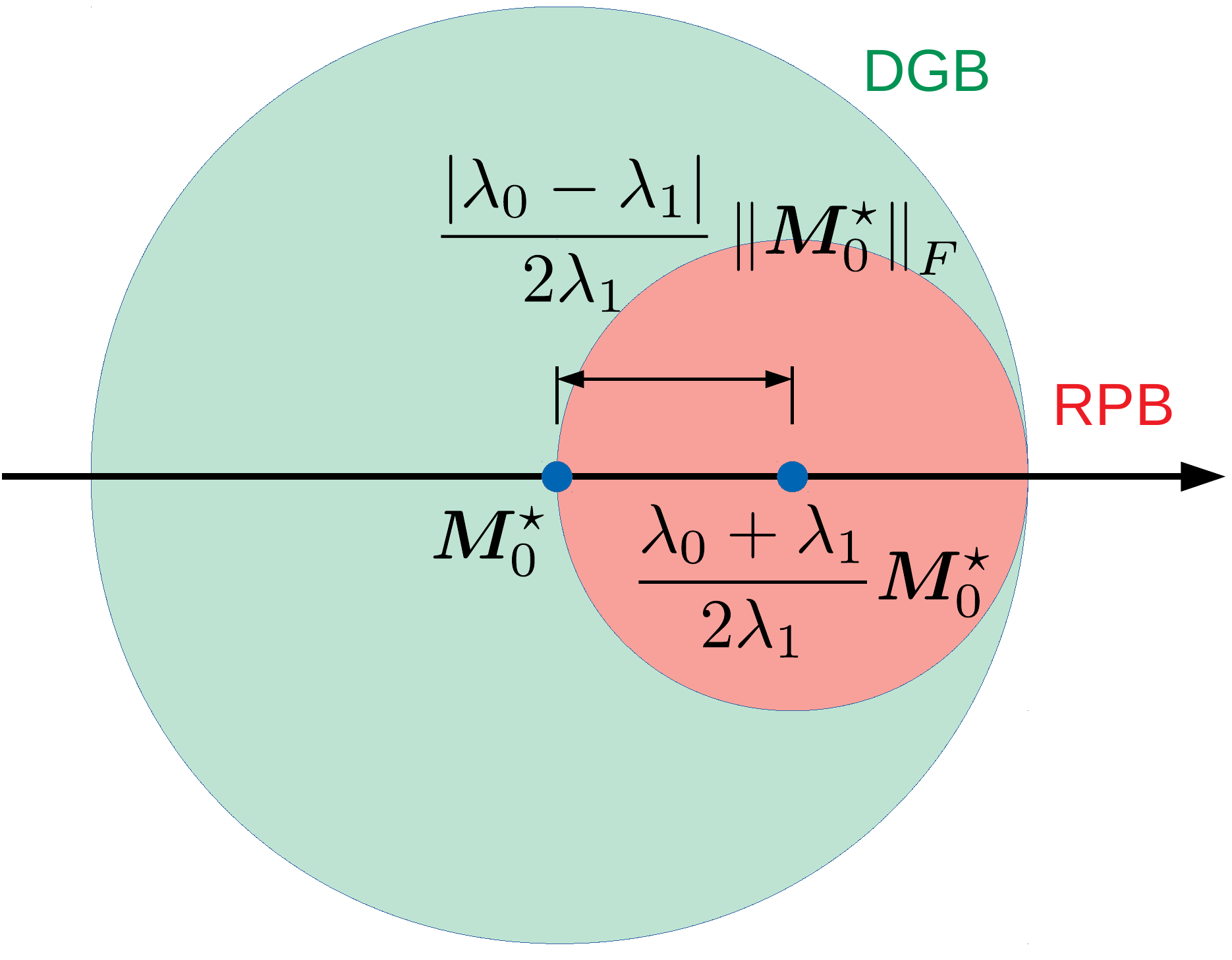}
 \label{fig:DGBvsRPB}
 }
 \hfill
 \subfigure[
 Relaxed Regularization Path Bound (RRPB).
 RRPB ``blurs'' the center of RPB based on the current approximate solution. 
 ]{\includegraphics[width=0.21\linewidth]{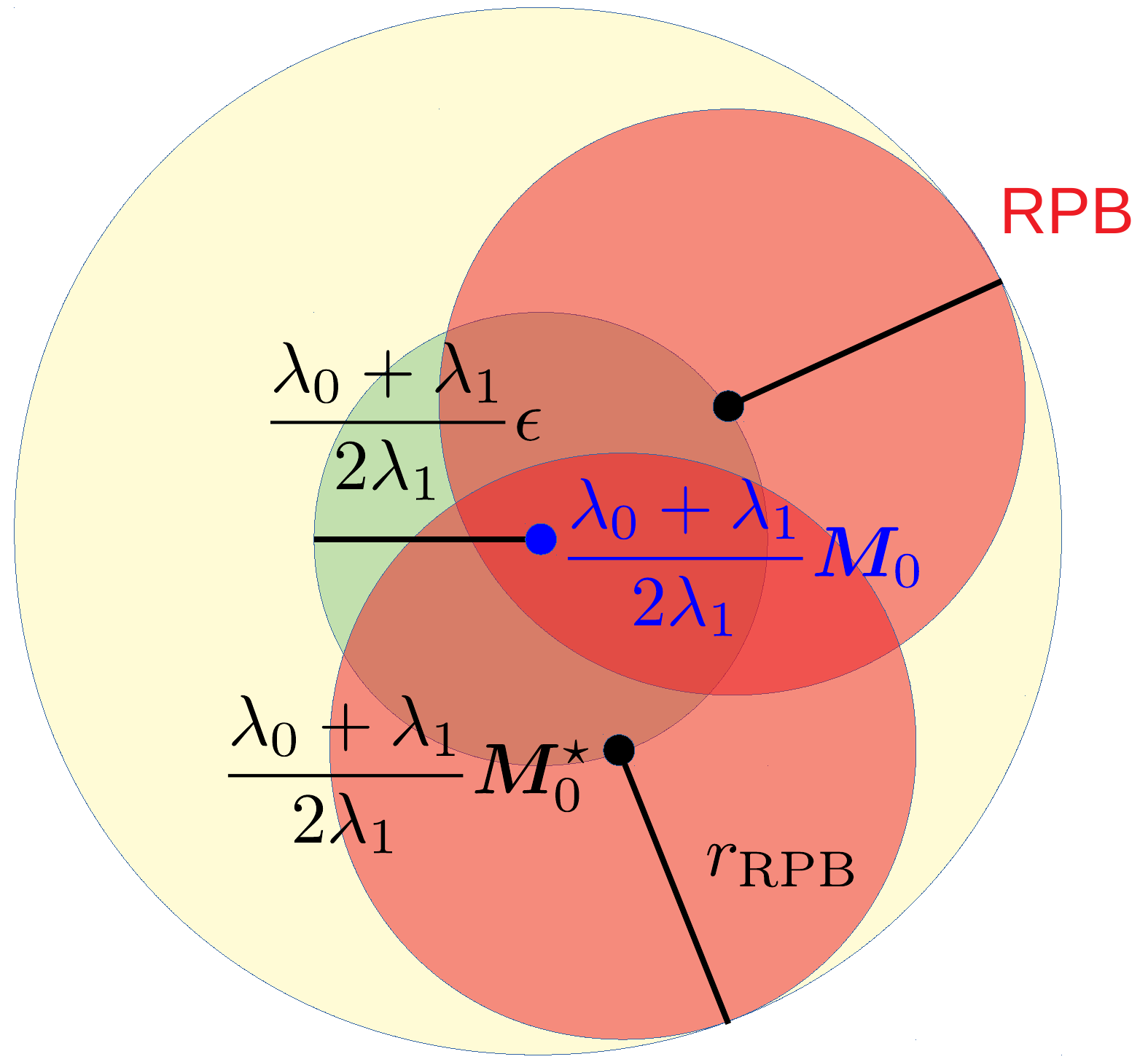}
 \label{fig:RRPB}
 }
 \caption{Illustrations of sphere bounds.}
\end{figure*}

Even when we substitute the optimal $\bm{M}^\star$ into the reference solution $\bm{M}$, the radius of GB is not guaranteed to be $0$.
By projecting the center of GB onto the feasible region (i.e., semi-definite cone), another GB based hypersphere can be derived, which has a radius converging to $0$ at the optimal.
We call this extension \emph{Projected Gradient Bound} (PGB) for which a schematic illustration is shown as \figurename~\ref{fig:project}.
In \figurename~\ref{fig:project}, the center of GB 
$\bm{Q}^{\rm GB}$ (abbreviation of $\bm{Q}^{\rm GB}(\bm{M})$)
% $\bm{Q}^{\rm GB} (\coloneqq \bm{Q}^{\rm GB}(\bm{M}))$ 
is projected onto the semi-definite cone which becomes a center of PGB $\bm{Q}_+^{\rm GB}$. %, and the radius is also shown as $r_{\rm PGB}$.
The sphere of PGB can be written as follows:
\begin{theorem}[PGB]
 \label{thm:PGB}
 Given any feasible solution $\bm{M} \succeq \bm{O}$, the optimal solution $\bm{M}^\star$ for $\lambda$ exists in the following hypersphere:
 \[
  \left\|\bm{M}^\star-\left[
  \bm{Q}^{\mathrm{GB}}(\bm{M})
  % \bm{M}-\frac{1}{2\lambda}\nabla P_{\lambda}(\bm{M})
  \right]_+\right\|_F^2
  % \\
  \le\left(\frac{1}{2\lambda}\left\|\nabla P_{\lambda}(\bm{M})\right\|_F\right)^2
  -\left\|\left[
  \bm{Q}^{\mathrm{GB}}(\bm{M})
  % \bm{M}-\frac{1}{2\lambda}\nabla P_{\lambda}(\bm{M})
  \right]_-\right\|_F^2.
 \]
\end{theorem}\noindent
See Appendix~\ref{app:PGBproof} for the proof.
% \begin{figure}[tb]
% 	\centering
% 	\includegraphics[width=0.9\linewidth]{figure/project.pdf}
% 	\caption{
% 	Projected Gradient Bound (PGB).
% 	%
% 	The center of GB $\bm{Q}^{\rm GB}$ is projected onto the semi-definite cone.
% 	%
% 	The radius is calculated as
% 	$r_{\rm PGB}^2=r_{\rm GB}^2-\|\bm{Q}^{\rm GB}-\bm{Q}_+^{\rm GB}\|_F^2$.}
% 	\label{fig:project}
% \end{figure}
% \begin{figure*}[t]
%  \begin{subfigure}[t]{.26\textwidth}
%   \includegraphics[width=\linewidth]{figure/project.pdf}
%   \caption{
%  Projected Gradient Bound (PGB).
%  %
%  The center of GB $\bm{Q}^{\rm GB}$ is projected onto the semi-definite cone.
%  %
%  The radius is calculated as
%  $r_{\rm PGB}^2=r_{\rm GB}^2-\|\bm{Q}^{\rm GB}-\bm{Q}_+^{\rm GB}\|_F^2$.
%   }
%  \end{subfigure}
%  \begin{subfigure}[t]{.23\textwidth}
%   \includegraphics[width=\linewidth]{figure/gap.pdf}
%   \caption{
%   Duality Gap Bound (DGB). 
%   %
%   The center is the reference solution, and the radius is determined from the duality gap.
%   }
%  \end{subfigure}
%  \begin{subfigure}[t]{.23\textwidth}
%   \includegraphics[width=\linewidth]{figure/DGBvsRPB.pdf}
%   \caption{
%  Relation between DGB and RPB (\textbf{Theorem} \ref{thm:DGBvsRPB}).}
%  \end{subfigure}
%  \begin{subfigure}[t]{.25\textwidth}
%   \includegraphics[width=\linewidth]{figure/RRPB.pdf}
%   \caption{
%  Relaxed Regularization Path Bound (RRPB).
%  %
%   RRPB ``blurs'' the center of RPB based on the current approximate solution. 
%   }
%  \end{subfigure}
%  \caption{}
% \end{figure*}
PGB contains the projections onto the positive and the negative semi-definite cone in the center and the radius, respectively.
These projections require the eigenvalue decomposition of $\bm{M}-\frac{1}{2\lambda}\nabla P_{\lambda}(\bm{M})$.
This decomposition, however, is necessary to perform only once for evaluating screening rules of all triplets.
In the standard optimization procedures of RTLM, including \cite{weinberger2009distance}, 
the eigenvalue decomposition of the $d \times d$ matrix is calculated at every iteration, 
and thus the computational complexity is not increased by PGB.

The following theorem shows a superior convergence property of PGB compared to GB:
\begin{theorem}
	There exist a subgradient $\nabla P_{\lambda}(\bm{M}^\star)$ such that the radius of PGB is $0$.
\end{theorem}\noindent
For the hinge loss, which is not differentiable at the ``kink'', the optimal dual variables provide subgradients which make the radius $0$.
This theorem is an immediate consequence from Appendix~\ref{app:PGBvsRPB}, which is a proof for the relation between PGB and the other bound derived in section~\ref{sec:RPB}.

% This theorem is proved by substituting $\lambda_0 = \lambda_1$ in the Appendix~\ref{app:PGBvsRPB}, which is a proof for the relation between PGB and the other bound derived in section~\ref{sec:RPB}, as a special case.

% \paragraph*{(GB with linear constraint and PGB)}

From \figurename~\ref{fig:project}, 
we see that the half space $\langle - \bm{Q}_-^{\rm GB}, \bm{X}\rangle \ge 0$, where $\bm{Q}_-^{\rm GB} = \bm{Q}^{\rm GB} - \bm{Q}^{\rm GB}_+$,  can be used as a linear relaxation of the semi-definite constraint 
for the linear constraint rule in section~\ref{sec:sphere+Linear}. 
Interestingly, GB with this linear constraint is tighter than PGB. 
This is proved in Appendix~\ref{app:PGBproof}, which is the proof of PGB. 

% $ \langle - \bm{Q}_-^{\rm GB}, \bm{X}\rangle \ge 0 $, 
%it gets tighter than PGB (\figurename~\ref{fig:project}). 
%This is proved in Appendix \ref{app:PGBproof} which is proof of PGB 
%by considering $\langle -\bm{Q}_-^{\rm GB}, \bm{X}\rangle \ge 0$ instead of $\bm{X}\succeq\bm{O}$. 

\subsubsection{Duality Gap Bound (DGB)}

In this section, we describe \emph{Duality Gap Bound} (DGB) 
in which the radius is represented by the duality gap:
% for LMML, which was previously used for the SVM screening \cite{shibagaki2016simultaneous}. 
\begin{theorem}[DGB]
	\label{thm:DGB}
	Let $\bm{M}$ be a feasible solution of the primal problem, and $\bm{\alpha}$ and $\bm{\Gamma}$ be feasible solutions of the dual problem, 
	then the optimal solution of the primal problem $\bm{M}^\star$ exists in the following hypersphere:
	\[
	\left\|\bm{M}^\star-\bm{M}\right\|_F^2\le {2(P_{\lambda}(\bm{M})-D_{\lambda}(\bm{\alpha},\bm{\Gamma}))}/{\lambda}.
	\]
\end{theorem}\noindent
\begin{proof}
 In general,  a function $f(\bm{x})$ is $m$-strongly convex function if $f(\bm{x})-\frac{m}{2}\left\|\bm{x}\right\|_2^2$ is a convex.
 % The objective function $P_\lambda(\bm{M})$ is a $\lambda$-strongly convex function, 
 % for which we employ the following definition:
 % 	\begin{definition}
 % 		\textbf{($m$-strongly convex function).}
 % 		\label{def:strong}
 % 		When $f(\bm{x})-\frac{m}{2}\left\|\bm{x}\right\|_2^2$ is a convex function, 
 % 		$f(\bm{x})$ is $m$-strongly convex function. 
 % 	\end{definition}\noindent
%	From the strongly convexity \cite{boyd2004convex}, we obtain
Since the objective function $P_\lambda(\bm{M})$ is a $\lambda$-strongly convex function, we obtain
	\[
		P_{\lambda}(\bm{M})\ge P_{\lambda}(\bm{M}^\star)+\langle \nabla P_{\lambda}(\bm{M}^\star), \bm{M}-\bm{M}^\star\rangle+\frac{\lambda}{2}\left\|\bm{M}-\bm{M}^\star\right\|_F^2.
	\]
	From the optimal condition \eqref{eq:primalopt}, the second term on the right hand side is greater than or equal to $0$, and from 
	weak duality, $P_{\lambda}(\bm{M}^\star)\ge D_{\lambda}(\bm{\alpha},\bm{\Gamma})$. 
	Therefore, we obtain \textbf{Theorem} \ref{thm:DGB}.
\end{proof} \noindent
Since the radius is proportional to the square root of the duality gap, 
DGB obviously converges to $0$ at the optimal solution (\figurename~\ref{fig:gap}). 
% \begin{figure}[tb]
% 	\centering
% 	\includegraphics[width=0.8\linewidth]{figure/gap.pdf}
% 	\caption{
% 	Duality Gap Bound (DGB). 
% 	The center is the reference solution, and the radius is determined from the duality gap.
% 	}
% 	\label{fig:gap}
% \end{figure}
For DGB, unlike the previous bounds, a dual feasible solution is necessary. 
This means that when a primal based optimization algorithm is employed, we need to create a dual feasible solution from a primal feasible solution.
A simple way to create a dual feasible solution is to substitute the current $\bm{M}$ into $\bm{M}^\star$ of \eqref{eq:alpha-star}.
% For the smoothed hinge loss, a simple way to create a dual solution is to use \eqref{eq:kkt1-2}.
% For the hinge loss, although a similar transformation is not available, any dual feasible solution can be substituted into DGB.
%
On the other hand, when a dual based optimization algorithm is employed, a primal feasible solution can be created by \eqref{eq:Mlambda}. % for the both cases.

For DGB, we further show that if the primal and dual reference solutions satisfy \eqref{eq:Mlambda}, the radius can be $\sqrt{2}$ times smaller.
We extend a dual based screening of SVM \cite{zimmert2015safe} for RTLM. 

\begin{theorem}[CDGB]
	\label{thm:CDGB}
	Let $\bm{\alpha}$ and $\bm{\Gamma}$ be the feasible solutions of the dual problem, then the optimal solution of the primal problem $\bm{M}^\star$ exists in the following hypersphere:
	\[
	\left\|\bm{M}^\star-\bm{M}_{\lambda}(\bm{\alpha},\bm{\Gamma})\right\|_F^2\le {G_{D_{\lambda}}(\bm{\alpha},\bm{\Gamma})}/{\lambda}.
	\]
\end{theorem}\noindent
\begin{proof}
 Let
 % \[
 $G_{D_{\lambda}}(\bm{\alpha},\bm{\Gamma})
 \coloneqq P_{\lambda}(\bm{M}_{\lambda}(\bm{\alpha},\bm{\Gamma})) - D_{\lambda}(\bm{\alpha},\bm{\Gamma})$
  % \]
	be the duality gap as a function of the dual feasible solutions $\bm{\alpha}$ and $\bm{\Gamma}$.
	On the other hand, the following equation is the duality gap as a function of the primal feasible solution $\bm{M}$ 
	in which the dual solutions are optimized:
	\begin{align*}
	 G_{P_{\lambda}}(\bm{M})& \coloneqq 
	 \hspace{-.8em}
	 \min_{
	 \scriptsize
	 \begin{array}{c}
	  \bm{0}\le\bm{\alpha}\le\bm{1}, \\
	  \bm{\Gamma}\succeq \bm{O},\\
	  \bm{M}_{\lambda}(\bm{\alpha},\bm{\Gamma})=\bm{M}
	 \end{array}
	 }
	 \hspace{-.8em}
	 G_{D_{\lambda}}(\bm{\alpha},\bm{\Gamma})
	 %		\\
	 &= P_{\lambda}(\bm{M})-
	 \hspace{-.8em}
	 \max_{
	 \scriptsize
	 \begin{array}{c}
	  \bm{0}\le\bm{\alpha}\le\bm{1}, \\
	  \bm{\Gamma}\succeq \bm{O}, \\
	  \bm{M}_{\lambda}(\bm{\alpha},\bm{\Gamma})=\bm{M}
	 \end{array}
	 }
	 \hspace{-.8em}
	 D_{\lambda}(\bm{\alpha},\bm{\Gamma}).
	\end{align*}
 From the definition, we obtain
 \begin{equation}
  \label{eq:G_D-geq-G_P}
		G_{D_{\lambda}}(\bm{\alpha},\bm{\Gamma}) \geq G_{P_{\lambda}}(\bm{M}_{\lambda}(\bm{\alpha},\bm{\Gamma})).
 \end{equation}
	% For $G_{P_{\lambda}}$, the following \textbf{Theorem} \ref{thm:GP} holds: 
	% \begin{theorem}
	% 	\textbf{(Strong convexity of $G_{P_{\lambda}}$).}
	% 	\label{thm:GP}
	% 	$G_{P_{\lambda}}$ is a $2\lambda$-strongly convex function. 
	% \end{theorem}\noindent
	From the strong convexity
 % (definition is show as \textbf{Definition}~\ref{def:strong} in Appendix~\ref{app:convexityGP}) 
 of $G_{P_{\lambda}}$ (Appendix~\ref{app:convexityGP}), we obtain
	\begin{equation}
		\label{eq:strong-convex-G_P} 
		G_{P_{\lambda}}(\bm{M}_{\lambda}(\bm{\alpha},\bm{\Gamma}))\ge 
		G_{P_{\lambda}}(\bm{M}^\star)
		+\langle \nabla G_{P_{\lambda}}(\bm{M}^\star), \bm{M}_{\lambda}(\bm{\alpha},\bm{\Gamma})-\bm{M}^\star\rangle
		+\lambda\left\|\bm{M}_{\lambda}(\bm{\alpha},\bm{\Gamma})-\bm{M}^\star\right\|_F^2.
	\end{equation}
	Considering the optimality of $G_{P_{\lambda}}(\bm{M}^\star)$ and combining \eqref{eq:G_D-geq-G_P} and \eqref{eq:strong-convex-G_P}, 
	\textbf{Theorem} \ref{thm:CDGB} can be derived. 
 See Appendix~\ref{app:CDGB} for the proof.
% See Appendix~\ref{app:convexityGP} and \ref{app:CDGB} for the proof.
\end{proof}\noindent
We call this bound \emph{Constrained Duality Gap Bound} (CDGB). 
Since CDGB also has a radius proportional to the square root of the duality gap, the radius converges to $0$ at the optimal solution. 
For primal based optimizers, additional calculation is necessary for $P_{\lambda}(\bm{M}_{\lambda}(\bm{\alpha},\bm{\Gamma}))$, 
while dual based optimizers calculates this term in the optimization process.

\subsubsection{Regularization Path Bound (RPB)}
\label{sec:RPB}
In \cite{wang2014scaling}, a hypersphere specific for \emph{regularization path} is proposed, in which the optimization problem should be solved for a sequence of $\lambda$s. 
Suppose that the optimization for $\lambda_0$ is already finished and the optimization for $\lambda_1$ is necessary to solve. 
Then, the same approach as \cite{wang2014scaling} is applicable to our RTLM, which derives a bound depending on the optimal solution for $\lambda_0$ as a reference solution:
\begin{theorem}[RPB]
	\label{thm:RPB}
	Let $\bm{M}_0^\star$ be the optimal solution for $\lambda_0$, the optimal solution $\bm{M}_1^\star$ for $\lambda_1$ exists in the following hypersphere: 
	\[
	\left\|\bm{M}_1^\star-\frac{\lambda_0+\lambda_1}{2\lambda_1}\bm{M}_0^\star\right\|_F^2
 \le\left(\frac{\lambda_0-\lambda_1}{2\lambda_1}\left\|\bm{M}_0^\star\right\|_F\right)^2.
	\]
\end{theorem}\noindent
\begin{proof}
	Let $\bm{\alpha}_i^\star$ and $\bm{\Gamma}_i^\star$ be the optimal dual solutions for $\lambda_i (i \in \{0,1\})$.
 From the optimality condition of the convex optimization problem \cite{bertsekas1999nonlinear}, 
 % Applying \textbf{Theorem}~\ref{thm:cvxopt}
%	\footnote{It is 
which is also called \emph{variational inequality} \cite{wang2014scaling},
% } 
to \eqref{eq:Dual1}, 
	we obtain the following two inequalities 
	\begin{gather*}
		\nabla_{\bm{\alpha}} D_{\lambda_0}(\bm{\alpha}_0^\star,\bm{\Gamma}_0^\star)^\top (\bm{\alpha}_1^\star-\bm{\alpha}_0^\star)
		+\langle \nabla_{\bm{\Gamma}} D_{\lambda_0}(\bm{\alpha}_0^\star,\bm{\Gamma}_0^\star), \bm{\Gamma}_1^\star-\bm{\Gamma}_0^\star \rangle
		\le0, \\
		\nabla_{\bm{\alpha}} D_{\lambda_1}(\bm{\alpha}_1^\star,\bm{\Gamma}_1^\star)^\top (\bm{\alpha}_0^\star-\bm{\alpha}_1^\star)
		+\langle \nabla_{\bm{\Gamma}} D_{\lambda_1}(\bm{\alpha}_1^\star,\bm{\Gamma}_1^\star), \bm{\Gamma}_0^\star-\bm{\Gamma}_1^\star \rangle
		\le0.
	\end{gather*}
 Note that dual variables should be in the feasible region $\bm{0}\le\bm{\alpha}\le\bm{1}, \bm{\Gamma}\succeq\bm{O}$.
 % Here, the feasible region $\mathcal{F}$ in \textbf{Theorem}~\ref{thm:cvxopt} is $\bm{0}\le\bm{\alpha}\le\bm{1}, \bm{\Gamma}\succeq\bm{O}$. 
 Combining these two inequalities, we obtain the \textbf{Theorem}~\ref{thm:RPB}.
 See Appendix~\ref{app:RPB} for the proof.
\end{proof}\noindent
We call this bound \emph{Regularization Path Bound} (RPB).

RPB requires the theoretically optimal solution $\bm{M}_0^\star$, which is numerically impossible.
Furthermore, since the reference solution is fixed on $\bm{M}_0^\star$, RPB can be performed only once for a specific pair of $\lambda_0$ and $\lambda_1$ even if the optimal $\bm{M}_0^\star$ is available. 
The other bounds can be performed multiple times during the optimization by regarding the current approximate solution as a reference solution. 
On the other hand, RPB provides interesting insights about relations with PGB and DGB.
%
% Although RPB requires the theoretically optimal solution $\bm{M}_0^\star$, which is numerically impossible, RPB provides an interesting insight about relations between PGB and DGB. 
The following theorem describes the relation between PGB and RPB:
\begin{theorem}[Relationship Between PGB and RPB]
 \label{thm:PGBvsRPB}
 Suppose that the optimal solution $\bm{M}_0^\star$ for $\lambda_0$ is substituted into  
 the reference solution $\bm{M}$ of PGB.
% Assuming that the reference solution $\bm{M}$ in PGB is the optimal solution $\bm{M}_0^\star$ for $\lambda_0$, 
Then, there exist a subgradient $\nabla P_{\lambda_1}(\bm{M}_0^\star)$ 
	by which PGB and RPB provides the same center and the radius for $\bm{M}_1^\star$.
\end{theorem}\noindent
See Appendix~\ref{app:PGBvsRPB} for the proof.
The following theorem describes the relation between DGB and RPB:
\begin{theorem}[Relationship Between DGB and RPB]
 \label{thm:DGBvsRPB}
Suppose that the optimal solutions $\bm{M}_0^\star, \bm{\alpha}_0^\star, \bm{\Gamma}_0^\star$ for $\lambda_0$ are substituted into the reference solutions $\bm{M}, \bm{\alpha}$ and $\bm{\Gamma}$ of DGB. 
Then, the radius of DGB and RPB for $\lambda_1$ has a relation $r_{\mathrm{DGB}}=2\,r_{\mathrm{RPB}}$, and the hypersphere of RPB is included in the hypersphere of DGB. 
\end{theorem}\noindent
See Appendix~\ref{app:DGBvsRPB} for the proof.
\figurename~\ref{fig:DGBvsRPB} illustrates the relation between DGB and RPB 
which shows theoretical advantage of RPB for the regularization path setting.
% \begin{figure}[tb]
% 	\centering
% 	\includegraphics[width=0.8\linewidth]{figure/DGBvsRPB.pdf}
% 	\caption{Relation between DGB and RPB (\textbf{Theorem} \ref{thm:DGBvsRPB}). }
% 	\label{fig:DGBvsRPB}
% \end{figure}
For practical use of RPB, we modify RPB in such a way that the approximate solution can be used as a reference solution.
Assuming that $\bm{M}_0$ satisfy
\[
\|\bm{M}_0^\star-\bm{M}_0\|_F\le\epsilon,
\]
where $\epsilon \ge 0$ is a constant.
% \begin{figure}[tb]
% 	\centering
% 	\includegraphics[width=0.7\linewidth]{figure/RRPB.pdf}
% 	\caption{
% 	Relaxed Regularization Path Bound (RRPB).
% 	RRPB ``blurs'' the center of RPB based on the current approximate solution.
% 	% Enlarge the radius to cover the blur of the optimal solution $\bm{M}_0^\star$.
% 	}
% 	\label{fig:RRPB}
% \end{figure}
Given $\bm{M}_0$ which satisfy the above condition, we obtain \emph{Relaxed Regularization Path Bound} (RRPB):
\begin{theorem}[RRPB]
	\label{thm:RRPB}
	Let $\bm{M}_0$ be an approximate solution for $\lambda_0$ which satisfies $\|\bm{M}_0^\star-\bm{M}_0\|_F\le\epsilon$. 
	The optimal solution $\bm{M}_1^\star$ for $\lambda_1$ exists in the following hypersphere:
\begin{align*} 
	\left\|\bm{M}_1^\star-\frac{\lambda_0+\lambda_1}{2\lambda_1}\bm{M}_0\right\|_F^2
	\le\left(
	\frac{|\lambda_0-\lambda_1|}{2\lambda_1}\left\|\bm{M}_0\right\|_F
	+\frac{|\lambda_0-\lambda_1|+\lambda_0+\lambda_1}{2\lambda_1}\epsilon
	\right)^2.
\end{align*}
\end{theorem}\noindent
See Appendix~\ref{app:RRPB} for the proof.
An intuition behind RRPB is shown in \figurename~\ref{fig:RRPB}, in which the approximation error for the center of RPB is depicted. 
In the theorem, RRPB also considers the error in the radius though it is not illustrated in the figure for simplicity.
To the best of our knowledge, this approach has not been introduced in other existing screening studies. 
% Then, since it is known that the center of the RPB $\frac{\lambda_0+\lambda_1}{2\lambda_1}\bm{M}_0^\star$ is located 
% within $\frac{\lambda_0+\lambda_1}{2\lambda_1}\epsilon$ 
% from the point using the approximate solution $\frac{\lambda_0+\lambda_1}{2\lambda_1}\bm{M}_0$, 
% the radius is enlarged so as to cover the uncertainty (blur) of this optimal solution (\figurename~\ref{fig:RRPB}). 

For example, $\epsilon$ can be set from \textbf{Theorem} \ref{thm:DGB} (DGB) as follows: 
\[
\epsilon=\sqrt{{2(P_{\lambda_0}(\bm{M}_0)-D_{\lambda_0}(\bm{\alpha}_0,\bm{\Gamma}_0))}/{\lambda_0}}.
\]
When the optimization for $\lambda_0$ terminates, the solution $\bm{M}_0$ should be accurate in terms of some stopping criterion such as the duality gap.
Then, $\epsilon$ is expected to be quite small, and RRPB can provide a tight bound for $\lambda_1$, which is close to the ideal (but not computable) RPB.
% Then, if $\lambda_0 \ne \lambda_1$, $\epsilon$ is sufficiently small when the approximate solution $\bm{M}_0$ for $\lambda_0$ is good enough. 
%
As a special case, by setting $\lambda_1 = \lambda_0$, RRPB is applicable to perform screening of $\lambda_1$ using any approximate solution having $\| \bm{M}_1^\star - \bm{M} \|_F \leq \epsilon$, and then RRPB is equivalent to DGB.
%Otherwise, i.e. $\lambda_0=\lambda_1$, RRPB is reduced to DGB. 
%
% Unlike RPB, since the reference solution of RRPB is a general feasible solution $\bm{M}_0$, for a specific pair of $\lambda_0$ and $\lambda_1$, we can perform the screening multiple times using different reference solutions.

% --------------------------------------------------
\subsection{Computational Cost}

Considering computational cost of the screening procedure, the rule evaluation ({\bf Step2}) described in section~\ref{sec:screening} is often dominant, because the rule needs to be evaluated for each one of triplets.
On the other hand, the sphere, constructed in {\bf Step1}, can be fixed during the screening procedure as long as the reference solution is fixed.

Sphere Rule (section \ref{sec:noSD}) needs $O(d^2)$ computations for the inner product $\langle \bm{H}_{ijl}, \bm{Q}\rangle$, but we can reuse this term from objective function $P_{\lambda}(\bm{M})$ calculation in the case of DGB, RPB and RRPB.
%
% Sphere Rule (section \ref{sec:noSD}) takes the calculation cost of the inner product $O(d^2)$, 
% but because $\bm{Q}$ is proportional to the current solution $\bm{M}$ for the case of DGB, RPB or RRPB, 
% we can reuse the inner product calculation used for calculating the objective function $P_{\lambda}(\bm{M})$. 
%
The computational cost of Sphere Rule with Semi-definite Constraint (section \ref{sec:sphere+SD}) is that of SDLS algorithm. 
% The calculation cost of Sphere Rule with Semi-definite Constraint (section \ref{sec:sphere+SD}) is that of SDLS algorithm. 
%
SDLS algorithm needs $O(d^3)$ because of the eigenvalue decomposition at every iteration, which may cause large computational cost.
% The computational complexity of the SDLS algorithm is $O(kd^3)$ as eigenvalue decomposition becomes dominant. 
% $k$ is the number of loops.
% Sphere Rule with Linear Constraint (section \ref{sec:sphere+Linear}) cal be evaluated $O(d^2)$ by using 
%
The calculation cost of Sphere Rule with Linear Constraint (section \ref{sec:sphere+Linear}) takes $O(d^2)$.

\section{Range Based Extension of Triplet Screening}
\label{sec:screeningRange}
The screening rules shown in section~\ref{sec:screening} provides the conditions for the problem of a fixed $\lambda$. 
In this section, by considering $\lambda$ as a variable, we derive a range of $\lambda$ in which the screening rule is guaranteed to be satisfied.
This is particularly useful for the regularization path calculation for which we need to optimize the metric for a sequence of $\lambda$s.
If a screening rule is satisfied for a triplet $(i,j,l)$ in a range $(\lambda_a, \lambda_b)$, 
we can fix the triplet $(i,j,l)$ in $\hat{\cL}$ or $\hat{\cR}$ as long as $\lambda$ is in $(\lambda_a, \lambda_b)$, 
without computing screening rules.

Let $\bm{Q}=\bm{A}+\bm{B}\frac{1}{\lambda}$ be a general form of hypersphere for some constant matrices 
$\bm{A} \in \RR^{d \times d}$ and $\bm{B} \in \RR^{d \times d}$, and $r^2=a+b\frac{1}{\lambda}+c\frac{1}{\lambda^2}$ 
be a general form of the radius for some constants $a \in \RR$, $b \in \RR$ and $c \in \RR$.
GB, DGB, RPB and RRPB can be in this form (see Appendix~\ref{app:general-sphere} for detail).
The sphere rule for $\cR^\star$ \eqref{eq:sphere-ruleR2} is 
equivalent to the intersection of the following two inequalities because of $r\ge0$ and $\|\bm{H}_{ijl}\|_F\ge0$:
\[
\langle \bm{H}_{ijl}, \bm{Q} \rangle-1>0, ~
(\langle \bm{H}_{ijl}, \bm{Q} \rangle-1)^2>r^2\|\bm{H}_{ijl}\|_F^2.
\]
Since 
$
\langle \bm{H}_{ijl}, \bm{Q} \rangle
=\langle \bm{H}_{ijl}, \bm{A} \rangle+\langle \bm{H}_{ijl}, \bm{B} \rangle\frac{1}{\lambda},
$
The first and second inequalities can be transformed into linear and quadratic functions of $\lambda$ respectively, 
for which it is easy to find the range of $\lambda$ satisfying these two inequalities.
The following theorem shows the range for the case of RRPB given a reference solution $\bm{M}_0$ which is an approximate solution for $\lambda_0$:
\begin{theorem}[Range Based Extension of RRPB]
 \label{thm:RRPBrange}
 Assuming \\
 $\langle \bm{H}_{ijl}, \bm{M}_0 \rangle - 2 + \| \bm{H}_{ijl} \|_F \| \bm{M}_{0} \|_F > 0$
 and
 $\| \bm{M}_0^\star - \bm{M}_0 \|_F \le \epsilon$, 
 a triplet $(i,j,l)$ is guaranteed to be in $\cR^\star$ for the following range of $\lambda$:
 \[
 \lambda \in \left( \lambda_a, \lambda_b \right),
 \]
 where
 \begin{align*}
  % Lower
  \lambda_a &=
  \frac
  {
  \lambda_0
  \left(
  \| \bm{M}_0 \|_F \| \bm{H}_{ijl} \|_F
  -
  \langle \bm{H}_{ijl}, \bm{M}_0 \rangle
  +
  2 \epsilon \| \bm{H}_{ijl} \|_F
  \right)
  }
  {
  \langle \bm{H}_{ijl}, \bm{M}_0 \rangle
  -
  2
  +
  \| \bm{H}_{ijl} \|_F \| \bm{M}_{0} \|_F
  }, 
  \\
  % Upper
  \lambda_b &= 
  \frac
  {
  \lambda_0
  \left(
  \| \bm{M}_0 \|_F \| \bm{H}_{ijl} \|_F
  +
  \langle \bm{H}_{ijl}, \bm{M}_0 \rangle
  \right)
  }
  {
  \| \bm{H}_{ijl} \|_F \| \bm{M}_{0} \|_F
  -
  \langle \bm{H}_{ijl}, \bm{M}_0 \rangle
  + 
  2  
  +
  2 \epsilon \| \bm{H}_{ijl} \|_F
  }.
 \end{align*}
\end{theorem} \noindent
See Appendix~\ref{app:RRPBrange} for the proof.

\section{Experiment}\label{sec:experiment}

We evaluate performance of safe triplet screening using the benchmark datasets shown in \tablename~\ref{tbl:dataset}, which are from LIBSVM \cite{CC01a} and Keras Dataset \cite{chollet2015keras}. 
To create a set of triplets, we follow the approach by \cite{shen2014efficient}, in which $k$ neighborhoods in the same class $\bm{x}_j$ and $k$ neighborhoods in different class $\bm{x}_l$ are sampled for each $\bm{x}_i$.
%
% According to the definition, triplets is too many, so consider the problem limiting the triplets to be used as follows. 
% For each sample $\bm{x}_i$, triplet set $\mathcal{T}$ is formed with $k$ neighborhood of the same class $\bm{x}_j$ and $k$ neighborhood of different class $\bm{x}_l$.
% \footnote{This is the same way of making triplet as \refname \cite{shen2014efficient}.}
% The data set to be used is as shown in \tablename~\ref{tbl:dataset} from LIBSVM\cite{CC01a} and Keras Dataset \footnote{\href{https://keras.io/datasets/}{https://keras.io/datasets/}}. 
% Data with a large number of dimensions are compressed using Principal Component Analysis (PCA) or AutoEncoder. 
\begin{table}[t]
 \caption{
 Summary of datasets.
 $*1 : $The dimension was reduced by AutoEncoder.
 $*2 : $The dimension was reduced by PCA.
 \#triplet and $\lambda_{\min}$ are the average value for sub-sampled random trials.
 }
 \label{tbl:dataset}
 \centering
 {\footnotesize\tabcolsep=1mm
 \begin{tabular}{l||r|r|r|r|r|r|r}
                   & \#dimension & \#sample & \#classes & $k$ & \#triplet & $\lambda_{\max}$ & $\lambda_{\min}$ \\ \hline
  segment                 & 19  & 2310  & 7  & 20 & 832000  & 2.5e$+$6 & 4.2e$+$0 \\
  % satimage                & 36  & 4435  & 6  & 15 & 898200  & 1.0e$+$7 & 8.8e$+$0 \\
  phishing                & 68  & 11055 & 2  & 7  & 487550  & 5.0e$+$3 & 2.0e$-$1 \\
  SensIT Vehicle          & 100 & 78823 & 3  & 3  & 638469  & 1.0e$+$4 & 2.9e$+$0 \\
  a9a$^{*1}$              & 16  & 32561 & 2  & 5  & 732625  & 1.2e$+$5 & 3.1e$+$2 \\
  mnist$^{*1}$            & 32  & 60000 & 10 & 5  & 1350025 & 7.0e$+$3 & 9.6e$-$1 \\
  cifar10$^{*1}$          & 200 & 50000 & 10 & 2  & 180004  & 2.0e$+$3 & 3.3e$+$1 \\
  rcv1.multiclass$^{*2}$  & 200 & 15564 & 53 & 3  & 126018  & 3.0e$+$2 & 6.0e$-$4
 \end{tabular} 
 }
\end{table}
We employed the regularization path setting in which RTLM is optimized for a sequence of $\lambda_0, \lambda_1, \ldots, \lambda_T$.
The initial $\lambda_0=\lambda_{\max}$ was set by a sufficiently large value in which $\mathcal{R}^\star$ starts increasing from the empty set.
To generate the next value of $\lambda$, we used $\lambda_{t}=0.9\lambda_{t-1}$, and the path terminated when the following condition is satisfied:
% $\lambda_{t}=0.9\lambda_{t-1}$ from $\lambda_0=\lambda_{\max}$ 
%
%
% Consider a regularization path according to $\lambda_{t}=0.9\lambda_{t-1}$ from $\lambda_0=\lambda_{\max}$ 
% which is sufficiently large that $\mathcal{R}^\star$ starts to increase. 
% When the loss term no longer decreases sufficiently, the regularization path is terminated. 
% The conditions are as follows.
\[
\frac{\mathrm{loss}(\lambda_{t-1})-\mathrm{loss}(\lambda_{t})}{\mathrm{loss}(\lambda_{t-1})}
\times 
\frac{\lambda_{t-1}}{\lambda_{t-1}-\lambda_{t}}<0.01,
\]
where $\mathrm{loss}(\lambda_{t})$ is the loss function value at $\lambda_t$.
We randomly selected 90\% of the instances of each dataset 5 times, and the average is shown as the experimental result.
As a base optimizer, we employed the projected gradient descent of the primal problem, and the iteration terminated when the duality gap becomes less than $10^{-6}$.
For the loss function $\ell$, we used the smoothed hinge loss of $\gamma=0.05$ (We also provides results for the hinge loss in Appendix~\ref{app:eval-hinge}).
We performed safe triplet screening every ten iterations of the gradient descent.
We refer to the first screening for a specific $\lambda_t$, in which the solution of previous $\lambda_{t-1}$ is used for the reference solution, as the \emph{regularization path screening}.
On the other hand, the screening performed during the optimization process (after regularization path screening) is called \emph{dynamic screening}.
For all experiments, we performed both of these screening procedures.
As a baseline, we call the RTLM optimization without screening \emph{naive optimization}.
When the regularization coefficient changes, $\bm{M}$ starts from the previous solution $\hat{\bm{M}}$ (warm start).
The step size of the gradient descent was determined by
% We randomly select 90\% of the samples of each dataset and experiment with regularization path.
% This trial is conducted 5 times, and the average is taken as the experiment result.
% We solve the primal problem of metric learning by the gradient method and regard it as convergence when the relative duality gap becomes less than $10^{-6}$.
% For the loss function $\ell$, use the smoothed hinge loss of $\gamma=0.05$. 
% Screening is done once every ten times updating of the primal variable $\bm{M}$. 
% When the regularization coefficient changes, $\bm{M}$ starts from the previous solution $\hat{\bm{M}}$(warm start).
% The update width is as follows with reference to \refname \cite{barzilai1988two}.
\[
\frac{1}{2}\left|
\frac{\langle \Delta \bm{M} , \Delta \bm{G}\rangle}{\langle \Delta \bm{G} , \Delta \bm{G}\rangle}
+\frac{\langle \Delta \bm{M} , \Delta \bm{M}\rangle}{\langle \Delta \bm{M} , \Delta \bm{G}\rangle}\right|,
\]
where $\Delta \bm{M}=\bm{M}_t-\bm{M}_{t-1}, \Delta \bm{G}=\nabla P_{\lambda}(\bm{M}_t)-\nabla P_{\lambda}(\bm{M}_{t-1})$ \cite{barzilai1988two}.
%
% For the sphere rule with semi-definite constraint \eqref{eq:P2}, the step size of SDLS dual ascent method was
% % When the semi-definite constraint is used, the update width of SDLS dual ascent method is decided as follows. 
% \[
% \alpha_{0} = \left\|\bm{H}_{ijl}\right\|_F^{-2},~
% \alpha_{t} = -\frac{y_{t} - y_{t-1}}{\nabla D_{\mathrm{SDLS}}(y_{t}) - \nabla D_{\mathrm{SDLS}}(y_{t-1})}.
% \]
In SDLS dual ascent, we used the conjugate gradient method \cite{yang1993conjugate} for finding the minimum eigenvalue.

\subsection{Comparing GB Based Rules}
% \subsection{Performance Comparison of Screening Rules}

\begin{figure}[tb]
 \centering
 \includegraphics[width=\linewidth]{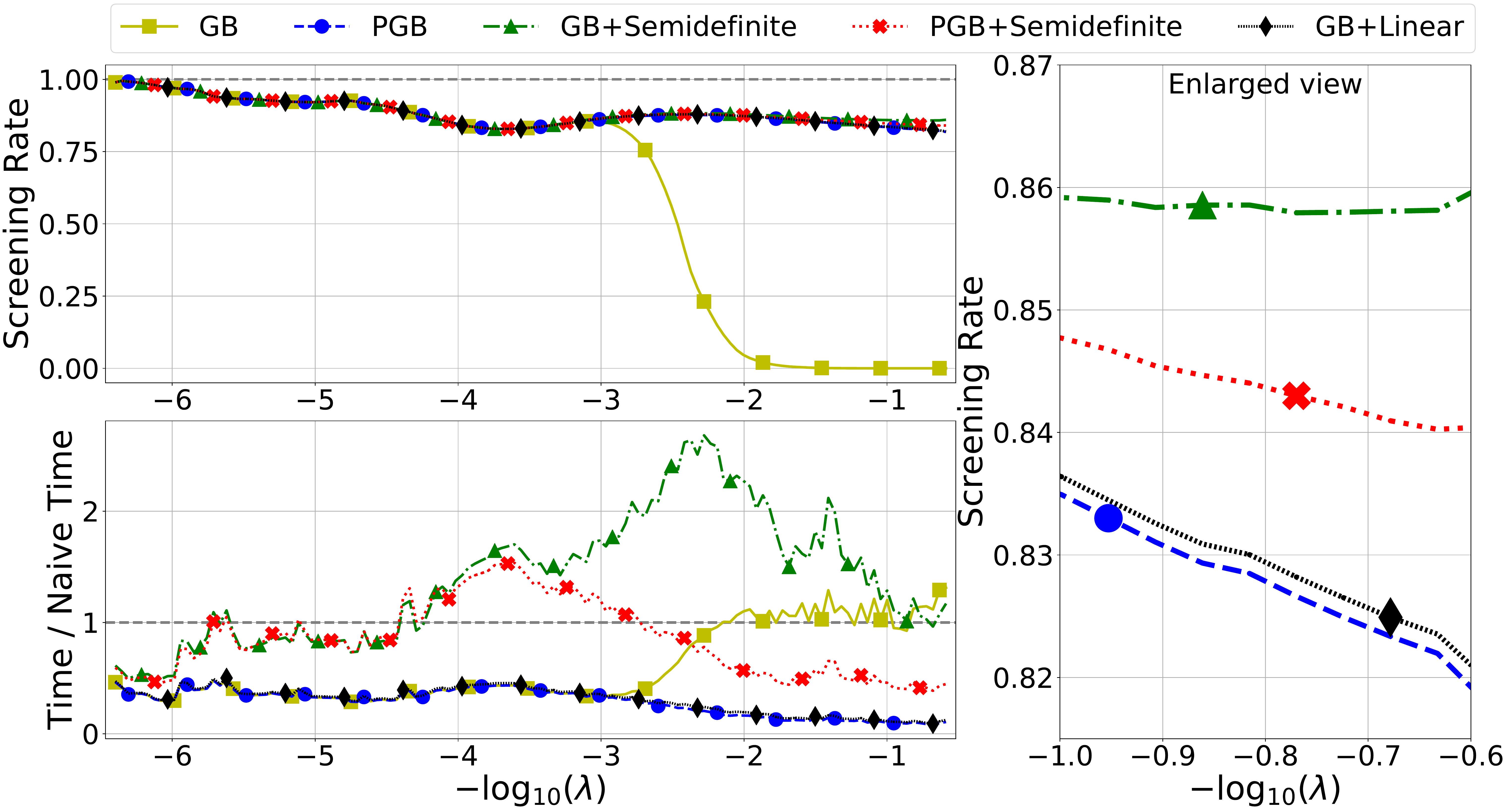}
 \caption{
 Screening rule comparison on segment dataset.
 % Screening Rule performance in segment. 
 %
 The top left plot shows performance of regularization path screening, and the bottom left plot shows the ratio of the CPU time compared with the naive optimization for each $\lambda$.
 %
 % The upper figure shows the first Screening Rate after $\lambda$ changes and the lower figure shows the time ratio with Naive without Screening. 
 %
 The right plot enlarges the upper left plot for a range $- \log_{10}(\lambda) \in [-1, -0.6]$. 
 }
 \label{fig:segmentComp}
\end{figure}

We first validate the screening performance (screening rate and CPU time) of each screening rule introduced in the section \ref{sec:screening}. 
%
% In this section, we check the screening performance (Screening Rate, Time) of each Screening Rule introduced in the section \ref{sec:screening}. 
We here use GB and PGB as spheres, and observe the effect of the semi-definite constraint in the rules.
%
% By considering semidefinite constraint, comparison is made using GB and PGB in order to see what kind of change is. 
%
As a representative result, comparison on the segment data is shown in \figurename~\ref{fig:segmentComp}. 
% The result of using segment data is shown in \figurename~\ref{fig:segmentComp}. 

First of all, we see that the rules except for GB keep the high screening rate for the entire regularization path shown as the top left plot.
Note that this rate is only for regularization path screening, meaning that dynamic screening can further increase screening rate during the optimization as we see next subsection.
The bottom left plot of the same figure shows PGB and GB+Linear are most efficient which achieved about $2$-$10$ times faster CPU time than the naive optimization. 
The screening rate of GB severely dropped on the later half of the regularization path.
As illustrated in \figurename~\ref{fig:project}, the center of GB can be outside of the semi-definite cone by which the sphere of GB contains a larger proportion of the region violating the constraint $\bm{M} \succeq \bm{O}$, compared with the spheres having their center inside the semi-definite cone.
This causes performance deterioration particularly for smaller $\lambda$, because the minimum of the loss term is usually in outside of the semi-definite cone. 
%
% Comparing Screening Rate, we can see that semidefinite constraint is becoming important in the latter half, since that of GB is 0 in the latter half. 

The screening rate of GB+Linear and GB+Semidefinite are slightly higher than that of PGB (the right plot), which can be seen from the geometrical relation of them illustrated in \figurename~\ref{fig:project}.
% The reason why Screening Rate of GB+Linear is higher than that of PGB is as shown in the section \ref{sec:sphere+Linear}.
%
GB+Semidefinite achieved the highest screening rate, but the eigenvalue decomposition is necessary to calculate repeatedly in SDLS, by which the CPU time increased in the later half of the path.
% (Note that the center of GB is not positive semi-definite)
%
% As can be seen from \figurename~\ref{fig:project}, GB+Semidefinite maintains the highest Screening Rate, but eigenvalue decomposition is needed many times, so the calculation time is increasing in the latter half. 
%
Although PGB+Semidefinite is also tighter than PGB, the CPU time increased around from $- \log_{10}(\lambda) \approx - 4$ to $-3$.
Since the center of PGB is positive semi-definite, only the minimum eigenvalue is required (see section~\ref{sec:sphere+SD}), but it still can increase the CPU time.
%
%Although PGB+Semidefinite becomes tighter than PGB, the computation time is also increasing in the second half as calculation of the minimum eigenvalue becomes necessary many times. 
%

Among screening methods compared here, our empirical analysis suggests that the sphere rule with PGB is most cost-effective, in which semi-definite constraint is implicitly incorporated at the projection process.
We did not observe that the other approach to considering the semi-definite (or relaxed linear) constraint in the rule substantially outperform PGB in terms of the CPU time despite their high screening rate. 
% What we can see from this result is that only Screening Rate of about PGB can be expected even if semidefinite constraint are taken into GB and PGB. 
% This fact became more prominent as the dimension increased and we confirmed that Screening Rate is almost the same as PGB. 
% \footnote{
We observed the same tendency for DGB. 
The screening rate did not largely change even if the semi-definite constraint is explicitly taken into account (see Appendix~\ref{app:compareDGB}).
% The same was true when considering semidefinite constraint on DGB. 
% Screening Rate seems to be almost unchanged 
% even if semidefinite constraint are taken into account for a hyperspheres whose center is semidefinite. 
% }
%
% On the other hand, since the computation time increases considerably due to eigenvalue decomposition and minimum eigenvalues, it can be said that the Screening Rule in Section \ref{sec:sphere+SD} is not practical. 

\subsection{Comparing Bounds}
\label{sec:exp-bounds}
% \subsection{Performance Comparison of Bounds}

\begin{figure}[tb]
 \centering
 \includegraphics[width=\linewidth]{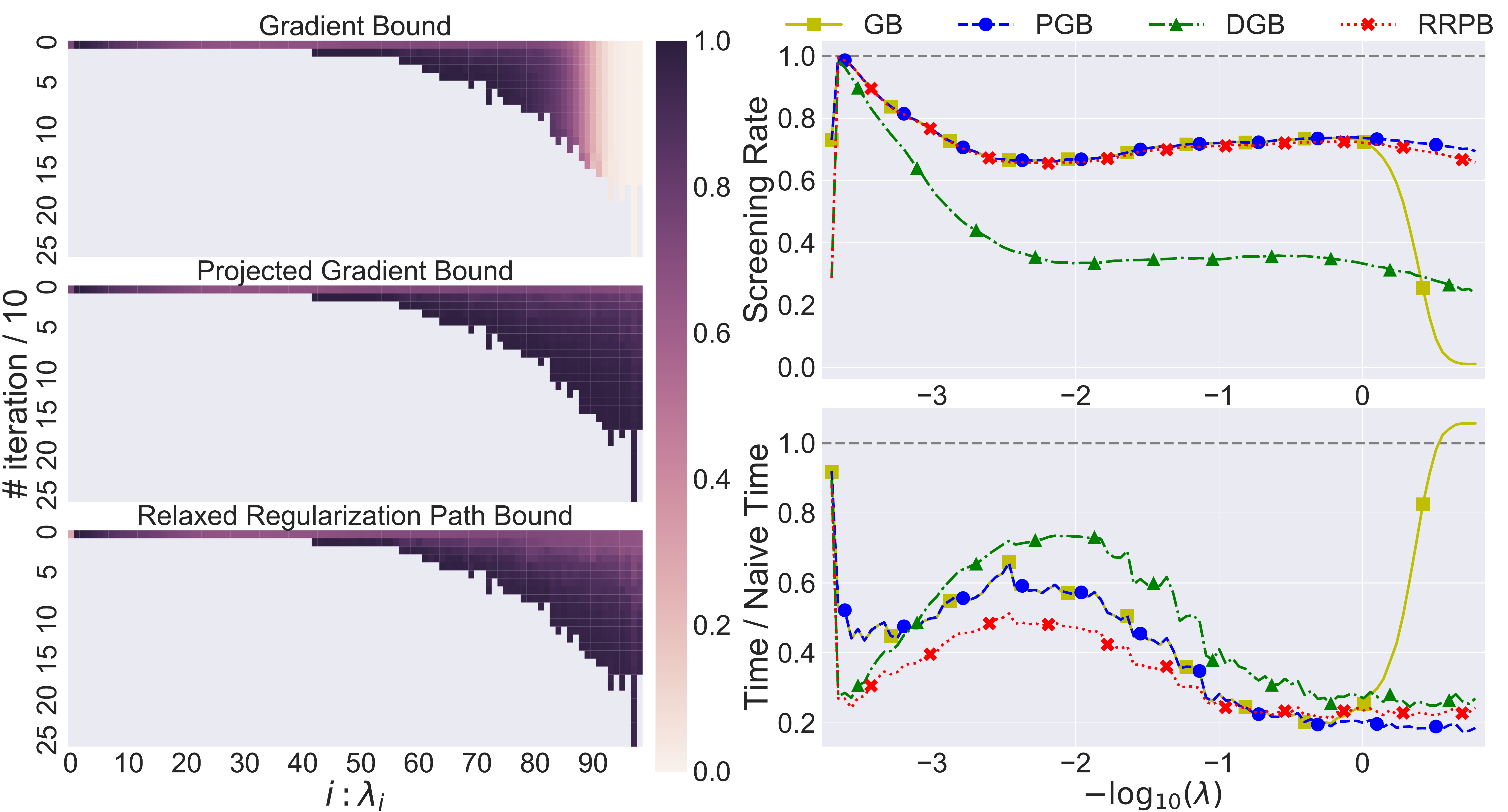}
 \caption{
 Comparison of sphere bounds on phishing dataset.
 The left heatmaps show the screening rate of dynamic screening.
 The vertical axis is the number of screening performed for $\lambda_i$ (once every ten iterations of gradient descent).
 The top right plot shows the rate of regularization path screening, and the bottom right plot shows the ratio of the CPU time compared with the naive optimization.
 }
 \label{fig:phishing}
\end{figure}

We next compare screening performance (screening rate and CPU time) of each bound introduced in the section \ref{sec:bound}.
% In this section, we check the screening performance (Screening Rate, Time) of each bound introduced in the section \ref{sec:bound}.
%
Based on the results in the previous section, we employed the sphere rule.
% Based on the results in the previous section, we perform a comparison using Screening Rule by only hypersphere. 
%
% The result of the satimage and phishing dataset are shown in \figurename~\ref{fig:satimage}, and \figurename~\ref{fig:phishing}, respectively. 
The result of the phishing dataset are shown in \figurename~\ref{fig:phishing}. 
Screening rate (the top right plot) of GB again dropped from the middle compared with the other spheres. 
Screening rate (the top right plot) of GB again dropped from the middle compared with the other spheres. 
The other spheres also have lower screening rates for  small $\lambda$s.
As we mention in section~\ref{sec:screeningRange}, the radiuses of GB, DGB, RPB and RRPB have the form $r^2=a+b\frac{1}{\lambda}+c\frac{1}{\lambda^2}$, meaning that if $\lambda \rightarrow 0$ then $r \rightarrow \infty$.
%
% \red{The radius of DGB is proportional to $1/\lambda$, and thus smaller $\lambda$ leads to a larger radius, and RRPB depends on DGB.}
%  
For PGB, although dependency on $\lambda$ can not be written explicitly, the same tendency was observed.
% Screening Rate of GB is 0 from the middle, but other bounds keep high Screening Rate in the second half.
%
We see that PGB and RRPB have similar results as suggested by \textbf{Theorem}~\ref{thm:PGBvsRPB}, and the screening rate of DGB is lower than RRPB as suggested by \textbf{Theorem}~\ref{thm:DGBvsRPB}.
%
% The reason why PGB and RRPB have similar results is shown in \textbf{Theorem} \ref{thm:PGBvsRPB}, and the reason why Screening Rate of DGB is lower than that of RRPB is as shown in \textbf{Theorem} \ref{thm:DGBvsRPB}.
%
Comparing PGB and RRPB, PGB achieved the higher screening rate, but RRPB shows the faster CPU time (the bottom right plots), because
% Comparing PGB and RRPB, Screening Rate is higher for PGB, but calculation time is less for RRPB.
%
PGB requires a matrix inner product calculation for each triplet.
% This is because PGB is accompanied by cost of inner product calculation for Screening. \red{(射影のコストはどうか？)}
%
We see that the bounds other than GB are more than two times faster than the naive calculation for most of $\lambda$s.
% Looking at the time ratio, bounds other than GB are almost less than half the time compared with Naive.

% \begin{figure}[tb]
%  \centering
%  \includegraphics[width=\linewidth]{experiment/satimage.pdf}
%  \caption{
%  %
%  Comparison of sphere bounds on satimage dataset.
%  % Bound performance in satimage. 
%  %
%  The left heatmaps show the screening rate of dynamic screening.
%  %
%  The vertical axis is the number of screening performed for $\lambda_i$ (once every ten iterations of gradient descent).
%  % The left figure represents the Screening Rate of Dynamic Screening. 
%  %
%  The top right plot shows the rate of regularization path screening, and the bottom right plot shows the ratio of the CPU time compared with the naive optimization.
%  % The upper right figure shows the first Screening Rate after $\lambda$ changes and the lower right figure shows the time ratio with Naive without Screening. 
%  }
%  \label{fig:satimage}
% \end{figure}

% The result using phishing data is shown in \figurename~\ref{fig:phishing}. 
%
Comparing the dynamic screening rate (the left three plots of 
% \figurename~\ref{fig:satimage} and 
\figurename~\ref{fig:phishing}) of PGB and RRPB, PGB has the higher screening rate.
% When comparing Dynamic Screening Rate of PGB and RRPB, PGB is higher.
%
For the regularization path screening (the top right), RRPB and PGB have similar screening rate, but for the dynamic screening, PGB has the higher rate. 
For the later half of the regularization path, the number of gradient descent iterations increases, by which the dynamic screening significantly effects on the CPU time, and PGB becomes faster despite its additional computation for the inner product. 
In Appendix~\ref{app:compBounds}, we show the CPU time for the entire path with some additional datasets.

\begin{figure}[tb]
	\centering
\includegraphics[clip,width=\linewidth]{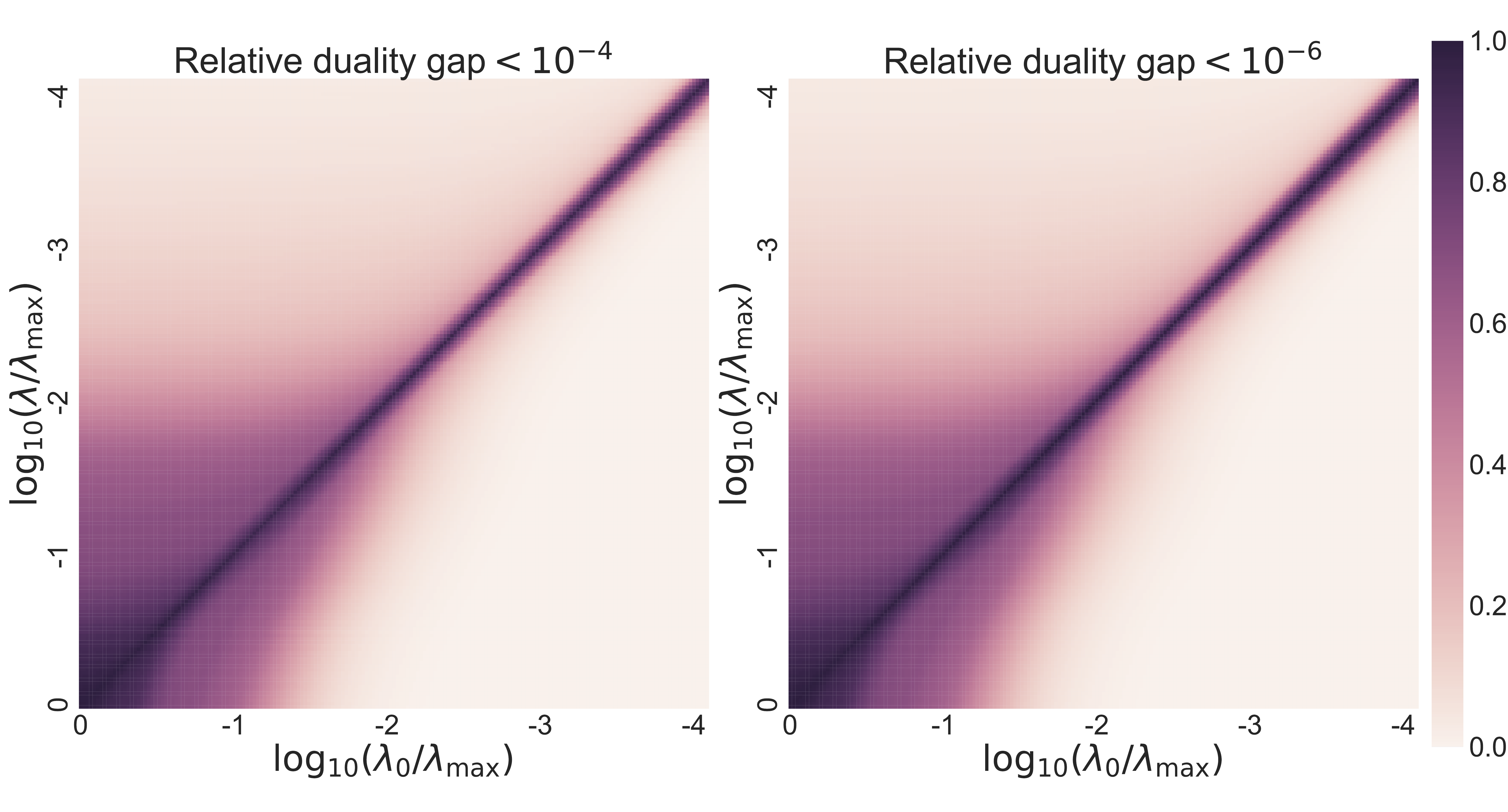}
	\caption{
	Screening rate of the range based screening on the segment dataset.
	The color indicates the screening rate for $\lambda$ in the vertical axis based on the reference solution at $\lambda_0$ in the horizontal axis. 
	The accuracy of the reference solution is $10^{-4}$ for the left plot and $10^{-6}$ for the right plot.
	} 
	\label{fig:rangeScreening}
\end{figure}

We further evaluate performance of the range based extension described in section~\ref{sec:screeningRange}.
\figurename~\ref{fig:rangeScreening} shows the rate of the range based screening for the segment dataset.
%
% satimage dataset.
%
% Since the range based condition is a stronger condition than the usual regularization path screening, the rate is always lower than the regularization path screening.
%
We see that a wide range of $\lambda$ can be screened particularly for small $\lambda$, and for large $\lambda$, although the range is smaller than the small $\lambda$ cases, high screening rate was observed for $\lambda$ close to $\lambda_0$.
A significant advantage of this approach is that, for triplets screened by the range, we do not need to evaluate screening rule anymore as long as $\lambda$ is in the range.
%  as we mention in section~\ref{sec:screeningRange}.

% --------------------------------------------------
% Rate of Range-based Screening
% --------------------------------------------------

\subsection{Practical Performance Evaluation}
\label{sec:pracEval}
% \subsection{Comparison with existing methods}

As a computationally more expensive setting, we consider investigating the regularization path in more detail by setting $\lambda_{t} = 0.99 \lambda_{t-1}$.
To evaluate practical performance, we combine our safe triplet screening with the well-known \emph{active set} heuristics.
% We do the same method as Weinberger et al's ActiveSet method with smoothed hinge loss. 
%
In the active set method, only a subset of triplets whose loss is greater than $0$ are treated as the active set.
% Like the Weinberger et al., triplets whose loss is not 0 are treated as ActiveSet, and ActiveSet is updated once every ten times.
%
The gradient is calculated by only using the active set, and the overall optimality is confirmed when the iteration converges.
% The gradient is calculated using only ActiveSet, and the overall optimality is confirmed at the time of convergence. 
%
We employed the active set update strategy shown by Weinberger et al. \cite{weinberger2009distance}, in which the active set is updated once every ten iterations.

\tablename~\ref{tbl:active} shows the CPU time comparison for the entire regularization path.
%
% The time of this ActiveSet method and that of the method combining Screening with ActiveSet method is compared. 
%
Based on the the results in the previous section, we employed RRPB and RRPB+PGB (evaluating rules based on both spheres) for the triplet screening. 
% Following the results in the previous section, we use RRPB, RRPB+PGB for Screening. 
%
Further, the range based screening described in section~\ref{sec:screeningRange} is also performed using RRPB, 
for which we evaluate the range at the beginning of the optimization for each $\lambda$.
% Furthermore, Screening by Screening possible range in Section \ref{sec:screeningRange} is also done by RRPB.
%
% Total time of regularization path computation is shown in \tablename~\ref{tbl:active}. 
%
We see that our safe triplet screening accelerates the optimization process about up to $10$ times from the simple active set method. 
The results for higher dimensional datasets with diagonal $\bm{M}$ are also shown in Appendix~\ref{app:highdimRslt}.

\begin{table}[tb]
 \caption{
 Total CPU time (sec) evaluation with the active set method.
 % Total time of regularization path. Comparison with ActiveSet.
 The results with $^\star$ indicates the fastest method.
 }
 \label{tbl:active}
 \centering
	{\footnotesize\tabcolsep=1mm
		% \begin{tabular}{c||r|r|r|r|r|r|r|r|r|r}
		% Method\textbackslash Dataset	
		% &iris	&wine	&segment	&satimage	&phishing	&SensIT		&a9a	&mnist	&cifar10&rcv1	\\ \hline
		% ActiveSet						
		% &51.5		&186.4		&539.9		&1171.2		&2010.1		&3876.6		&90.9		&457.0		&4366.6		&27464.6	\\
		% ActiveSet+RRPB
		% &$\pmb{9.6}$	&$\pmb{67.0}$	&125.5		&396.5		&948.4		&1961.7		&$\pmb{24.5}$	&$\pmb{293.0}$	&3331.8		&19974.7	\\
		% ActiveSet+RRPB+PGB
		% &10.1		&69.9		&$\pmb{124.3}$	&$\pmb{364.9}$	&$\pmb{917.8}$	&$\pmb{1860.8}$	&27.1		&334.1		&$\pmb{2965.4}$	&$\pmb{19212.9}$\\
		% \end{tabular}
		% \begin{tabular}{l||r|r|r|r|r|r}
		% Method\textbackslash Dataset	&phishing			&SensIT				&a9a			&mnist			&cifar10		&rcv				\\ \hline
		% ActiveSet						&7989.5				&16352.1			&758.7			&3788.1			&11085.7		&				-	\\
		% ActiveSet+RRPB					&$\pmb{2126.2}$		&3555.6				&$\pmb{70.1}$	&$\pmb{871.1}$	&1431.3			&43174.9			\\
		% ActiveSet+RRPB+PGB				&2133.2				&$\pmb{3046.9}$		&72.1			&897.9			&$\pmb{1279.7}$	&$\pmb{38231.1}$	\\
		% \end{tabular}
		\begin{tabular}{l||r|r|r|r|r|r}
		Method\textbackslash Dataset	&phishing			&SensIT				&a9a			&mnist			&cifar10		&rcv				\\ \hline
		ActiveSet						&7989.5				&16352.1			&758.7			&3788.1			&11085.7		&94996.3			\\
		ActiveSet+RRPB					&$^\star$2126.2		&3555.6				&$^\star$70.1	&$^\star$871.1	&1431.3			&43174.9			\\
		ActiveSet+RRPB+PGB				&2133.2				&$^\star$3046.9		&72.1			&897.9			&$^\star$1279.7	&$^\star$38231.1	\\
		\end{tabular}
	}
\end{table}

\section{Summary}\label{sec:summary}

We introduced \emph{safe triplet screening} for large margin metric learning.
The three screening rules and the six sphere bounds were derived, and their relation was analyzed.
We further proposed range based extension for the regularization path calculation.
Our screening technique for metric learning is particularly significant compared with other screening studies due to massiveness of triplets and the semi-definite constraint. 
Our numerical experiments verified effectiveness of safe triplet screening using several benchmark datasets.

This work was financially supported by grants from the Japanese Ministry of Education, Culture, Sports, Science and Technology awarded to I.T. (16H06538, 17H00758) and M.K. (16H06538, 17H04694); from Japan Science and Technology Agency (JST) CREST awarded to I.T. (JPMJCR1302, JPMJCR1502) and PRESTO awarded to M.K. (JPMJPR15N2); from the ``Materials research by Information Integration'' Initiative (MI2I) project of the Support Program for Starting Up Innovation Hub from JST awarded to I.T., and M.K.; and from RIKEN Center for Advanced Intelligence Project awarded to I.T.
% This work was supported by MEXT KAKENHI to I.T. (16H06538, 17H00758) and M.K. (16H06538, 17H04694); from JST CREST awarded to I.T. (JPMJCR1302, JPMJCR1502) and PRESTO awarded to M.K. (JPMJPR15N2); from the MI2I project of the Support Program for Starting Up Innovation Hub from JST awarded to I.T., and M.K.; and from RIKEN Center for AIP awarded to I.T.
% \end{acks}

\bibliographystyle{unsrt}
%\bibliography{bibliography}

\begin{thebibliography}{10}

\bibitem{weinberger2009distance}
Kilian~Q Weinberger and Lawrence~K Saul.
\newblock Distance metric learning for large margin nearest neighbor
  classification.
\newblock {\em JMLR}, 10(Feb):207--244, 2009.

\bibitem{schultz2004learning}
Matthew Schultz and Thorsten Joachims.
\newblock Learning a distance metric from relative comparisons.
\newblock In {\em Advances in NIPS}, pages 41--48, 2004.

\bibitem{davis2007information}
Jason~V Davis, Brian Kulis, Prateek Jain, Suvrit Sra, and Inderjit~S Dhillon.
\newblock Information-theoretic metric learning.
\newblock In {\em Proc. of the 24th ICML}, pages 209--216. ACM, 2007.

\bibitem{kulis2013metric}
Brian Kulis et~al.
\newblock Metric learning: A survey.
\newblock {\em Foundations and Trends{\textregistered} in Machine Learning},
  5(4):287--364, 2013.

\bibitem{xing2002distance}
Eric~P. Xing, Andrew~Y. Ng, Michael~I. Jordan, and Stuart Russell.
\newblock Distance metric learning, with application to clustering with
  side-information.
\newblock In {\em Advances in NIPS}, pages 521--528, Cambridge, MA, USA, 2002.
  MIT Press.

\bibitem{mcfee2002metric}
Brian McFee and Gert R.~G. Lanckriet.
\newblock Metric learning to rank.
\newblock In {\em Proc. of the 27th ICML}, pages 775--782, Haifa, Israel, 2010.

\bibitem{schroff2015facenet}
Florian Schroff, Dmitry Kalenichenko, and James Philbin.
\newblock Facenet: A unified embedding for face recognition and clustering.
\newblock In {\em IEEE Conf. on CVPR}, pages 815--823, Boston, MA, USA, 2015.
  IEEE Computer Society.

\bibitem{hoffer2015deep}
Elad Hoffer and Nir Ailon.
\newblock Deep metric learning using triplet network.
\newblock In {\em Similarity-Based Pattern Recognition}, pages 84--92, Cham,
  2015. Springer.

\bibitem{jain2009online}
Prateek Jain, Brian Kulis, Inderjit~S. Dhillon, and Kristen Grauman.
\newblock Online metric learning and fast similarity search.
\newblock In {\em Advances in NIPS}, pages 761--768. Curran Associates, Inc.,
  2009.

\bibitem{capitanine2016constraint}
Hoel~Le Capitaine.
\newblock Constraint selection in metric learning.
\newblock {\em arXiv, CoRR}, abs/1612.04853, 2016.

\bibitem{ghaoui2010safe}
Laurent~El Ghaoui, Vivian Viallon, and Tarek Rabbani.
\newblock Safe feature elimination for the lasso and sparse supervised learning
  problems.
\newblock {\em arXiv:1009.4219}, 2010.

\bibitem{wang2013lasso}
Jie Wang, Jiayu Zhou, Peter Wonka, and Jieping Ye.
\newblock Lasso screening rules via dual polytope projection.
\newblock In {\em Advances in NIPS}, pages 1070--1078. Curran Associates, Inc.,
  2013.

\bibitem{liu2014safe}
Jun Liu, Zheng Zhao, Jie Wang, and Jieping Ye.
\newblock Safe screening with variational inequalities and its application to
  lasso.
\newblock In {\em Proc. of the 31st ICML}, pages 289--297, 2014.

\bibitem{fercoq2015mind}
Olivier Fercoq, Alexandre Gramfort, and Joseph Salmon.
\newblock Mind the duality gap: safer rules for the lasso.
\newblock {\em arXiv:1505.03410}, 2015.

\bibitem{xiang2017screening}
Zhen~James Xiang, Yun Wang, and Peter~J Ramadge.
\newblock Screening tests for lasso problems.
\newblock {\em IEEE TPAMI}, 39(5):1008--1027, 2017.

\bibitem{ogawa2013safe}
Kohei Ogawa, Yoshiki Suzuki, and Ichiro Takeuchi.
\newblock Safe screening of non-support vectors in pathwise svm computation.
\newblock In {\em Proc. of the 30th ICML}, pages 1382--1390, 2013.

\bibitem{zhou2015safe}
Qiang Zhou and Qi~Zhao.
\newblock Safe subspace screening for nuclear norm regularized least squares
  problems.
\newblock In {\em Proc. of the 32nd ICML}, volume~37, pages 1103--1112, Lille,
  France, 2015. PMLR.

\bibitem{ndiaye2016safe}
Eugene Ndiaye, Olivier Fercoq, Alexandre Gramfort, and Joseph Salmon.
\newblock Gap safe screening rules for sparse-group lasso.
\newblock In {\em Advances in NIPS}, pages 388--396. Curran Associates, Inc.,
  2016.

\bibitem{boyd2005least}
Stephen Boyd and Lin Xiao.
\newblock Least-squares covariance matrix adjustment.
\newblock {\em SIAM Journal on Matrix Analysis and Applications},
  27(2):532--546, 2005.

\bibitem{malick2004dual}
J{\'e}r{\^o}me Malick.
\newblock A dual approach to semidefinite least-squares problems.
\newblock {\em SIAM Journal on Matrix Analysis and Applications},
  26(1):272--284, 2004.

\bibitem{shen2014efficient}
Chunhua Shen, Junae Kim, Fayao Liu, Lei Wang, and Anton Van Den~Hengel.
\newblock Efficient dual approach to distance metric learning.
\newblock {\em IEEE TNNLS}, 25(2):394--406, 2014.

\bibitem{lehoucq1996deflation}
Richard~B Lehoucq and Danny~C Sorensen.
\newblock Deflation techniques for an implicitly restarted arnoldi iteration.
\newblock {\em SIAM Journal on Matrix Analysis and Applications},
  17(4):789--821, 1996.

\bibitem{boyd2004convex}
Stephen Boyd and Lieven Vandenberghe.
\newblock {\em Convex optimization}.
\newblock Cambridge university press, 2004.

\bibitem{bertsekas1999nonlinear}
Dimitri~P Bertsekas.
\newblock {\em Nonlinear programming}.
\newblock Athena scientific Belmont, 1999.

\bibitem{shibagaki2015regularization}
Atsushi Shibagaki, Yoshiki Suzuki, Masayuki Karasuyama, and Ichiro Takeuchi.
\newblock Regularization path of cross-validation error lower bounds.
\newblock In {\em Advances in NIPS}, pages 1675--1683, 2015.

\bibitem{zimmert2015safe}
Julian Zimmert, Christian~Schroeder de~Witt, Giancarlo Kerg, and Marius Kloft.
\newblock Safe screening for support vector machines.
\newblock In {\em NIPS 2015 Workshop on Optimization in Machine Learning},
  2015.

\bibitem{wang2014scaling}
Jie Wang, Peter Wonka, and Jieping Ye.
\newblock Scaling svm and least absolute deviations via exact data reduction.
\newblock In {\em Proc. of the 31st ICML}, pages 523--531, 2014.

\bibitem{CC01a}
Chih-Chung Chang and Chih-Jen Lin.
\newblock {LIBSVM}: A library for support vector machines.
\newblock {\em ACM Transactions on Intelligent Systems and Technology},
  2:27:1--27:27, 2011.

\bibitem{chollet2015keras}
Fran\c{c}ois Chollet et~al.
\newblock Keras.
\newblock \url{https://github.com/keras-team/keras}, 2015.

\bibitem{barzilai1988two}
Jonathan Barzilai and Jonathan~M Borwein.
\newblock Two-point step size gradient methods.
\newblock {\em IMA journal of numerical analysis}, 8(1):141--148, 1988.

\bibitem{yang1993conjugate}
H~Yang.
\newblock Conjugate gradient methods for the rayleigh quotient minimization of
  generalized eigenvalue problems.
\newblock {\em Computing}, 51(1):79--94, 1993.

\end{thebibliography}

\appendix
%Appendix A

% --------------------------------------------------
\section{Dual Formulation}
\label{app:dual}

To derive dual problem, we first rewrite the primal problem as
\[
\begin{array}{cc}
	\displaystyle\min_{\bm{M},\,\bm{t} }&\displaystyle\sum_{ijl}\ell\left(t_{ijl}\right)+\lambda R(\bm{M})\\
	\mathrm{s.t.}&\bm{M}\succeq\bm{O},~t_{ijl}=\langle \bm{M},\bm{H}_{ijl}\rangle,
\end{array}
\]
where $\bm{t}$ is a $|\cT|$ dimensional vector which contains all $t_{ijl}$ for $(i,j,l) \in \cT$, 
$\sum_{ijl}$ denotes $\sum_{(i,j,l) \in \cT}$, 
and 
\[
R(\bm{M})=\frac{1}{2}\left\|\bm{M}\right\|_F^2.
\]
The Lagrange function is 
\[
L(\bm{M}, \bm{t}, \bm{\alpha}, \bm{\Gamma})\coloneqq\sum_{ijl}\ell\left(t_{ijl}\right)+\lambda R(\bm{M})
  + \sum_{ijl}\alpha_{ijl}(t_{ijl}-\langle \bm{M}, \bm{H}_{ijl}\rangle)- \langle \bm{M},\bm{\Gamma}\rangle, 
\]
where $\bm{\alpha} \in \RR^{|\mathcal{T}|}$ and $\bm{\Gamma} \in \RR_+^{d\times d}$ are Lagrange multipliers. 
Let 
\begin{align}
	\label{eq:conjugate-ell}
	\ell^*(-\alpha_{ijl})&\coloneqq\sup_{t_{ijl}}\{(-\alpha_{ijl})t_{ijl}-\ell\left(t_{ijl}\right)\}, \\
	\label{eq:conjugate-R}
	R^*(\bm{M}_{\lambda}(\bm{\alpha},\bm{\Gamma}) )&\coloneqq\sup_{\bm{M}}\{ \langle \bm{M}_{\lambda}(\bm{\alpha}, \bm{\Gamma}), \bm{M}\rangle-R(\bm{M})\},
\end{align}
be convex conjugate functions \cite{boyd2004convex} of $\ell$ and $R$, where 
\begin{equation}\label{eq:Mlambda-app}
\bm{M}_{\lambda}(\bm{\alpha}, \bm{\Gamma})\coloneqq\frac{1}{\lambda}\Bigl[\sum_{ijl}\alpha_{ijl}\bm{H}_{ijl}+\bm{\Gamma}\Bigr].
\end{equation}
Then, the dual function is written as, 
\begin{align*}
D_{\lambda}(\bm{\alpha},\bm{\Gamma})	&\coloneqq \inf_{\bm{M},\bm{t}}L(\bm{M}, \bm{t}, \bm{\alpha}, \bm{\Gamma})\\
					&=-\sum_{ijl}\sup_{t_{ijl}}\{(-\alpha_{ijl})t_{ijl}-\ell\left(t_{ijl}\right)\}-\lambda\sup_{\bm{M}}\{ \langle \bm{M}_{\lambda}(\bm{\alpha}, \bm{\Gamma}), \bm{M}\rangle-R(\bm{M})\}\\
					&=-\sum_{ijl}\ell^*(-\alpha_{ijl})-\lambda R^*(\bm{M}_{\lambda}(\bm{\alpha},\bm{\Gamma})).
\end{align*}
From the Karush-Kuhn-Tucker (KKT) condition, we obtain
\begin{subequations}
\begin{align}
	&\nabla_{\bm{M}} L=\lambda \nabla R(\bm{M})- \lambda\bm{M}_{\lambda}(\bm{\alpha},\bm{\Gamma})=\bm{O}, \label{eq:kkt1-1}\\
	&\nabla_{t_{ijl}} L=\nabla\ell(t_{ijl})+\alpha_{ijl}=0, \label{eq:kkt1-2}\\
	&\bm{\Gamma}\succeq\bm{O},~\bm{M}\succeq\bm{O},~\langle \bm{M},\bm{H}_{ijl}\rangle = t_{ijl},~\langle\bm{M},\bm{\Gamma}\rangle=0, \label{eq:kkt1-3}
\end{align}
\end{subequations}
where in the case of hinge loss, 
\[
\nabla
\ell(x)=\begin{cases}
0,		&x>1,\\
-c,		&x=1,\\
-1,		&x<1,
\end{cases}
\]
where $\forall c\in[0,1]$, and in the case of smoothed hinge loss,
\[
\nabla
\ell(x)=\begin{cases}
0,							&x>1,\\
-\frac{1}{\gamma}(1-x),		&1-\gamma\le x\le 1,\\
-1,							&x<1-\gamma.
\end{cases}
\]
From these two equations and \eqref{eq:kkt1-2}, we see $\bm{0}\le\bm{\alpha}\le\bm{1}$.
Substituting \eqref{eq:kkt1-2} into \eqref{eq:conjugate-ell} and considering the above constraint, the conjugate of the loss function $\ell$ can be transformed into
\[
\ell^*(-\alpha_{ijl})=\frac{\gamma}{2}\alpha_{ijl}^2-\alpha_{ijl}.
\]
Note that this equation holds for the both cases of the hinge loss (by setting $\gamma = 0$) and the smoothed hinge loss ($\gamma > 0$).
Substituting \eqref{eq:kkt1-1} into \eqref{eq:conjugate-R}, the conjugate of the regularization term $R$ is written as
\[
R^*(\bm{M}_{\lambda}(\bm{\alpha},\bm{\Gamma}))
=R(\bm{M}_{\lambda}(\bm{\alpha},\bm{\Gamma}))
=\frac{1}{2}\|\bm{M}_{\lambda}(\bm{\alpha},\bm{\Gamma})\|_F^2.
\]
Therefore, the dual problem is 
\[
	\tag{Dual1}\label{eq:Dual1-app}
	\max_{\bm{0}\le\bm{\alpha}\le\bm{1},\,\bm{\Gamma}\succeq\bm{O}} D_{\lambda}(\bm{\alpha},\bm{\Gamma})
	=-\sum_{ijl}\ell^*(-\alpha_{ijl})-\frac{\lambda}{2}\|\bm{M}_{\lambda}(\bm{\alpha},\bm{\Gamma})\|_F^2.
\]
Since the second term 
$\displaystyle\max_{\bm{\Gamma}\succeq\bm{O}}-\frac{1}{2}\left\|\bm{M}_{\lambda}(\bm{\alpha},\bm{\Gamma})\right\|_F^2$
is equivalent to the projection onto a semi-definite cone \cite{boyd2005least,malick2004dual}, 
the above problem \eqref{eq:Dual1-app} can be simplified as 
\[
	\tag{Dual2}\label{eq:Dual2-app}
	\max_{\bm{0}\le\bm{\alpha}\le\bm{1}}~
	D_{\lambda}(\bm{\alpha})\coloneqq
	%-\sum_{ijl}\ell^*(-\alpha_{ijl})
	-\frac{\gamma}{2}\|\bm{\alpha}\|_2^2+\bm{\alpha}^\top\bm{1}
	-\frac{\lambda}{2}\bigl\|\bm{M}_{\lambda}(\bm{\alpha})\bigr\|_F^2,
\]
where
\[
\bm{M}_{\lambda}(\bm{\alpha})\coloneqq\frac{1}{\lambda}\Bigl[\sum_{ijl}\alpha_{ijl}\bm{H}_{ijl}\Bigr]_+.
\]
For the optimal $\bm{M}^\star$, each one of triplets in $\cT$ can be categorized into the following three groups:
\begin{equation}
	\begin{split}
		\mathcal{L}^\star&\coloneqq\{(i,j,l)\in\mathcal{T} \mid \langle \bm{H}_{ijl},\bm{M}^\star \rangle < 1-\gamma \}, \\
		\mathcal{C}^\star&\coloneqq\{(i,j,l)\in\mathcal{T} \mid 1-\gamma\le\langle \bm{H}_{ijl},\bm{M}^\star \rangle \le 1 \}, \\
		\mathcal{R}^\star&\coloneqq\{(i,j,l)\in\mathcal{T} \mid \langle \bm{H}_{ijl},\bm{M}^\star \rangle > 1 \}.
	\end{split}
	\label{eq:LCR-app}
\end{equation}
From the equations \eqref{eq:kkt1-2} and \eqref{eq:kkt1-3}, 
we see $\alpha_{ijl}^\star=\nabla\ell(\langle \bm{M}^\star,\bm{H}_{ijl}\rangle)$, 
by which the following rules are obtained.
\begin{equation}
	\begin{split}
	(i,j,l)\in\mathcal{L}^\star&\Rightarrow \alpha_{ijl}^\star=1, \\
	(i,j,l)\in\mathcal{C}^\star&\Rightarrow \alpha_{ijl}^\star\in[0,1], \\
	(i,j,l)\in\mathcal{R}^\star&\Rightarrow \alpha_{ijl}^\star=0.
	\end{split}
	\label{eq:LCRdual-app}
\end{equation}

\section{Sphere Rule with Semi-definite Constraint for Diagonal Case}
\label{app:P3}

We consider a special case that $\bm{M}$ is a diagonal matrix because we can evaluate screening rules with the semi-definite constraint much easier than the general case. 
When $\bm{M}$ is a diagonal matrix, the semi-definite constraint is reduced to the nonnegativity constraint. Then, the minimization problem \eqref{eq:P2} is simplified as
\[
\tag{P3}\label{eq:P3}
%	\begin{array}{cc}\displaystyle
		\min_{\bm{x}\in\mathbb{R}^d}~ \bm{x}^\top\bm{h}_{ijl}~~
		\mathrm{s.t.}~\left\|\bm{x}-\bm{q}\right\|_2^2\le r^2,~\bm{x}\ge\bm{0}, 
%	\end{array}
\]
where $\bm{h}_{ijl}\coloneqq\mathrm{diag}(\bm{H}_{ijl})$. 
Considering the KKT condition of \eqref{eq:P3}, this optimization problem can be solved analytically with $O(d^2)$.
% (see Appendix~\ref{app:P3}). 
SDLS dual ascent is also applicable, 
which can be faster than the analytical calculation for high dimensional case because the computations required for one iteration is $O(d)$. 

We show the analytical solution of \eqref{eq:P3} below.
Let 
\[
L(\bm{x},\alpha,\bm{\beta})\coloneqq\bm{x}^\top\bm{h}_{ijl}-\alpha(r^2-\left\|\bm{x}-\bm{q}\right\|_2^2)-\bm{\beta}^\top\bm{x}. 
\]
be the Lagrange function of \eqref{eq:P3}.
From the KKT condition, we obtain
\begin{subequations}
\begin{align}
	\partial L/\partial\bm{x}=\bm{h}_{ijl}+2\alpha(\bm{x}-\bm{q})-\bm{\beta}=\bm{0}, \label{eq:kkt2-1}\\
	\alpha(r^2-\left\|\bm{x}-\bm{q}\right\|_2^2)=0,~\beta_kx_k=0,\label{eq:kkt2-2}\\
	\alpha\ge0,~ r^2-\left\|\bm{x}-\bm{q}\right\|_2^2\ge0,~ \bm{\beta}\ge\bm{0},~ \bm{x}\ge\bm{0}. \label{eq:kkt2-3}
\end{align}
\end{subequations}
Rearranging an element of \eqref{eq:kkt2-1}, we obtain
\begin{align*}
 \beta_k = 2 \alpha x_k  + (h_{ijl,k} - 2 \alpha q_k). 
 % \\
 % 2 \alpha x_k = \beta_k - (h_{ijl,k} - 2 \alpha q_k).
\end{align*}
When we assume $h_{ijl,k}-2\alpha q_k>0$, we see the following rules:
\begin{align*}
 & h_{ijl,k}-2\alpha q_k>0 \\
 & \Rightarrow
 \beta_k = 
 \underbrace{2\alpha x_k}_{\ge0}+\underbrace{(h_{ijl,k}-2\alpha q_k)}_{>0}
 >
 0 \\
 & \Rightarrow
 x_k  = 0.
\end{align*}
The last equation is derived from the complementary condition $\beta_k x_k = 0$.
On the other hand, when we assume $h_{ijl,k}-2\alpha q_k\leq0$, we have
\begin{align*}
 & h_{ijl,k}-2\alpha q_k \leq 0 \\
 & \Rightarrow
 2 \alpha x_k - \beta_k = - (h_{ijl,k} - 2 \alpha q_k) \geq 0 \\
 & \Rightarrow \beta_k = 0 \\
 & \Rightarrow x_k =  q_k - h_{ijl,k}  / 2 \alpha.
 % - (h_{ijl,k} - 2 \alpha q_k)  / 2 \alpha
\end{align*}
The second last equation is derived from $x_k \geq 0$, $\beta_k \geq 0$ and the complementary condition $\beta_k x_k = 0$.
Using the above two derivations, we obtain
\begin{align}
 x_k =
 \begin{cases}
  q_k - h_{ijl,k}  / 2 \alpha, & \text{ if } h_{ijl,k}-2\alpha q_k \leq 0, \\
  0, & \text{ otherwise. }
 \end{cases}
 \label{eq:opt-x-diag}
\end{align}
If we can determine $k$ which satisfies $h_{ijl,k}-2\alpha q_k \leq 0$, i.e., 
$\{ k \mid h_{ijl,k}-2\alpha q_k \leq 0 \}$
, we can calculate $\alpha$ by substituting \eqref{eq:opt-x-diag} into the condition
\begin{align*}
 \| \bm{x} - \bm{q} \|_2^2 = r,
\end{align*}
which indicates that the optimal $\bm{x}$ must be at the boundary of the sphere.
We can easily investigate all possible patters of
$\{ k \mid h_{ijl,k}-2\alpha q_k \leq 0 \}$,
which only needs to consider at most $d$ different values of $\alpha$s.
Let $\alpha_1, \alpha_2, \ldots$ be a set of values defined by sorting the elements in the following set with increasing order:
\begin{align*}
 \left\{
 h_{ijl,k} / 2 q_k \mid
 h_{ijl,k} / 2 q_k \geq 0, 
 q_k \ne 0
 \right\}.
\end{align*}
Defining $\alpha_0 = 0$, we consider a set of intervals $[\alpha_k,\alpha_{k+1}]$ for $k \geq 0$.
Assuming $\alpha \in [\alpha_k,\alpha_{k+1}]$, 
we can define 
$\{ k \mid h_{ijl,k}-2\alpha q_k \leq 0 \}$
and calculate \eqref{eq:opt-x-diag}.
If all the KKT conditions \eqref{eq:kkt2-1}-\eqref{eq:kkt2-3} are satisfied, we can obtain the optimal $\bm{x}$, otherwise the next interval 
$[\alpha_{k+1},\alpha_{k+2}]$
should be investigated.
By repeating this procedure at most $d$ times the optimal solution can be found. 
Since the computational cost for one specific interval $[\alpha_k,\alpha_{k+1}]$ is $O(d)$, the total computational cost is $O(d^2)$.

\section{Proof of Theorem \ref{thm:P4}}
\label{app:solveP4}

The Lagrange function is defined as follows:
\[
L(\bm{X}, \alpha, \beta)\coloneqq\langle \bm{X}, \bm{H}_{ijl}\rangle
-\alpha\frac{1}{2}(r^2-\left\|\bm{X}-\bm{Q}\right\|_F^2)-\beta\langle \bm{P}, \bm{X}\rangle.
\]
From the KKT condition, we obtain
\begin{subequations}
\begin{align}
	\partial L/\partial \bm{X}=\bm{H}_{ijl}+\alpha(\bm{X}-\bm{Q})-\beta \bm{P}=\bm{O}.\label{eq:kkt3-1}\\
	\alpha\ge0,~\beta\ge0,~\left\|\bm{X}-\bm{Q}\right\|_F^2\le r^2,~\langle \bm{P}, \bm{X}\rangle \ge 0.\label{eq:kkt3-2}\\
	\alpha(r^2-\left\|\bm{X}-\bm{Q}\right\|_F^2)=0,~\beta\langle \bm{P}, \bm{X}\rangle=0.\label{eq:kkt3-3}
\end{align}
\end{subequations}
If $\alpha=0$, then $\bm{H}_{ijl}=\beta\bm{P}$ from \eqref{eq:kkt3-1}, and the value of the objective function becomes $\langle \bm{X}, \bm{H}_{ijl}\rangle=\beta\langle \bm{X}, \bm{P}\rangle=0$ from \eqref{eq:kkt3-3}. 
Let us consider the case of $\alpha\ne0$.
From \eqref{eq:kkt3-3}, we see $\left\|\bm{X}-\bm{Q}\right\|_F^2=r^2$. 
If $\beta=0$, the linear constraint is 
not an active constraint (i.e., $\langle \bm{P},\bm{X} \rangle > 0$ at the optimal), and
so it is the same as the problem \eqref{eq:P1}, which can be analytically solved. 
If this solution satisfies the linear constraint $\langle \bm{P}, \bm{X}\rangle \ge 0$, it becomes the optimal solution.
Next, we consider the case of $\beta\ne0$. 
From \eqref{eq:kkt3-1} and \eqref{eq:kkt3-3}, $\alpha$ and $\beta$ are obtained as follows.
\[
\alpha=\pm\sqrt{\frac{\|\bm{P}\|_F^2\|\bm{H}_{ijl}\|_F^2-\langle \bm{P},\bm{H}_{ijl}\rangle^2}{r^2\|\bm{P}\|_F^2-\langle\bm{P},\bm{Q}\rangle^2}},
\beta=\frac{\langle \bm{P},\bm{H}_{ijl}\rangle-\alpha\langle \bm{P},\bm{Q}\rangle}{\|\bm{P}\|_F^2}.
\]
For the solutions of the two $\alpha$, $\alpha>0$ gives the minimum value from \eqref{eq:kkt3-2}.

% --------------------------------------------------
\section{Proof of Theorem \ref{thm:GB} (GB)}
\label{app:GB}

The following theorem is a well-known optimality condition for the general convex optimization problem:
\begin{theorem}[Optimality condition of convex optimization, \cite{bertsekas1999nonlinear}]
	\label{thm:cvxopt}
	In the minimization problem 
	$\min_{\bm{x}\in\mathcal{F}}f(\bm{x})$
	where the feasible region $\mathcal{F}$ and the function $f(\bm{x})$ are convex, 
	the necessary and sufficient condition that $\bm{x}^\star$ is the optimal solution is 
	\[
	\exists \nabla f(\bm{x}^\star)\in\partial f(\bm{x}^\star)\left[\nabla f(\bm{x}^\star)^\top (\bm{x}^\star-\bm{x})\le0,~\forall \bm{x}\in\mathcal{F}\right], 
	\]
	where $\partial f(\bm{x}^\star)$ represents the set of subgradient in $\bm{x}^\star$.
\end{theorem}\noindent

From \textbf{Theorem} \ref{thm:cvxopt}, the following holds for the optimal solution $\bm{M}^\star$.
\begin{equation}\label{eq:primal_opt}
\langle \nabla P_{\lambda}(\bm{M}^\star), \bm{M}^\star-\bm{M}\rangle \le 0,~ \forall \bm{M} \succeq \bm{O}.
\end{equation}
Let $\bm{\Xi}_{ijl}(\bm{M})$ be the subgradient of the loss function $\ell(\langle\bm{M},\bm{H}_{ijl}\rangle)$ at $\bm{M}$.
Then, $\nabla P_{\lambda}(\bm{M})$ is written as
\begin{equation}\label{eq:primal_gradient}
\nabla P_{\lambda}(\bm{M})=\sum_{ijl}\bm{\Xi}_{ijl}(\bm{M})+\lambda \bm{M}.
\end{equation}
From the convexity of the (smoothed) hinge loss function $\ell(\langle\bm{M},\bm{H}_{ijl}\rangle)$, we obtain
\begin{gather*}
\ell(\langle\bm{M}^\star,\bm{H}_{ijl}\rangle)\ge\ell(\langle\bm{M},\bm{H}_{ijl}\rangle)+\langle \bm{\Xi}_{ijl}(\bm{M}),\bm{M}^\star-\bm{M}\rangle, \\
\ell(\langle\bm{M},\bm{H}_{ijl}\rangle)\ge\ell(\langle\bm{M}^\star,\bm{H}_{ijl}\rangle)+\langle \bm{\Xi}_{ijl}(\bm{M}^\star),\bm{M}-\bm{M}^\star\rangle,
\end{gather*}
for any subgradient.
By adding these two equations, we see
\begin{equation}\label{eq:loss_inequality}
\langle \bm{\Xi}_{ijl}(\bm{M}^\star),\bm{M}^\star-\bm{M}\rangle\ge\langle \bm{\Xi}_{ijl}(\bm{M}),\bm{M}^\star-\bm{M}\rangle.
\end{equation}
Combining above \eqref{eq:primal_opt}, \eqref{eq:primal_gradient} and \eqref{eq:loss_inequality} results in 
\begin{gather*}
\Bigl\langle \sum_{ijl}\bm{\Xi}_{ijl}(\bm{M})+\lambda \bm{M}^\star, \bm{M}^\star-\bm{M}\Bigr\rangle \le 0\\
\Leftrightarrow
\langle \nabla P_{\lambda}(\bm{M})-\lambda\bm{M}+\lambda\bm{M}^\star, \bm{M}^\star-\bm{M}\rangle \le 0. 
\end{gather*}
By transforming this inequality based on completing the square, we obtain GB.

% --------------------------------------------------
\section{Proof of Theorem \ref{thm:PGB} (PGB)}
\label{app:PGBproof}
Let $\bm{Q}^{\mathrm{GB}}$ be the center of GB hypersphere, and $r_{\mathrm{GB}}$ be the radius.
The optimal solution exists in the following set:
% At this time, the optimal solution belongs to the next set. 
\begin{equation}\label{eq:gb_set}
\{\bm{X}\mid\|\bm{X}-\bm{Q}^{\mathrm{GB}}\|_F^2\le r_{\mathrm{GB}}^2,~\bm{X}\succeq\bm{O}\}.
\end{equation}
By transforming the sphere of GB, we obtain
\begin{align*}
\|\bm{X}-\bm{Q}^{\mathrm{GB}}\|_F^2=&\|\bm{X}-(\bm{Q}_+^{\mathrm{GB}}+\bm{Q}_-^{\mathrm{GB}})\|_F^2\\
=&\|\bm{X}-\bm{Q}_+^{\mathrm{GB}}\|_F^2+2\langle \bm{X},-\bm{Q}_-^{\mathrm{GB}}\rangle+2\langle \bm{Q}_+^{\mathrm{GB}},\bm{Q}_-^{\mathrm{GB}}\rangle+\|\bm{Q}_-^{\mathrm{GB}}\|_F^2.
\end{align*}
Since $\bm{X}\succeq \bm{O}$ and $-\bm{Q}_-^{\mathrm{GB}}\succeq \bm{O}$, we see
$\langle \bm{X},-\bm{Q}_-^{\mathrm{GB}}\rangle\ge0$. 
Further, using $\langle \bm{Q}_+^{\mathrm{GB}},\bm{Q}_-^{\mathrm{GB}}\rangle=0$, we obtain the following sphere:
\[
\begin{array}{c}
	r_{\mathrm{GB}}^2\ge\|\bm{X}-\bm{Q}^{\mathrm{GB}}\|_F^2\ge\|\bm{X}-\bm{Q}_+^{\mathrm{GB}}\|_F^2+\|\bm{Q}_-^{\mathrm{GB}}\|_F^2. \\
	\therefore~\|\bm{X}-\bm{Q}_+^{\mathrm{GB}}\|_F^2\le r_{\mathrm{GB}}^2-\|\bm{Q}_-^{\mathrm{GB}}\|_F^2.
\end{array}
\]
Letting $\bm{Q}^{\rm PGB}\coloneqq\bm{Q}_+^{\mathrm{GB}}$ and $r_{\rm PGB}^2\coloneqq r_{\mathrm{GB}}^2-\|\bm{Q}_-^{\mathrm{GB}}\|_F^2$, 
PGB is obtained. 
Note that by considering $\langle \bm{X},-\bm{Q}_-^{\mathrm{GB}}\rangle\ge0$ instead of $\bm{X}\succeq \bm{O}$ in \eqref{eq:gb_set}, 
we can immediately see that
GB with Linear constraint $\langle \bm{X},-\bm{Q}_-^{\mathrm{GB}}\rangle\ge0$
is tighter than PGB. 

% --------------------------------------------------
\section{Proof of CDGB}

% --------------------------------------------------
% \section{Proof of Theorem \ref{thm:GP} (Strong convexity of $G_{P_{\lambda}}$)}
\subsection{Proof of Strong Convexity of $G_{P_{\lambda}}$}
\label{app:convexityGP}

We first defines $m$-strongly convex function as follows:
\begin{definition}[$m$-strongly convex function]
 \label{def:strong}
 When $f(\bm{x})-\frac{m}{2}\left\|\bm{x}\right\|_2^2$ is a convex function, 
 $f(\bm{x})$ is $m$-strongly convex function. 
\end{definition}\noindent

According to definition~\ref{def:strong}, 
in order to show that $G_{P_{\lambda}}$ is strongly convex, we need to show that the term other than $\lambda\left\|\bm{M}\right\|_F^2$ is convex. 
\[
G_{P_{\lambda}}(\bm{M})=\underbrace{\sum_{ijl}\ell(\langle \bm{M},\bm{H}_{ijl}\rangle)}_{\mathrm{convex}}+\lambda\left\|\bm{M}\right\|_F^2\\
+\underbrace{\min_{
\scriptsize
\begin{array}{c}
\bm{0}\le\bm{\alpha}\le\bm{1}, \bm{\Gamma} \succeq \bm{O},
\bm{M}_{\lambda}(\bm{\alpha},\bm{\Gamma})=\bm{M}
\end{array}
}\underbrace{\sum_{ijl}\ell^*(-\alpha_{ijl})}_{\eqqcolon g(\bm{\alpha})}
}_{\eqqcolon f(\bm{M})}.
\]
Since the loss $\ell$ is convex, we need to show that $f(\bm{M})$ is convex.
This can be shown as below.
\[
	f(\bm{M})=
	\min_{
	\scriptsize
	\begin{array}{c}
	\bm{0}\le\bm{\alpha}\le\bm{1}, \bm{\Gamma} \succeq \bm{O},\\
	\displaystyle\frac{1}{\lambda}\bigl[\sum_{ijl}\alpha_{ijl}\bm{H}_{ijl}+\bm{\Gamma}\bigr]=\bm{M}
	\end{array}
	}g(\bm{\alpha})
	=
	\min_{
	\scriptsize
	\begin{array}{c}
	\bm{0}\le\bm{\alpha}\le \bm{1}, \\
	\displaystyle\frac{1}{\lambda}\sum_{ijl}\alpha_{ijl}\bm{H}_{ijl}\preceq \bm{M}
	\end{array}
	}g(\bm{\alpha}).
\]
Consider a point 
$
\bm{M}_2=t\,\bm{M}_0 + (1-t)\,\bm{M}_1~(t\in[0,1])
$
internally dividing two points $\bm{M}_0$ and $\bm{M}_1$.
Let 
\begin{align*}
 \bm{\alpha}_i^\star 
 \coloneqq
 % :=
 \argmin_{
	\scriptsize
	\begin{array}{c}
	\bm{0}\le\bm{\alpha}\le \bm{1}, \\
	\displaystyle\frac{1}{\lambda}\sum_{ijl}\alpha_{ijl}\bm{H}_{ijl}\preceq \bm{M}_i
	\end{array}
	}g(\bm{\alpha}),
\end{align*}
which means that
$\bm{\alpha}_i^\star$
is the minimizer of this problem for a given
% the optimal solution of $\min$ for 
$\bm{M}_i (i \in \{0, 1, 2\})$,
and from the definition, we see
$
f(\bm{M}_i)=g(\bm{\alpha}_i^\star).
$
Further, let 
$
\bm{\alpha}_2 = t\,\bm{\alpha}_0^\star + (1-t)\,\bm{\alpha}_1^\star
$.
Then, $\bm{0}\le\bm{\alpha}_2\le\bm{1}$ and $\frac{1}{\lambda}\sum_{ijl}\alpha_{2,ijl}\bm{H}_{ijl}\preceq \bm{M}_2$. 
Since $g$ is convex because of the convexity of $\ell^*$, we have
\begin{align*}
t\,f(\bm{M}_0) + (1-t)\,f(\bm{M}_1)&=t\,g(\bm{\alpha}_0^\star) + (1-t)\,g(\bm{\alpha}_1^\star)\\
									&\ge g(\underbrace{t \bm{\alpha}_0^\star+(1-t) \bm{\alpha}_1^\star}_{\bm{\alpha}_2})\\
									&\ge g(\bm{\alpha}_2^\star)= f(\underbrace{t\,\bm{M}_0 + (1-t)\,\bm{M}_1}_{\bm{M}_2}).
\end{align*}
Hence, $f(\bm{M})$ is convex and $G_{P_{\lambda}}$ is a strongly convex function.
%
% In \refname \cite{zimmert2015safe} we referred to, Danskin's theorem is used to show that the function corresponding to $-f$ is concave, but in this paper we followed the definition of convex. 

% --------------------------------------------------
\subsection{Proof of Theorem \ref{thm:CDGB} (CDGB)}
\label{app:CDGB}
%定理\ref{thm:GP}，\ref{thm:strong}から，

From the strong convexity of $G_{P_\lambda}$ shown in Appendix~\ref{app:convexityGP},
% From the nature of the strongly convex function of $G_{P_\lambda}$, 
the following holds
for any $\bm{M}_{\lambda}(\bm{\alpha},\bm{\Gamma}),\bm{M}^\star\succeq\bm{O}$:
% \begin{multline*}
\[
 G_{P_{\lambda}}(\bm{M}_{\lambda}(\bm{\alpha},\bm{\Gamma}))\ge G_{P_{\lambda}}(\bm{M}^\star)
 +\langle \nabla G_{P_{\lambda}}(\bm{M}^\star), \bm{M}_{\lambda}(\bm{\alpha},\bm{\Gamma})-\bm{M}^\star\rangle
 +\lambda\left\|\bm{M}_{\lambda}(\bm{\alpha},\bm{\Gamma})-\bm{M}^\star\right\|_F^2. 
\]
% \end{multline*}
We assume that $\bm{M}^\star$ is the optimal solution of the primal problem. 
Then, since $\bm{M}^\star$ is also a solution to the convex optimization problem $\min_{\bm{M}\succeq\bm{O}}G_{P_{\lambda}}(\bm{M})$, we see
$\langle \nabla G_{P_{\lambda}}(\bm{M}^\star), \bm{M}_{\lambda}(\bm{\alpha},\bm{\Gamma})-\bm{M}^\star\rangle\ge 0$ 
from \textbf{Theorem} \ref{thm:cvxopt}. 
Considering 
$G_{P_{\lambda}}(\bm{M}^\star)=0$ 
and
$G_{D_{\lambda}}(\bm{\alpha},\bm{\Gamma})\ge G_{P_{\lambda}}(\bm{M}_{\lambda}(\bm{\alpha},\bm{\Gamma}))$, both of which are from the definition, we obtain
%
% In the optimal solution $\bm{M}^\star$, since the duality gap is 0, $G_{P_{\lambda}}(\bm{M}^\star)=0$.
% Furthermore, from the definition of $G_{P_{\lambda}}$, 
% $G_{D_{\lambda}}(\bm{\alpha},\bm{\Gamma})\ge G_{P_{\lambda}}(\bm{M}_{\lambda}(\bm{\alpha},\bm{\Gamma}))$. 
% Utilizing these, 
\[
G_{D_{\lambda}}(\bm{\alpha},\bm{\Gamma})\ge G_{P_{\lambda}}(\bm{M}_{\lambda}(\bm{\alpha},\bm{\Gamma}))
\ge\lambda\left\|\bm{M}_{\lambda}(\bm{\alpha},\bm{\Gamma})-\bm{M}^\star\right\|_F^2.
\]
Dividing by $\lambda$, CDGB is derived.

\section{Proof of Theorem \ref{thm:RPB} (RPB)}
\label{app:RPB}
From the optimality condition (\textbf{Theorem} \ref{thm:cvxopt}) in the dual problem \eqref{eq:Dual1} for $\lambda_0,\lambda_1$, 
\begin{gather*}
	\nabla_{\bm{\alpha}} D_{\lambda_0}(\bm{\alpha}_0^\star,\bm{\Gamma}_0^\star)^\top (\bm{\alpha}_1^\star-\bm{\alpha}_0^\star)
	+\langle \nabla_{\bm{\Gamma}} D_{\lambda_0}(\bm{\alpha}_0^\star,\bm{\Gamma}_0^\star), \bm{\Gamma}_1^\star-\bm{\Gamma}_0^\star \rangle
	\le0, \\
	\nabla_{\bm{\alpha}} D_{\lambda_1}(\bm{\alpha}_1^\star,\bm{\Gamma}_1^\star)^\top (\bm{\alpha}_0^\star-\bm{\alpha}_1^\star)
	+\langle \nabla_{\bm{\Gamma}} D_{\lambda_1}(\bm{\alpha}_1^\star,\bm{\Gamma}_1^\star), \bm{\Gamma}_0^\star-\bm{\Gamma}_1^\star \rangle
	\le0.
\end{gather*}
By adding these two equations, we obtain
\[
[\nabla_{\bm{\alpha}} D_{\lambda_0}(\bm{\alpha}_0^\star,\bm{\Gamma}_0^\star)-\nabla_{\bm{\alpha}} D_{\lambda_1}(\bm{\alpha}_1^\star,\bm{\Gamma}_1^\star)]^\top
(\bm{\alpha}_1^\star-\bm{\alpha}_0^\star)
+\langle \nabla_{\bm{\Gamma}} D_{\lambda_0}(\bm{\alpha}_0^\star,\bm{\Gamma}_0^\star)-\nabla_{\bm{\Gamma}} D_{\lambda_1}(\bm{\alpha}_1^\star,\bm{\Gamma}_1^\star), 
\bm{\Gamma}_1^\star-\bm{\Gamma}_0^\star \rangle
\le0.
\]
Next, we consider the following difference of gradient 
\begin{align*}
\nabla_{\bm{\alpha}_{ijl}} D_{\lambda_0}(\bm{\alpha}_0^\star,\bm{\Gamma}_0^\star)
-\nabla_{\bm{\alpha}_{ijl}} D_{\lambda_1}(\bm{\alpha}_1^\star,\bm{\Gamma}_1^\star)
&=-\gamma({\alpha_0^\star}_{ijl}-{\alpha_1^\star}_{ijl})-\langle \bm{H}_{ijl},\bm{M}_0^\star-\bm{M}_1^\star\rangle, \\
\nabla_{\bm{\Gamma}} D_{\lambda_0}(\bm{\alpha}_0^\star,\bm{\Gamma}_0^\star)
-\nabla_{\bm{\Gamma}} D_{\lambda_1}(\bm{\alpha}_1^\star,\bm{\Gamma}_1^\star)
&=-(\bm{M}_0^\star-\bm{M}_1^\star).
\end{align*}
% Here, representing 
Defining 
$
\bm{H}_t^\star\coloneqq \sum_{ijl}{\alpha_t^\star}_{ijl}\bm{H}_{ijl},
$
% we can express
$\bm{M}_t^\star$ is re-written as
$
\bm{M}_t^\star=\frac{1}{\lambda_t}[\bm{H}_t^\star+\bm{\Gamma}_t^\star],
$.
Then, we see
% by using these equations, 
\begin{align*}
&\gamma\|\bm{\alpha}_1^\star-\bm{\alpha}_0^\star\|_2^2-\langle \bm{H}_1^\star-\bm{H}_0^\star,\bm{M}_0^\star-\bm{M}_1^\star\rangle-\langle \bm{M}_0^\star-\bm{M}_1^\star, \bm{\Gamma}_1^\star-\bm{\Gamma}_0^\star \rangle\le 0\\
\Leftrightarrow
&\gamma\|\bm{\alpha}_1^\star-\bm{\alpha}_0^\star\|_2^2
-\langle \lambda_1\bm{M}_1^\star-\lambda_0\bm{M}_0^\star,\bm{M}_0^\star-\bm{M}_1^\star\rangle \le 0\\
\Rightarrow&
-\langle \lambda_1\bm{M}_1^\star-\lambda_0\bm{M}_0^\star,\bm{M}_0^\star-\bm{M}_1^\star\rangle \le 0.
\end{align*}
By transforming this inequality based on completing the square, RPB is obtained.

% --------------------------------------------------
\section{Proof of Theorem \ref{thm:PGBvsRPB} (Relationship Between PGB and RPB)}
\label{app:PGBvsRPB}

When the dual variable is used as the subgradient of the (smoothed) hinge loss at the optimal solution $\bm{M}_0^\star$ of $\lambda_0$ (from \eqref{eq:LCRdual-app}, we see that the optimal dual variable provides valid subgradient), the gradient of the objective function in the case of $\lambda_1$ is written as follows:
% When the value of the dual variable is used as the value of the subgradient of the (smoothed) hinge loss at the optimal solution $\bm{M}_0^\star$ in the case of $\lambda_0$, the gradient of the objective function in the case of $\lambda_1$ can be written as follows:
\[
\nabla P_{\lambda_1}(\bm{M}_0^\star)=-\bm{H}_0^\star+\lambda_1\bm{M}_0^\star,
\]
where 
\[
\bm{H}_0^\star\coloneqq-\sum_{ijl}\nabla\ell(\langle \bm{M}_0^\star, \bm{H}_{ijl}\rangle)\bm{H}_{ijl}=\sum_{ijl}{\alpha_0^\star}_{ijl}\bm{H}_{ijl}.
\]
Since $\lambda_0\bm{M}_0^\star={\bm{H}_0^\star}_+$,
\begin{align*}
	\nabla P_{\lambda_1}(\bm{M}_0^\star)&=-({\bm{H}_0^\star}_++{\bm{H}_0^\star}_-)+\lambda_1\bm{M}_0^\star\\
						&=(\lambda_1-\lambda_0)\bm{M}_0^\star-{\bm{H}_0^\star}_-.
\end{align*}
Then, the center and radius of GB are 
\begin{align*}
\bm{Q}^{\mathrm{GB}}&=\bm{M}_0^\star-\frac{1}{2\lambda_1}\nabla P_{\lambda_1}(\bm{M}_0^\star)
=\frac{(\lambda_0+\lambda_1)\bm{M}_0^\star+{\bm{H}_0^\star}_-}{2\lambda_1}, \\
r_{\mathrm{GB}}^2&=\frac{\left\|(\lambda_1-\lambda_0)\bm{M}_0^\star-{\bm{H}_0^\star}_-\right\|_F^2}{4\lambda_1^2}\\
&=\frac{\left\|(\lambda_1-\lambda_0)\bm{M}_0^\star\right\|_F^2-2(\lambda_1-\lambda_0)\langle\bm{M}_0^\star,{\bm{H}_0^\star}_-\rangle+\left\|{\bm{H}_0^\star}_-\right\|_F^2}{4\lambda_1^2}\\
&=\frac{\left\|(\lambda_0-\lambda_1)\bm{M}_0^\star\right\|_F^2+\left\|{\bm{H}_0^\star}_-\right\|_F^2}{4\lambda_1^2}.
\end{align*}
Here, the last equation of $r_{\mathrm{GB}}^2$ uses the fact that $\bm{M}_0^*$ and ${\bm{H}_0^*}_-$ are orthogonal.
Using $\bm{Q}^{\mathrm{GB}}$ and $r_{\mathrm{GB}}^2$, 
the center and radius of PGB are found to be 
\begin{align*}
\bm{Q}^{\mathrm{PGB}}=\bm{Q}_+^{\mathrm{GB}}=\frac{(\lambda_0+\lambda_1)\bm{M}_0^\star}{2\lambda_1},~\bm{Q}_-^{\mathrm{GB}}=\frac{{\bm{H}_0^\star}_-}{2\lambda_1}, \\
 r_{\mathrm{PGB}}^2=r_{\mathrm{GB}}^2-\left\|\bm{Q}_-^{\mathrm{GB}}\right\|_F^2=\frac{\left\|(\lambda_0-\lambda_1)\bm{M}_0^\star\right\|_F^2}{4\lambda_1^2}.
\end{align*}
Therefore, PGB coincides with RPB.

% --------------------------------------------------
\section{Proof of Theorem \ref{thm:DGBvsRPB} (Relationship Between DGB and RPB)} \label{app:DGBvsRPB}
At the optimal solution $\bm{M}_0^\star, \bm{\alpha}_0^\star$ and $\bm{\Gamma}_0^\star$ of $\lambda_0$, we obtain the following equation from $P_{\lambda_0}(\bm{M}_0^\star)=D_{\lambda_0}(\bm{\alpha}_0^\star, \bm{\Gamma}_0^\star)$ 
and $\bm{M}_{\lambda_0}(\bm{\alpha}_0^\star,\bm{\Gamma}_0^\star)=\bm{M}_0^\star$: 
\[
\sum_{ijl}\ell(\langle \bm{M}_0^\star, \bm{H}_{ijl}\rangle)+\sum_{ijl}\ell^*(-{\alpha_0^\star}_{ijl})=-\lambda_0\left\|\bm{M}_0^\star\right\|_F^2.
\]
We also see
$\bm{M}_{\lambda_1}(\bm{\alpha}_0^\star,\bm{\Gamma}_0^\star)=\frac{\lambda_0}{\lambda_1}\bm{M}_{\lambda_0}(\bm{\alpha}_0^\star, \bm{\Gamma}_0^\star)=\frac{\lambda_0}{\lambda_1}\bm{M}_0^\star$. 
Using these results, the value of duality gap for $\lambda_1$ is 
\[
	P_{\lambda_1}(\bm{M}_0^\star)-D_{\lambda_1}(\bm{\alpha}_0^\star,\bm{\Gamma}_0^\star)
	=\frac{(\lambda_0-\lambda_1)^2}{2\lambda_1}\left\|\bm{M}_0^\star\right\|_F^2.
\]
Therefore, the radius of DGB $r_{\mathrm{DGB}}$ and the radius of RPB $r_{\mathrm{RPB}}$ satisfy the following relationship.
\begin{align*}
r_{\mathrm{DGB}}^2&=\frac{2(P_{\lambda_1}(\bm{M}_0^\star)-D_{\lambda_1}(\bm{\alpha}_0^\star,\bm{\Gamma}_0^\star))}{\lambda_1}\\
&=\frac{(\lambda_0-\lambda_1)^2}{\lambda_1^2}\left\|\bm{M}_0^\star\right\|_F^2
=4\,r_{\mathrm{RPB}}^2.
\end{align*}
Also, the center of these hyperspheres are 
\[
\bm{Q}^{\rm DGB}=\bm{M}_0^\star,~
\bm{Q}^{\rm RPB}=\frac{\lambda_0+\lambda_1}{2\lambda_1}\bm{M}_0^\star,
\]
and the distance between the centers is 
\[
\|\bm{Q}^{\rm DGB}-\bm{Q}^{\rm RPB}\|_F
=\frac{|\lambda_0-\lambda_1|}{2\lambda_1}\|\bm{M}_0^\star\|_F=r_{\rm RPB}.
\]
We thus see that DGB includes RPB illustrated as \figurename~\ref{fig:DGBvsRPB}.

% --------------------------------------------------
\section{Proof of Theorem \ref{thm:RRPB} (RRPB)} 
\label{app:RRPB}

Considering
a hypersphere that expands the RPB radius by $\frac{\lambda_0+\lambda_1}{2\lambda_1}\epsilon$ and 
replaces the RPB center with 
$\frac{\lambda_0+\lambda_1}{2\lambda_1}\bm{M}_0$,
we obtain 

% From \figurename~\ref{fig:RRPB}, 
% considering the following hypersphere that expand the RPB radius by $\frac{\lambda_0+\lambda_1}{2\lambda_1}\epsilon$ 
% when replacing the RPB center with $\frac{\lambda_0+\lambda_1}{2\lambda_1}\bm{M}_0$, 
% it can cover any RPB made from $\bm{M}_0^\star$ moving $\|\bm{M}_0^\star-\bm{M}_0\|_F\le\epsilon$.
\[
\left\|\bm{M}_1^\star-\frac{\lambda_0+\lambda_1}{2\lambda_1}\bm{M}_0\right\|_F
\le
\frac{|\lambda_0-\lambda_1|}{2\lambda_1}\left\|\bm{M}_0^\star\right\|_F
+\frac{\lambda_0+\lambda_1}{2\lambda_1}\epsilon.
\]
Since $\epsilon$ is defined by
$\|\bm{M}_0^\star-\bm{M}_0\|_F\le\epsilon$,
this sphere covers 
any RPB made by $\bm{M}_0^\star$ which satisfies $\|\bm{M}_0^\star-\bm{M}_0\|_F\le\epsilon$
(see \figurename~\ref{fig:RRPB} for a geometrical illustration).
Using the reverse triangle inequality
\[
\|\bm{M}_0^\star\|_F-\|\bm{M}_0\|_F\le\|\bm{M}_0^\star-\bm{M}_0\|_F\le\epsilon,
\]
the following is obtained.
\[
\left\|\bm{M}_1^\star-\frac{\lambda_0+\lambda_1}{2\lambda_1}\bm{M}_0\right\|_F
\le
\frac{|\lambda_0-\lambda_1|}{2\lambda_1}(\left\|\bm{M}_0\right\|_F+\epsilon)
+\frac{\lambda_0+\lambda_1}{2\lambda_1}\epsilon.
\]
By rearranging this, RRPB is obtained.

% --------------------------------------------------
\section{Range Based Extension}

% --------------------------------------------------
\subsection{Generalized Form of GB, DGB, RPB and RRPB}
\label{app:general-sphere}

\paragraph*{(GB)}

The gradient is written as
\[
\nabla P_{\lambda}(\bm{M})=\bm{\Xi} + \lambda\bm{M}.
\]
Then the squared norm of this gradient is
\[
 \|\nabla P_{\lambda}(\bm{M})\|_F^2=\|\bm{\Xi}\|_F^2+2\lambda\langle \bm{\Xi}, \bm{M}\rangle+\lambda^2\|\bm{M}\|_F^2.
\]
By substituting this into the center and the radius of GB, we obtain
%\[
%r_{\rm GB}^2 = \frac{1}{4\lambda^2}\|\nabla P_{\lambda}(\bm{M})\|_F^2, \\
%\bm{Q}^{\rm GB} = \bm{M}-\frac{1}{2\lambda}(\bm{\Xi}+\lambda\bm{M})
%\]
%
\begin{align*}
 r_{\rm GB}^2 
 &= \frac{1}{4\lambda^2}\|\nabla P_{\lambda}(\bm{M})\|_F^2
 \\
 &=
 \frac{1}{4\lambda^2} 
 \left(
 \|\bm{\Xi}\|_F^2+2\lambda\langle \bm{\Xi}, \bm{M}\rangle+\lambda^2\|\bm{M}\|_F^2
 \right)
 \\
 &=
 \frac{1}{4}  \|\bm{M}\|_F^2 
 +  \frac{1}{2\lambda} \langle \bm{\Xi}, \bm{M}\rangle
 \frac{1}{4\lambda^2}  \|\bm{\Xi}\|_F^2,
 \\
 % \|\nabla P_{\lambda}(\bm{M})\|_F^2
 \bm{Q}^{\rm GB} &= 
 \bm{M}-\frac{1}{2\lambda}(\bm{\Xi}+\lambda\bm{M}) \\
 &= \frac{1}{2} \bm{M}  - \frac{1}{2\lambda} \bm{\Xi}.
\end{align*}

\paragraph*{(DGB)}

The duality gap is written as 
\[
\mathrm{gap}=\sum_{ijl}(\ell(\langle \bm{M}, \bm{H}_{ijl}\rangle)+\ell^*(-\alpha_{ijl}))+\frac{\lambda}{2}\|\bm{M}\|_F^2+\frac{1}{2\lambda}\|\sum_{ijl}\alpha_{ijl}\bm{H}_{ijl}+\bm{\Gamma}\|_F^2
\]
% DGB:
% \[
% _{\rm DGB}^2 = \frac{2\rm gap}{\lambda}, 
%\bm{Q}^{\rm DGB} = \bm{M}
%\]
Then, the center and the radius of DGB are
\begin{align*}
r_{\rm DGB}^2 
 =& \frac{2\rm gap}{\lambda} \\
 =& 
 \frac{2}{\lambda} \Bigl(
 \sum_{ijl}(\ell(\langle \bm{M}, \bm{H}_{ijl}\rangle)+\ell^*(-\alpha_{ijl})) +\frac{\lambda}{2}\|\bm{M}\|_F^2+\frac{1}{2\lambda}\|\sum_{ijl}\alpha_{ijl}\bm{H}_{ijl}+\bm{\Gamma}\|_F^2  \Bigr) \\
 =& 
 \|\bm{M}\|_F^2
 + \frac{2}{\lambda} \left(
 \sum_{ijl}(\ell(\langle \bm{M}, \bm{H}_{ijl}\rangle)+\ell^*(-\alpha_{ijl})) \right)  + \frac{1}{\lambda^2}\|\sum_{ijl}\alpha_{ijl}\bm{H}_{ijl}+\bm{\Gamma}\|_F^2,
 \\
 \bm{Q}^{\rm DGB} &= \bm{M}.
\end{align*}

\paragraph*{(RPB)}

In RPB, we regard $\lambda_1$ as a target $\lambda$ for which we consider the range.
From the definition, we see
\begin{align*}
 \bm{Q}^{\rm RPB} 
 &= 
 \frac{\lambda_0+\lambda}{2\lambda}\bm{M}_0^\star
 \\
 &= 
 \frac{1}{2}\bm{M}_0^\star +
 \frac{\lambda_0}{2\lambda}\bm{M}_0^\star,\\
 r_{\rm RPB} 
 &=
 \frac{\lambda_0-\lambda}{2\lambda}\left\|\bm{M}_0^\star\right\|_F \\
 &= 
 - \frac{1}{2}\left\|\bm{M}_0^\star\right\|_F
 + \frac{\lambda_0}{2\lambda}\left\|\bm{M}_0^\star\right\|_F.
\end{align*}

\paragraph*{(RRPB)}

Here again, we regard $\lambda_1$ as a target $\lambda$ for which we consider the range.
First, we assume $ \lambda \le \lambda_0 $, then we have
% In RRPB, we replace $ \lambda_1 $ with $ \lambda $ and assume $ \lambda \le \lambda_0 $. 
\begin{align*}
 \bm{Q}^{\rm RRPB} 
 &= \frac{\lambda_0+\lambda}{2\lambda}\bm{M}_0 \\
 &= \frac{1}{2}\bm{M}_0 + \frac{\lambda_0}{2\lambda}\bm{M}_0,  \\
 r_{\rm RRPB} 
 &= \frac{\lambda_0-\lambda}{2\lambda}\|\bm{M}_0\|_F+ \frac{\lambda_0}{\lambda}\epsilon  \\
 &= - \frac{1}{2}\|\bm{M}_0\|_F
 + \frac{1}{\lambda} \left(
 \frac{\lambda_0}{2}\|\bm{M}_0\|_F+ \lambda_0\epsilon  
 \right).
\end{align*}
In the case of $\lambda\ge\lambda_0$, we have
\begin{align*}
 \bm{Q}^{\rm RRPB} 
 % &= \frac{\lambda_0+\lambda}{2\lambda}\bm{M}_0 \\
 &= \frac{1}{2}\bm{M}_0 + \frac{\lambda_0}{2\lambda}\bm{M}_0,\\
 r_{\rm RRPB} 
 &= \frac{\lambda-\lambda_0}{2\lambda}\|\bm{M}_0\|_F + \epsilon \\
 &= \left(\epsilon +  \frac{1}{2}\|\bm{M}_0\|_F \right) -\frac{\lambda_0}{2\lambda}\|\bm{M}_0\|_F.
\end{align*}

% --------------------------------------------------
\subsection{Proof of Theorem \ref{thm:RRPBrange} (Range Based Extension of RRPB)}
\label{app:RRPBrange}
In RRPB, we replace $ \lambda_1 $ with $ \lambda $ and assume $ \lambda \le \lambda_0 $. 
Then, 
\[
	\bm{Q}^{\rm RRPB}=\frac{\lambda_0+\lambda}{2\lambda}\bm{M}_0,~~
	r_{\rm RRPB}=\frac{\lambda_0-\lambda}{2\lambda}\|\bm{M}_0\|_F+ \frac{\lambda_0}{\lambda}\epsilon.
\]
From Sphere rule \eqref{eq:sphere-ruleR2}, we obtain
\begin{align*}
&\frac{\lambda+\lambda_0}{2\lambda}\langle \bm{H}_{ijl}, \bm{M}_0\rangle-(\frac{\lambda_0-\lambda}{2\lambda}\|\bm{M}_0\|_F+ \frac{\lambda_0}{\lambda}\epsilon)\|\bm{H}_{ijl}\|_F>1\\
\Leftrightarrow&
\underbrace{(\langle \bm{H}_{ijl}, \bm{M}_0\rangle-2+\|\bm{H}_{ijl}\|_F\|\bm{M}_0\|_F)}_{>0 \text{ is required.}}\lambda >\lambda_0(\underbrace{\|\bm{M}_0\|_F\|\bm{H}_{ijl}\|_F-\langle \bm{H}_{ijl}, \bm{M}_0\rangle}_{\ge0}+2\epsilon\|\bm{H}_{ijl}\|_F).
\end{align*}
From Cauchy-Schwarz inequality, the right hand side is equal or greater than 0.
Therefore, the left hand side must be greater than 0.
\[
\therefore~
\lambda_0\ge
\lambda>\frac{\lambda_0(\|\bm{M}_0\|_F\|\bm{H}_{ijl}\|_F-\langle \bm{H}_{ijl}, \bm{M}_0\rangle+2\epsilon\|\bm{H}_{ijl}\|_F)}{\langle \bm{H}_{ijl}, \bm{M}_0\rangle-2+\|\bm{H}_{ijl}\|_F\|\bm{M}_0\|_F}.
\]

In the case of $\lambda\ge\lambda_0$, 
\[
	\bm{Q}^{\rm RRPB}=\frac{\lambda_0+\lambda}{2\lambda}\bm{M}_0,~~
	r_{\rm RRPB}=\frac{\lambda-\lambda_0}{2\lambda}\|\bm{M}_0\|_F + \epsilon.
\]
From Sphere Rule \eqref{eq:sphere-ruleR2}, 
\begin{align*}
&\frac{\lambda+\lambda_0}{2\lambda}\langle \bm{H}_{ijl}, \bm{M}_0\rangle-(\frac{\lambda-\lambda_0}{2\lambda}\|\bm{M}_0\|_F+ \epsilon)\|\bm{H}_{ijl}\|_F>1\\
\Leftrightarrow&
(\underbrace{\|\bm{H}_{ijl}\|_F\|\bm{M}_0\|_F-\langle \bm{H}_{ijl}, \bm{M}_0\rangle}_{\ge0}+2+2\epsilon\|\bm{H}_{ijl}\|_F)\lambda<\lambda_0(\|\bm{M}_0\|_F\|\bm{H}_{ijl}\|_F+\langle \bm{H}_{ijl}, \bm{M}_0\rangle).
\end{align*}
Similarly, from Cauchy-Schwarz inequality, the left hand side is greater than 0.
\[
\therefore~
\lambda_0\le
\lambda
<\frac{\lambda_0(\|\bm{M}_0\|_F\|\bm{H}_{ijl}\|_F+\langle \bm{H}_{ijl}, \bm{M}_0\rangle)}{\|\bm{H}_{ijl}\|_F\|\bm{M}_0\|_F-\langle \bm{H}_{ijl}, \bm{M}_0\rangle+2+2\epsilon\|\bm{H}_{ijl}\|_F}.
\]

% --------------------------------------------------
\subsection{Additional Remark for Range Extension of RRPB}

Since RRPB can evaluate the bound for $\lambda \neq \lambda_0$ based only on the optimality $\epsilon (\geq \| \bm{M}_0^\star - \bm{M}_0 \|_F)$ for $\lambda_0$, RRPB is particularly suitable to the range based screening among the spheres we derived so far.
For the other spheres (i.e., GB and DGB), we need all triplets including triplets currently screened into $\hat{\cL}$ and $\hat{\cR}$ at $\lambda_0$, to which we refer as $\hat{\cL}_{\lambda_0}$ and $\hat{\cR}_{\lambda_0}$. 
For example, in the case of DGB, the duality gap for $\lambda \in (\lambda_a, \lambda_b)$ should be considered, which depends on $P_{\lambda}(\bm{M}_0)$.
However, we cannot replace $P_{\lambda}(\bm{M}_0)$ for $\lambda \neq \lambda_0$ with $\tilde{P}_{\lambda}(\bm{M}_0)$, without guaranteeing the following conditions:
\begin{align*}
 \langle \bm{M}_0, \bm{H}_{ijl}\rangle &< 1-\gamma, \text{ for } \forall (i,j,l)\in
 \hat{\cL}_{\lambda_0},
 \\
 \langle \bm{M}_0, \bm{H}_{ijl}\rangle &> 1, \text{ for } \forall (i,j,l)\in
 \hat{\cR}_{\lambda_0}.
\end{align*}
When the reference solution $\bm{M}_0$ is theoretically optimal for $\lambda_0$, these conditions also hold, but this cannot be true for the numerical calculation, and further usually we stop the optimization with some tolerance level specified beforehand.

Although it is possible to create $\hat{\cL}_{\lambda_0}$ and $\hat{\cR}_{\lambda_0}$ by taking the tolerance level of optimization into consideration, the range calculation becomes quite complicated in this case shown as the next subsection.
% (see Appendix~\ref{sec:approxBound}.

\subsection{Bound Including Approximate Solution}
\label{sec:approxBound}

The sphere derived in the section~\ref{sec:bound} contains the optimal solution.
By enlarging the radius, we can also create a bound which contains the approximate solution obtained by an output of any optimization algorithm with some termination condition. 
If the termination condition is duality gap $ \mathrm{gap} \le \epsilon $, 
the distance between approximate solution $ \hat{\bm{M}} $ and optimal solution $ \bm{M}^\star $ 
satisfies
% is 
%suppressed by
\[
\|\bm{M}^\star-\hat{\bm{M}}\|_F\le\sqrt{\frac{2\rm gap}{\lambda} }\le\sqrt{\frac{2\epsilon}{\lambda} }.
\]
Since the distance between the approximate solution and the optimal solution is at most $ \sqrt{\frac{2\epsilon}{\lambda}} $, 
enlarging the radius of the hypersphere 
% containing the optimal solution 
by $ \sqrt{\frac{2\epsilon}{\lambda}} $, 
% it 
we
can guarantee that the bound includes an approximate solution. 
Using the radius $ r $ of the section \ref{sec:screeningRange} for the expanded radius $ R $,
\[
R=\underbrace{\sqrt{a+b\frac{1}{\lambda}+c\frac{1}{\lambda^2} }}_{r} + \sqrt{\frac{2\epsilon}{\lambda}}.
\]
% Then, trying 
To find a range of $ \lambda $, which can be screened, we need to consider a complicated quartic inequality of $\lambda$.
Another drawback of this approach is that the increase of the radius may cause the drop of the screening rate.
% Also, there is a problem that screening rate drops as the radius increases.

% --------------------------------------------------
\section{Additional Experimental Results}

We here show some additional results which we omit in the main text.
\tablename~\ref{tbl:dataset-app} shows datasets, some of which are already shown in section~\ref{sec:experiment}.

\begin{table*}[t]
 \caption{
 Summary of datasets.
 $*1 : $The dimension was reduced by AutoEncoder.
 $*2 : $The dimension was reduced by PCA.
 \#triplet and $\lambda_{\min}$ are the average value for sub-sampled random trials.
 }
 \label{tbl:dataset-app}
	\centering
	{\footnotesize\tabcolsep=1mm
		\begin{tabular}{l||c|c|c|c|c|c|c|c|c|c}
		dataset				&iris	&wine	&segment	&satimage	&phishing	&SensIT Vehicle	&a9a	&mnist	&cifar10&rcv1.multiclass\\ \hline
		\#dimension			&4			&13			&19			&36			&68			&100	&16${}^{*1}$&32${}^{*1}$&200${}^{*1}$&200${}^{*2}$\\
		\#samples			&150		&178		&2310		&4435		&11055		&78823		&32561		&60000		&50000		&15564		\\
		\#classes			&3			&3			&7			&6			&2			&3			&2			&10			&10			&53			\\
		$k$					&$\infty$	&$\infty$	&20			&15			&7			&3			&5			&5			&2			&3			\\
		\#triplet			&546668		&910224		&832000		&898200		&487550		&638469		&732625		&1350025	&180004		&126018		\\
		$\lambda_{\max}$	&1.3e$+$7	&2.0e$+$7	&2.5e$+$6	&1.0e$+$7	&5.0e$+$3	&1.0e$+$4	&1.2e$+$5	&7.0e$+$3	&2.0e$+$3	&3.0e$+$2	\\
		$\lambda_{\min}$	&2.3e$+$1	&5.1e$+$1	&4.2e$+$0	&8.8e$+$0	&2.0e$-$1	&2.9e$+$0	&3.1e$+$2	&9.6e$-$1	&3.3e$+$1	&6.0e$-$4	\\
		\end{tabular}
	}
\end{table*}

% --------------------------------------------------
\subsection{Evaluation for Hinge Loss}
\label{app:eval-hinge}

\begin{figure}[t]
 \centering
 \includegraphics[clip,width=\linewidth]{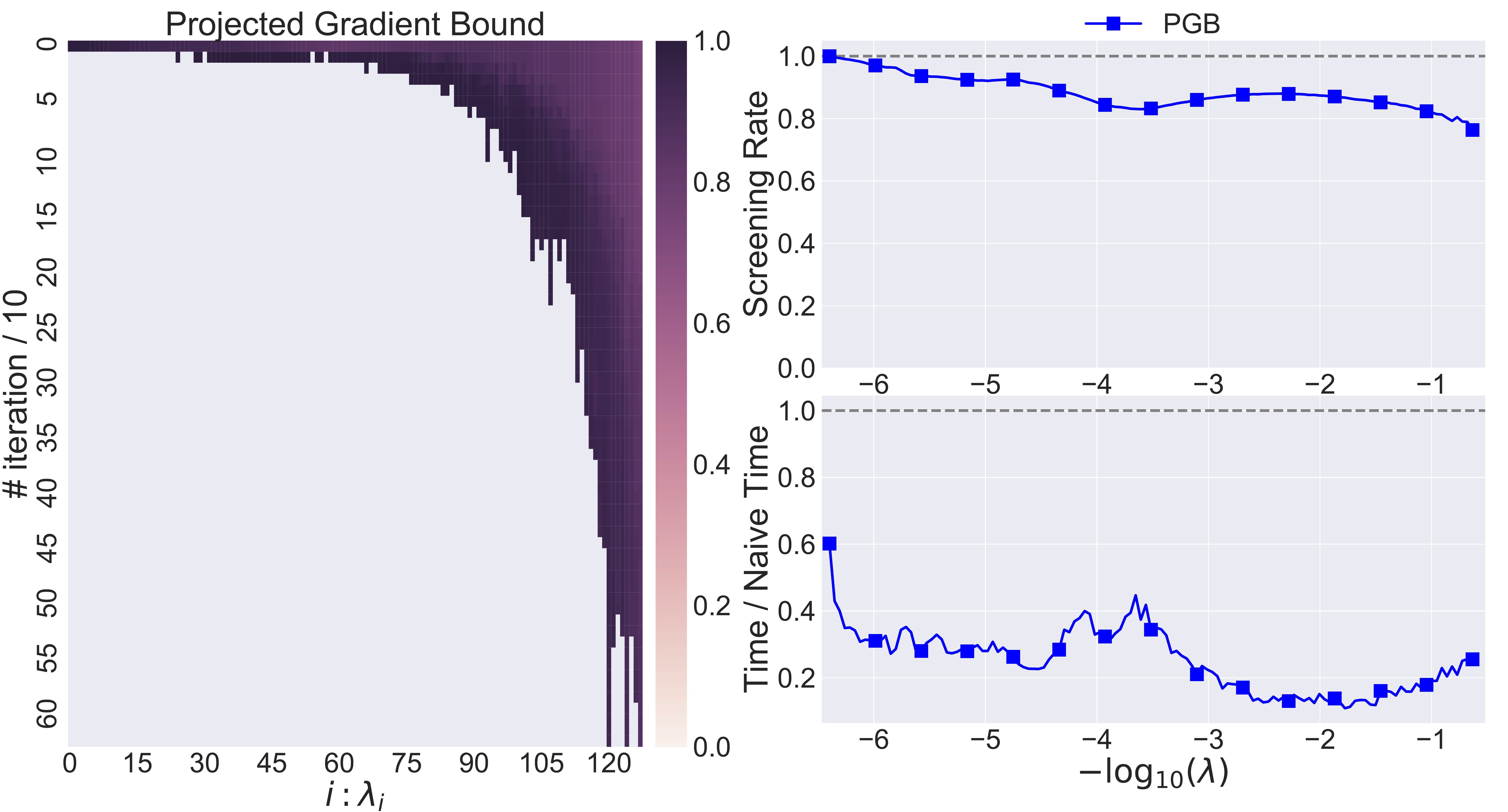}
 \caption{
 Performance evaluation of PGB for the hinge loss setting. % ($\gamma = 0$).
 }
 \label{fig:hingePGB}
\end{figure}

\figurename~\ref{fig:hingePGB} shows the screening result of the PGB sphere rule for segment data.
Here, the loss function of RTLM is the hinge loss function, and the other settings are the same as the experiments in the main text.
We see that PGB achieved the high screening rate and the CPU time was substantially improved.

% --------------------------------------------------
% Hinge loss + PGB + Segment data
% --------------------------------------------------

% --------------------------------------------------
\subsection{Comparing DGB Based Rules}
\label{app:compareDGB}

\begin{figure}[t]
 \centering
 \includegraphics[clip,width=\linewidth]{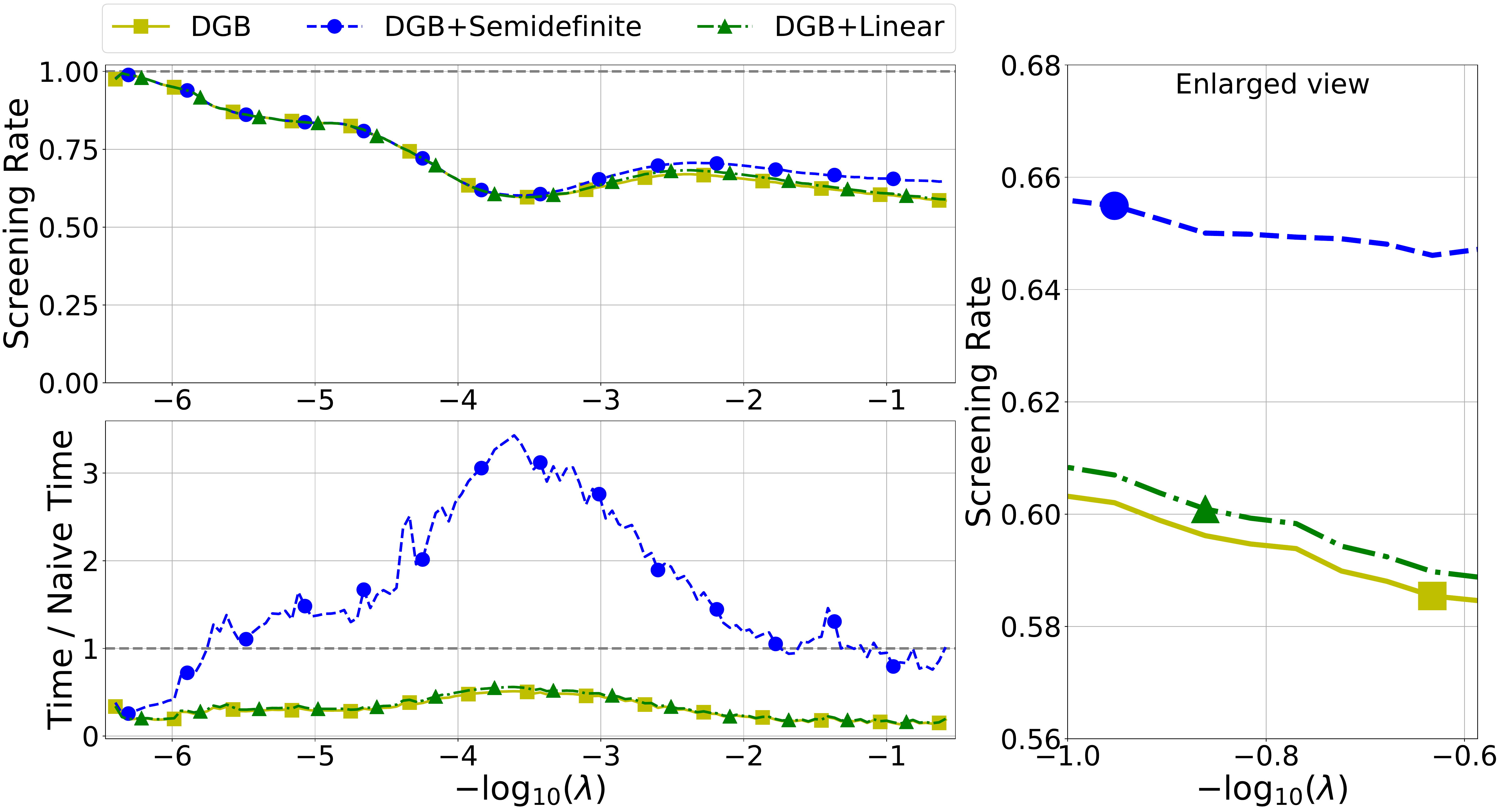}
 \caption{
 Screening rule comparison on the segment dataset (DGB).
 }
 \label{fig:segmentDGB}
\end{figure}

Using DGB, we compared performance of the three rules in section~\ref{sec:screening}.
\figurename~\ref{fig:segmentDGB} shows the results.
We see the similar tendency to the case of GB shown in \figurename~\ref{fig:segmentComp}. 
Although the semi-definite and the linear constraint slightly improves the rate, clear improvement in the CPU time was not observed.

% --------------------------------------------------
% DGB with constraint
% --------------------------------------------------

% --------------------------------------------------
\subsection{Total Time for Bound Comparison}
\label{app:compBounds}

The total CPU time for regularization path performed in section~\ref{sec:exp-bounds} is shown in \tablename~\ref{tbl:time}.

\begin{table}[t]
 \centering
 \caption{
 Total CPU time of regularization path (sec). 
 % Total time of regularization path.
 Parentheses indicate the time taken for the screening evaluation.
 RRPB + PGB indicates that both of sphere rules derived by each sphere are performed.
 }
 \label{tbl:time}
	{\scriptsize\tabcolsep=1mm
		\begin{tabular}{l||r|r|r|r|r|r}
		Bound\textbackslash Dataset	&iris		&wine		&segment	&satimage	&phishing	&SensIT		\\ \hline \hline
		---								&110.9		&564.1		&1327.8		&1970.9		&5584.7		&4679.8		\\ \hline
		GB								&71.5		&545.3		&1220.0		&1727.2		&3139.1		&3950.2		\\ 
										&(14.4)		&(59.6)		&(122.6)	&(197.3)	&(263.4)	&(445.1)	\\ \hline
		PGB								&31.6		&129.3		&235.6		&653.5		&1300.3		&2534.9		\\
										&(12.8)		&(35.9)		&(59.4)		&(131.0)	&(143.9)	&(352.0)	\\ \hline
		DGB								&23.7		&153.3		&300.9		&799.7		&1748.3		&2927.4		\\
										&(0.5)		&(1.1)		&(1.2)		&(1.5)		&(0.9)		&(0.8)		\\ \hline
		RRPB							&$^\star$20.0&$^\star$104.7&206.3		&651.2		&1397.6		&2596.2		\\
										&(0.5)		&(1.1)		&(1.2)		&(1.5)		&(0.8)		&(0.8)		\\ \hline \hline
		RRPB + PGB						&20.8		&105.1		&$^\star$188.1&$^\star$569.6&$^\star$1247.4&$^\star$2382.1\\
										&(0.9)		&(6.5)		&(11.2)		&(39.7)		&(60.7)		&(186.8)	\\
		\end{tabular}
	}
\end{table}

% --------------------------------------------------
\subsection{Experiment for Higher Dimensional Data in Diagonal Matrix}
\label{app:highdimRslt}

\begin{table}[t]
 \caption{Total time (seconds) of the regularization path in diagonal matrix. 
 The results with $^\star$ indicate the fastest method.
 The result of gisette dataset by ActiveSet was not obtained because of the time limitation.
 }
	\label{tbl:diag}
	\centering
	{\scriptsize\tabcolsep=1mm
	\begin{tabular}{lrrrr}
	Method\textbackslash Dataset	&usps			&madelon		&colon-cancer	&gisette			\\ \hline
	ActiveSet						&2485.5			&7005.8			&3149.8			& -					\\
	ActiveSet+RRPB					&$^\star$326.7	&593.4			&632.2			&133870.0			\\
	ActiveSet+RRPB+PGB				&336.6			&$^\star$562.4	&$^\star$628.2	&$^\star$127123.8 	\\ \hline
	\#dimension						&256			&500			&2000			&5000				\\
	\#samples						&7291			&2000			&62				&6000				\\
	\#triplet						&656200			&720400			&38696			&1215225			\\
	$k$								&10				&20				&$\infty$		&15					\\
	$\lambda_{\max}$				&1.0e+7			&2.0e+14		&5.0e+7			&4.5e+8				\\
	$\lambda_{\min}$				&1.9e+3			&4.7e+11		&7.0e+3			&2.1e+3				\\
	\end{tabular}
	}
\end{table}

We here used higher dimensional datasets.
To mitigate computational difficulty, we employed restrict $\bm{M}$ to diagonal matrix.
% We restrict $\bm{M}$ to diagonal matrix $\mathrm{diag}(\bm{m})$ and learn distance metric $\bm{m}$.
%
In the same setting as section~\ref{sec:pracEval}, for the higher dimensional datasets, the result for comparison with the ActiveSet method is shown in \figurename~\ref{tbl:diag}. 
Even in the case of the diagonal matrix, we see that our triplet screening is effective.
% it turns out that combining Screening is faster. 
%
In the gisette dataset, the active set method did not terminate even after 250,000 sec, at which about $3/4$ of the regularization path was calculated.
Since the larger number of iterations of the gradient descent is usually necessary for smaller $\lambda$, the entire calculation would take more than $250,000 \times (4/3)\approx 333,333$.
\end{document}